\documentclass[lettersize,journal]{IEEEtran}
\usepackage{mathtools}      
\usepackage{algorithm}
\usepackage{algorithmic}
\usepackage{amsthm}
\usepackage{amsmath}
\usepackage{amsfonts}       
\usepackage{nicefrac}       
\usepackage{dsfont}         
\usepackage{amssymb}        
\usepackage{physics}        
\usepackage{xcolor} 
\usepackage{colortbl}
\usepackage[utf8]{inputenc} 
\usepackage[T1]{fontenc}    
\usepackage{bbm}            
\usepackage{hyperref}       
\usepackage{orcidlink}      
\usepackage{url}            
\usepackage{microtype}      
\usepackage[caption=false]{subfig}
\usepackage{textcomp} 
\usepackage{pifont}   
\usepackage{numprint} 
\usepackage{stfloats} 
\usepackage{float}    
\usepackage{verbatim} 
\usepackage{graphicx} 
\usepackage{epsfig} 
\usepackage{cite}
\usepackage{enumitem} 

\usepackage{todonotes}
\usepackage{booktabs}   
\usepackage{array}      
\usepackage{multirow}   
\usepackage{multicol}
\usepackage{xspace}     
\usepackage{tikz} 
\usetikzlibrary{calc}
\usepackage{linegoal}
\usepackage[edges]{forest}
\definecolor{quantblue}{RGB}{70,130,180}   
\definecolor{calibred}{RGB}{220,20,60}     
\definecolor{lightblue}{RGB}{173,216,230}  
\definecolor{lightred}{RGB}{255,182,193}   
\tikzset{%
    parent/.style = {align=center,text width=4.2cm, rounded corners=4pt, fill=gray!50, draw=black, line width=0.4mm, rotate=90}, 
    referenceblock_l1/.style = {
        align=left, 
        rounded corners=2pt, 
        fill=white, draw=black, 
        line width=0.2mm,  
        text width=13.2cm,
    },
    referenceblock_l2/.style = {
        align=left, 
        rounded corners=2pt, 
        fill=white, draw=black, 
        line width=0.2mm,  
        text width=11.3cm,
    },
    referenceblock_l3/.style = {
        align=left, 
        rounded corners=2pt, 
        fill=white, draw=black, 
        line width=0.2mm,  
        text width=9.4cm,
    },
    quantchild/.style = {align=center, text width=1.6cm, rounded corners=3pt, fill=quantblue!30, draw=quantblue, line width=0.3mm},
    quantgrandchild/.style = {align=center, text width=1.4cm, rounded corners=3pt, fill=lightblue!30, draw=quantblue, line width=0.2mm},
    calibchild/.style = {align=center, text width=1.6cm, rounded corners=3pt, fill=calibred!30, draw=calibred, line width=0.3mm},
    calibgrandchild/.style = {align=center, text width=1.4cm, rounded corners=3pt, fill=lightred!30, draw=calibred, line width=0.2mm},           
}

\newcommand{\cmark}{\textcolor{black}{\ding{51}}}%
\newcommand{\xmark}{\textcolor{black}{\ding{55}}}%
\newcommand{\smark}{\textcolor{black}{$\sim$}}%

\newif\ifreviewmode
\reviewmodefalse  

\ifreviewmode
  \newcommand{\edit}[1]{\textcolor{purple}{#1}}
  \newcommand{\reviewerOne}[1]{\textcolor{teal}{#1}}
  \newcommand{\reviewerTwo}[1]{\textcolor{violet}{#1}}
  \newcommand{\revision}[1]{\textcolor{orange}{#1}}
\else
  \newcommand{\edit}[1]{#1}
  \newcommand{\reviewerOne}[1]{#1}
  \newcommand{\reviewerTwo}[1]{#1}
  \newcommand{\revision}[1]{#1}
\fi

\makeatletter
\DeclareRobustCommand\onedot{\futurelet\@let@token\@onedot}
\def\@onedot{\ifx\@let@token.\else.\null\fi\xspace}

\def\etal{\emph{et al}\onedot}
\makeatother

\usepackage{lastpage} 

\newcommand{\Title}{Comparing Uncertainty Measurement and Mitigation Methods for Large Language Models: A Systematic Review}
\begin{document}

\title{\Title}

\author{ 
Toghrul Abbasli\,\orcidlink{0009-0006-8160-2346}, Kentaroh Toyoda\,\orcidlink{0000-0002-6233-3121}, Yuan Wang\,\orcidlink{0000-0002-5258-1500}, Leon Witt\,\orcidlink{0000-0002-9984-3213}, Muhammad Asif Ali\,\orcidlink{0000-0002-9365-4817}, Yukai Miao\,\orcidlink{0000-0002-6401-5979}, Dan Li\,\orcidlink{0000-0002-7581-8865}, and Qingsong Wei\,\orcidlink{0000-0003-4570-7328}
\thanks{Manuscript received 16 April 2025; revised 23 September 2025, 11 March 2026 and 15 July 2026; accepted 4 August 2026. This research is supported by the RIE2025 Industry Alignment Fund – Industry Collaboration Project (IAF-ICP) (Award No: I2301E0020) and Japan-Singapore Joint Call: Japan Science and Technology Agency (JST) and Agency for Science, Technology and Research (A*STAR) 2024 (Award No: R24I6IR141), administered by A*STAR, and is supported by the Beijing Outstanding Young Scientist Program (No.JWZQ20240101008). \textit{(Corresponding author: Toghrul Abbasli.)}
}
\thanks{Toghrul Abbasli, and Dan Li are with the Department of Computer Science and Technology, Tsinghua University, Beijing, China (e-mail: tgl22@mails.tsinghua.edu.cn; tolidan@tsinghua.edu.cn).}
\thanks{Kentaroh Toyoda is with Vulcan Research, AIFT, Singapore, Keio Global Research Institute (KGRI), Japan (e-mail: kentaroh.toyoda@ieee.org).}%
\thanks{Yuan Wang is with China Mobile Research Institute, China (e-mail: wangyuanyjy@chinamobile.com).}
\thanks{Leon Witt is with Shanghai Institute for Mathematics and Interdisciplinary Sciences (SIMIS), China (email: leon.witt@simis.cn).}
\thanks{Muhammad Asif Ali is with the Department of Computer, Electrical and Mathematical Sciences and Engineering (CEMSE), King Abdullah University of Science and Technology, Thuwal, Makkah, Kingdom of Saudi Arabia, and also with Information Technology University, Lahore, Pakistan (e-mail: muhammadasif@itu.edu.pk).}
\thanks{Yukai Miao is with Zhongguancun Laboratory, Beijing, China (e-mail: miaoyk@zgclab.edu.cn  ).}
\thanks{Qingsong Wei is with Institute of Advanced Intelligence and Computing (IAIC), Agency for Science, Technology and Research (A*STAR), Singapore (e-mail: wei\_qingsong@a-star.edu.sg).}
}

\markboth{\Title}%
{Anonymous \MakeLowercase{\textit{et al.}}: \Title}


\maketitle

\begin{abstract}
Large Language Models (LLMs) have been transformative across many domains. However, hallucination, i.e., confidently
outputting incorrect information, remains one of the leading challenges for LLMs. This raises the question 
of how to accurately assess and quantify the uncertainty of LLMs.
Extensive literature on traditional models has explored Uncertainty Quantification (UQ) to measure uncertainty 
and employed calibration techniques to address the misalignment between uncertainty and accuracy. While some 
of these methods have been adapted for LLMs
, the literature 
lacks an in-depth analysis of their effectiveness and does not offer a comprehensive benchmark to enable insightful 
comparison among existing solutions. In this work, we fill this gap via a systematic survey of representative 
prior works on UQ and calibration for LLMs and introduce a 
rigorous benchmark. Using three widely used reliability datasets, we empirically evaluate seven
related methods, which justify the significant findings of our review.
Finally, we provide outlooks for key future directions and outline open challenges. 
To the best of our knowledge, this survey is one of the first dedicated studies to review the calibration methods and relevant metrics for LLMs.
\footnote{Review list and source code\hspace{0.2cm}: \hspace{0.3cm}\url{https://github.com/userTogrul/large-model-calibration-and-uncertainty}}

\end{abstract}


\section{Introduction}
\label{sec:introduction}

\subsection{Overview of Uncertainty and Calibration}

Large language models (LLMs) encompass a massive amount of parameters and are trained using extensively large-scale datasets to support diverse applications such as machine translation, content summarization, question answering, coding, and open conversation \cite{brown2020languagegpt3, achiam2023gpt4, hurst2024gpt4o, guo2025deepseekrone}.  In past years, research and development practice has witnessed the power of scaling law, which shows that up-scaling the size of LLMs dramatically improves algorithmic, inference, and reasoning performance, a behavior widely known as the emergent phenomena~\cite{brown2020languagegpt3, achiam2023gpt4, chang2024surveyllmevaluation, phasetransition2025}. However, these advancements have trade-offs, such as formidable costs and complexity. Meanwhile, the increase in the size of LLMs makes the interpretation and understanding of the internal scientific mechanisms and principles even harder~\cite{walker2003defining, kailkhura2019reliablex, ovadia2019can, zhou2023navigating, ulmer2024apricot}. 

For these reasons, research on the reliability of LLMs is crucial for enhancing stability and security while uncovering risks like misinformation, degraded cognitive capabilities, and hallucinations, especially in sectors such as finance, healthcare, and autonomous systems~\cite{deng2023agreementscore, danruo2023uncertainty, chang2024surveyllmevaluation}. In other words, to achieve reliable performance, it is imperative to enable LLMs to \emph{``know what they do not know''} in addition to tuning the prediction accuracy alone~\cite{niculescu2005predictingreldiag, yuksekgonul2024atypicalitycalib}. To this end, \textit{uncertainty} serves as an important indicator of noisy, in-concrete categories or unknown content in decision-making \cite{blundell2015bayesbybackprop, tran2022plex, yin2023llmknowwhattheydont}. 
Since uncertainty originates from both the internal structure of the model and the external environment, \revision{uncertainty quantification (UQ), i.e., detecting and measuring these sources, is} fundamental to enhance the reliability of LLMs \cite{kendallgal2017whatuncertaintiesweneed, malinin2018predictivepriornetworks, danruo2023uncertainty}. 
Meanwhile, LLMs must be capable of communicating the uncertainty of their predictions, a feature essential for users to know when to trust the model outcomes \cite{lin2022teachingmodelstoexpress, xu2024sayself, ganguli2022redteamingllms, ulmer2024apricot}.


Although uncertainty can offer valuable insights into an LLM's competence, these models often exhibit miscalibrated generations, displaying either over-confidence or under-confidence in their predictions, which provide incorrect measures of uncertainty \cite{Braverman2019CalibrationER, hendrycks2021unsolved, Kalai2023CalibratedLMmustHallucinate, yuksekgonul2024atypicalitycalib, zhang2024luq}. This misalignment often results in a probability of prediction generation that does not reflect the model's actual evaluation performance and introduces bias to the outputs \cite{guo2017calibration, 
brown2020languagegpt3, hendrycks2021unsolved, si2022promptinggpt3reliable, cohen2023lmvslm}. Hence, calibrating models' confidence is crucial for ensuring the reliable measurement and the accuracy of their probabilistic inferences. 

While recent surveys focus on works about uncertainty-aware techniques and hallucination \cite{xiao2021hallucinationanduncertainty, zablotskaia2023uncertaintybenchmark, zhou2023batchcalibration, 
ulmer2024apricot, farquhar2024detecting, baan2023uncertaintyfromtheorytoapplications, geng2024surveyllmcal, gawlikowski2023survey}, we observe a lack of systematic review of calibration and evaluation of these methods for LLMs. This work seeks to bridge that gap by providing a comprehensive review of LLM calibration and uncertainty, by answering to the research questions (RQs) we set, and with the empirical analysis of them for the generation task. 






Our findings from the systematic review with PRISMA \cite{page2021prisma} method and generative question answering (QA) experiments revealed that recent LLMs are quite under-confident and over-confident in their predictions, and calibration methods are necessary to tackle this issue. While existing metrics are straightforward for measuring accuracy, calibration, and uncertainty, they fail to capture diverse nature predictions and the semantic and syntactic information of longer-form generations. We find that existing methods of calibration need to be adapted to internal layers and generation tasks to elicit reliable confidence. On the other hand, smaller LMs require a better calibration method.

\subsection{Contributions}
\label{sec:contribution}
The main contributions of the paper can be listed as follows:

\begin{enumerate}
    \item To the best of our knowledge, this work is among the first to systematically review a broad set of calibration and uncertainty quantification (UQ) methods, considering both open-box and closed-box LLMs.

    \item We conduct comparative experiments on LLMs using both calibration-specific and general performance metrics, addressing the current lack of effective evaluation regarding model truthfulness and uncertainty reasoning in generative QA tasks. While existing metrics are adequate for measuring accuracy and calibration, they do not fully capture the complexity of uncertainty or the semantic and syntactic richness of longer-form generations. We recommend adapting current methods to better account for internal model architectures and generation tasks in order to improve confidence estimation.

    \item Our results show that smaller language models require improved calibration, whereas larger LLMs, with greater model and vocabulary sizes, can reduce output uncertainty through reasoning steps. For example, we observe that applying scaling to Mistral v0.3 7B improves its confidence estimates, and that models employing Chain-of-Thought (CoT) tend to be better calibrated. Experiments with Natural Questions (NQ) exhibit increased miscalibration, suggesting that factors such as question difficulty, distribution shift, and ambiguity affect model uncertainty.

    \item We identify and summarize key research challenges and open problems to guide and accelerate future work. We hope our study increases awareness of the importance of aligning uncertainty estimation and accuracy in LLMs.
\end{enumerate}

\noindent{\bf Organization.} \edit{The remainder of the paper is structured as follows: Section~\ref{sec:methodology} presents the problem statement, reviews related surveys, and details our survey methodology. Section~\ref{sec:llm_uncertainty} provides the definition of uncertainty and UQ methods for DL and LLMs.} Section~\ref{sec:llm_calibration} examines calibration approaches for LLMs and the applicability of existing methods. Section~\ref{sec:evaluation} covers the evaluation of these methods and analyzes the results. Section~\ref{sec:emerging_topics} discusses key findings, emerging topics, and future research challenges. Finally, Section~\ref{sec:conclusion} concludes the paper.

\section{Survey Methodologies}
\label{sec:methodology}

\subsection{Related Survey Papers}

We examine existing survey papers to identify key research questions and determine gaps that need further investigation.
Table~\ref{tab:survey_comparison} shows a detailed comparison of related review works. We mark topics that are covered as \cmark, topics that are not covered as \xmark, and topics that are briefly mentioned or slightly covered as \smark. Categories: UQ: uncertainty quantification; DL: classical deep learning setting; LLM: large language models setting.

\revision{Prior surveys fall into three groups: (i) DL-focused UQ~\cite{zhang2021modern, gawlikowski2023survey, baan2023uncertaintyfromtheorytoapplications, yang2024generalizedoodsurvey, wenchong2026asurveyonuqmethodsfordl}, (ii) LLM-trustworthiness and hallucination-oriented reviews~\cite{liu2023trustworthy, ji2023surveyofhallucinationinnlg, chang2024surveyllmevaluation, hendrycks2021unsolved}, and (iii) LLM UQ or calibration treated in isolation~\cite{geng2024surveyllmcal, shorinwa2025surveyuqllm, xia2025asurveyofuellms}; none jointly address UQ \emph{and} calibration for LLMs with a method-level comparison grounded in an empirical evaluation, which is the gap our work fills.}

\edit{Existing surveys on uncertainty quantification (UQ) and calibration predominantly focus on classical DL models~\cite{gawlikowski2023survey, hendrycks2021unsolved}, leaving key challenges for LLMs underexplored. For example,~\cite{liu2023trustworthy} centers on LLM trustworthiness and alignment with human preferences, but only briefly touches on miscalibration. Others, such as~\cite{zhang2021modern}, restrict their scope to Monte Carlo (MC)-based UQ for classical DL. Recent surveys~\cite{geng2024surveyllmcal, zablotskaia2023uncertaintybenchmark} discuss calibration and UQ in LLMs and propose benchmarks, yet do not systematically compare state-of-the-art methods or address the unique challenges of LLMs. Additional surveys~\cite{kumar2019verified, yang2024generalizedoodsurvey} focus on applications like selective inference and out-of-distribution (OOD) detection rather than UQ methodology itself. While~\cite{shorinwa2025surveyuqllm} attempts to bridge DL and LLMs in UQ, it provides only a limited review of calibration and lacks a thorough comparison. The survey in~\cite{xia2025asurveyofuellms} broadly covers UQ for LLMs with calibration as a secondary component, offering stronger method-level comparison and evaluation-oriented synthesis. In contrast,~\cite{ji2023surveyofhallucinationinnlg} focuses on hallucination in natural language generation, so its treatment of UQ and calibration is limited and indirect, without a dedicated comparison of UQ/calibration methods or a calibration-centered evaluation framework.}

\reviewerTwo{Most recently, \cite{wenchong2026asurveyonuqmethodsfordl} present an extensive taxonomy of UQ methods for deep neural networks (DNNs) organized by uncertainty source (data vs.\ model), summarize the strengths and weaknesses of each category, and illustrate applications such as active learning, OOD robustness, and deep reinforcement learning; LLMs are identified as a future-work direction rather than a focal topic. Our review is complementary: it is explicitly LLM-centric, covers both UQ \emph{and} calibration, and grounds the comparison with a generative-task benchmark across multiple closed-box and open-box LLMs. We position both works in Table~\ref{tab:survey_comparison}, where \cite{wenchong2026asurveyonuqmethodsfordl} covers DL-side UQ fully and method comparison strongly but does not provide LLM-specific UQ or empirical evaluation.}

\edit{Recent TNNLS work on uncertainty and calibration likewise remains centered on classical DL and vision: Dong~\etal~\cite{dong2025confcalibsurvey} survey confidence calibration for class-imbalanced deep classification (about sixty methods across six categories); Wang~\etal~\cite{wang2025conformalkeypoint} use inductive conformal prediction for coverage-guaranteed pose uncertainty; and Zhao~\etal~\cite{zhao2024oodcrossclass} improve out-of-distribution confidence for image classifiers. None address LLMs or open-ended generative QA, and the only survey among them~\cite{dong2025confcalibsurvey} (added to Table~\ref{tab:survey_comparison}) is confined to class-imbalanced classification calibration. Our review fills this gap, extending UQ and calibration to LLMs with a generative-task benchmark.}

\edit{
In contrast, our work is, to our knowledge, among the first to jointly benchmark UQ and calibration for LLMs on open-ended generative QA: we evaluate state-of-the-art approaches on current models and datasets, highlight LLM-specific challenges, and offer a detailed comparative evaluation, providing a much-needed resource for researchers and practitioners.
}

\begin{table}[!t]
    \caption{A comparison of survey papers based on topic criteria.}
    \centering
    \begin{tabular}{
    >{\arraybackslash}p{0.06\textwidth}
    >{\centering\arraybackslash}p{0.05\textwidth}
    >{\centering\arraybackslash}p{0.05\textwidth}
    >{\centering\arraybackslash}p{0.06\textwidth}
    >{\centering\arraybackslash}p{0.06\textwidth}
    >{\centering\arraybackslash}p{0.06\textwidth}}
        \toprule
        \multirow{2}{=}{\textbf{Review}} & 
        \multicolumn{2}{c}{\textbf{UQ}} & 
        \multirow{2}{*}{\textbf{Calibration}} &  
        \multirow{2}{=}{\centering \textbf{Method \\ Comparison}} & 
        \multirow{2}{*}{\textbf{Evaluation}} \\ 
        \cmidrule(lr){2-3} & \textbf{DL} & \textbf{LLM} & & & \\
        \midrule
        \cite{zhang2021modern} & \cmark & \xmark & \xmark & \smark & \xmark \\
        \cite{hendrycks2021unsolved} & \smark & \smark & \smark & \xmark & \xmark \\
        \cite{baan2023uncertaintyfromtheorytoapplications} & \cmark & \smark & \xmark & \cmark & \xmark \\
        \cite{gawlikowski2023survey} & \cmark & \xmark & \cmark & \xmark & \xmark  \\
        \cite{liu2023trustworthy} & \xmark & \cmark & \cmark & \xmark & \smark \\
        \cite{ji2023surveyofhallucinationinnlg} & \xmark & \smark & \xmark & \smark & \smark \\
        \cite{chang2024surveyllmevaluation} & \xmark & \cmark & \xmark & \xmark & \xmark \\
        \cite{yang2024generalizedoodsurvey} & \smark & \xmark & \xmark & \xmark & \xmark \\
        \cite{geng2024surveyllmcal} & \smark & \cmark & \cmark & \smark & \xmark \\
        \cite{shorinwa2025surveyuqllm} & \cmark & \cmark & \smark & \xmark & \xmark \\
        \cite{xia2025asurveyofuellms} & \smark & \cmark & \xmark & \smark & \cmark \\
        \reviewerTwo{\cite{wenchong2026asurveyonuqmethodsfordl}} & \reviewerTwo{\cmark} & \reviewerTwo{\xmark} & \reviewerTwo{\smark} & \reviewerTwo{\cmark} & \reviewerTwo{\xmark} \\
        \edit{\cite{dong2025confcalibsurvey}} & \edit{\smark} & \edit{\xmark} & \edit{\cmark} & \edit{\cmark} & \edit{\smark} \\
        \rowcolor{yellow!50}
        \textbf{Ours} & \cmark & \cmark & \cmark & \cmark & \cmark \\
        \bottomrule
    \end{tabular}
    \label{tab:survey_comparison}
\end{table}

\subsection{Research Questions}

Since existing uncertainty measurement and calibration methods are mostly designed for classical DL, we ask how they should be redefined, adapted, or extended for LLMs. We address the following RQs:

\begin{enumerate}[label=RQ\arabic*.]
    \item Uncertainty in LLMs: What do we mean by uncertainty for LLMs? 
    \item Calibration in LLMs: What are existing calibration methods for LLMs?
    \item Evaluation: How well do the state-of-the-art calibration methods work in LLM settings?
    \item Research Gaps: How relevant are the existing works, and what are the emerging topics?
\end{enumerate}





Guided by these RQs, we review the gaps, recent methods, and opportunities in UQ and calibration for LLMs and provide an empirical evaluation on generation tasks.

\subsection{Review Methodology}
\label{sec:review_method}

Our review follows the PRISMA framework~\cite{page2021prisma}. After defining objectives and research questions, the identification stage follows the protocol of~\cite{kitchenham2004procedures}: we set inclusion and exclusion criteria, run keyword-based searches, and extract data from databases and registers. In screening, we assess titles and abstracts for relevance and define the data to extract from each study (quality indicators and eligibility thresholds). Finally, in the inclusion stage, we perform full-text analysis and make final decisions based on the RQs, maintaining lists of included and excluded studies. \revision{Having established our scope and methodology, we now address each research question in turn, beginning with the foundations of uncertainty in LLMs.}

\section{RQ1: Uncertainty in LLMs}
\label{sec:llm_uncertainty}

\subsection{Definition of Uncertainty}
\label{sec:uncertainty}

Definitions of uncertainty and confidence vary in the existing literature, whether it is probabilistically measured or verbally expressed by the LLMs, and some works use these terms interchangeably \cite{zhou2023navigating, geng2024surveyllmcal}. We define uncertainty as a rate of fluctuation and instability of a model's confidence over many trials and tasks. Likewise, we define the confidence of model predictions as a measure of how sure the model is about the correctness of its outcome. 
The nature of uncertainty is primarily categorized into two types
~\cite{walker2003defining, kendallgal2017whatuncertaintiesweneed}, (1) epistemic uncertainty and (2) aleatoric uncertainty.

Epistemic uncertainty is a type of uncertainty that happens because of lacking knowledge to make models more accurate. This type of uncertainty is also called model uncertainty \cite{chua2023tacklinguncert}. For modeling this uncertainty, latent penultimate representations can be fit into the probability distribution using the Gaussian mixture model (GMM) and distance metrics \cite{vazhentsev2023hybridue, wang2024subjectiveuqcal}. Model uncertainty can be reduced by introducing more data or architecture to the model \cite{kendallgal2017whatuncertaintiesweneed}.

Aleatoric uncertainty is a type of uncertainty caused by the variability or loss in the input data. This type of uncertainty is also called data uncertainty \cite{chua2023tacklinguncert}.  Existing methods use softmax output probabilities to apply probabilistic entropy to capture the aleatoric uncertainty of the model \cite{vazhentsev2023hybridue, wang2024subjectiveuqcal}. Incorrect, ambiguous inputs and corrupted data sources are major sources of aleatoric uncertainty\cite{hou2023decomposinguncertaintyinputc}, which is considered the root of hallucination \cite{aichberger2025sdlg}. While data uncertainty is irreducible, recent attempts use probabilistic contrastive learning to represent it \cite{kirchhof2023probabilistic}.

Note, based on the difference between train and test dataset distribution, distributional uncertainty may also exist and is connected to both aleatoric and epistemic uncertainties \cite{gawlikowski2023survey, baan2023uncertaintyfromtheorytoapplications}. Next, we describe classical DL methods to quantify uncertainty.


\subsection{Uncertainty Quantification for Classical DL}
\label{sec:methods_uq}

\begin{table*}[!t]
    \caption{A summary of methods for uncertainty quantification(UQ) for classical deep learning. 
    }
    \centering
    \label{tab:base_methods_outline}
    \begin{tabular}{
    @{}p{.24\textwidth}
    p{.30\textwidth}
    p{.30\textwidth}
    >{\centering\arraybackslash}p{.08\textwidth}@{}
    }
        \toprule
        \multicolumn{1}{c}{\textbf{Quantification Methods}}  & \multicolumn{1}{c}{\textbf{Advantages}}  & \multicolumn{1}{c}{\textbf{Limitations}} & \multirow{1}{*}{\textbf{LLM scalable}} \\ 
        \midrule

        BNN \cite{buntine1991bayesian, nix1994estimating, blundell2015bayesbybackprop, upadhyay2022bayescap} & Treat network weights as probability distributions for modeling the uncertainty. & Computational intensity of inference.  & \xmark \\
        Uncertainty-Aware training \cite{yan2023distortion} & Directly optimizes for uncertain samples. & Lack of works on designing effective loss functions and efficient integration. &  \smark \\
        Self-supervised learning \cite{kirchhof2023probabilistic} & Provides additional information about data distribution & Effectiveness depends on the chosen self-supervised task. & \smark \\
        
        \midrule

        Ensemble \cite{dietterich2000experimental} & Provides uncertainty through variability in predictions. & Can be computationally expensive. & \xmark  \\
        Deep Ensembles \cite{lakshminarayanan2017simple, dietterich2000experimental} & Better estimation with diversity. & Similar to regular ensembles, it can be computationally expensive. & \xmark \\
        MCD \cite{gal2016dropout} & Captures uncertainty with Bayesian approximation. & Computationally expensive due to multiple forward passes. & \smark \\

        $O(N)$ Energy \cite{phasetransition2025} & Measure the intensity of interactions among tokens within the generated sentence. & Performs better on large models (>7B). & \smark \\

        \bottomrule
    \end{tabular}
\end{table*}


UQ focuses on representing various types of uncertainty \cite{baan2023uncertaintyfromtheorytoapplications}. We compare UQ methods for DL in Table~\ref{tab:base_methods_outline}. Lower side of the table lists the sampling-based and the upper side of the table lists training and entropy-based UQ methods, the \emph{LLM scalable} column indicates whether a UQ approach is computationally feasible or scalable to apply on large model architectures. {\xmark} indicates inapplicable, {\smark} indicates can be applied to certain layers.

Bayesian neural networks (BNNs) \cite{buntine1991bayesian, blundell2015bayesbybackprop, upadhyay2022bayescap} and related approaches of variational inference \cite{Zhang2019CyclicalSG, Daxberger2021laplaceredux} provide efficiency, intuitiveness, and conscientious treatment of uncertainty by obtaining the posterior distribution over model weights using Laplace approximation (LA). In BNN, Bayesian layers can be added to the neural network during the large-scale training to model the probabilistic uncertainty alongside the result \cite{Tran2018BayesianLA, Cinquin2021PathologiesIP, Daxberger2021laplaceredux}. Such layers can also calibrate uncertainty using Kullback-Leibler (KL) regularization, joint learning, and fine-tuning by maintaining some of the pre-trained weights \cite{He2023PreservingPF}. While being popular, BNNs require access to the internal state and influence training procedures.

Uncertainty-aware training incorporates a penalty term to the loss function to mitigate overly confident predictions. The objective of that loss can be the consistency of difference in diverse uncertainty estimates \cite{yan2023distortion, krishnan2020improving}. In \cite{bengs2023secondorderloss}, they minimize second-order loss to quantify the epistemic uncertainty.

Self-supervised learning \cite{kirchhof2023probabilistic} leverages learning tasks to improve the model's understanding of data and uncertainty. They train the encoder by adding ambiguity to inputs for probabilistic embeddings of aleatoric uncertainty. The pseudo-uncertainty measure can be obtained using contrastive learning to estimate the original uncertainty by comparing augmented data points \cite{he2023clur}.

The ensemble \cite{dietterich2000experimental} method trains multiple models with different initializations and combines predictions. A line of works \cite{ovadia2019can, wen2020batchensemble} claims that deep ensembles\cite{lakshminarayanan2017simple, dietterich2000experimental}, which is the method to train an ensemble of diverse models to capture different aspects of the data distribution, can better quantify the uncertainty with a relatively small samples.  Dirichlet prior networks (DPN) use a single model for the ensemble of predictive distributions overweights and capture distributional uncertainty using the sampled categorical distributions \cite{malinin2018predictivepriornetworks, danruo2023uncertainty}. However, prior networks require explicit modification to the training process. In \cite{phasetransition2025}, they treat transformers as a physics O($N$) model and apply multiple phase transitions to obtain the energy of generated sequences for uncertainty.

While related to regular ensembles, Monte-Carlo dropout (MCD) \cite{gal2016dropout} is a simpler and computationally less intensive approximation of Bayesian inference but still requires open-box access to the model parameters or the last layer. MCD applies dropout during training and inference and runs multiple passes with dropout at test time. MCD can also be combined with Bayesian belief networks to capture different kinds of uncertainty \cite{loquercio2020generaluedl}. However, sampling-based methods provide over-confident and incorrect uncertainty measures if the model's predictions are also not well-calibrated \cite{gao2024spuq, zhang2024luq}.

\subsection{Applicability in LLMs}

\reviewerTwo{In principle, the surveyed methods target different uncertainty components, but in practice they only partially decouple aleatoric and epistemic uncertainty in our benchmarks. Methods based on sampling variance, ensembles, or debate-style disagreement are intended to capture epistemic effects, whereas entropy- and confidence-based signals are often interpreted as reflecting aleatoric noise. However, LLM generation introduces coupled sources of variation (including prompt underspecification, decoding stochasticity, and semantic ambiguity) that simultaneously affect both components. As a result, the observed uncertainty signals are often conflated, especially in open-ended generation tasks where multiple valid answers exist. Therefore, we treat the aleatoric-epistemic distinction as an approximate analytical lens rather than a strict empirical separation, and we interpret benchmark results accordingly. For instance, uncertainty on TruthfulQA is predominantly epistemic, whereas on TriviaQA it is largely aleatoric, arising from ambiguity in the questions and answers~\cite{xia2025asurveyofuellms}.}

While the methods described so far are generally sufficient to quantify uncertainty in classical DL, efforts to adapt these methods to LLMs remain limited \cite{hou2023decomposinguncertaintyinputc, gao2024spuq}. Figure~\ref{fig:type-of-uncert-large} summarizes methods for quantifying uncertainty in LLM settings. In Table~\ref{tab:uq_llm_methods}, we cover the commonalities and differences of existing methods to provide a decision framework for practical needs. We also classify existing methods by the LLMs and datasets used in Table~\ref{tab:review_methods_outline}.

\begin{table*}[!t]
    \centering
    \caption{\edit{A summary of commonly used UQ methods for LLMs. Key Terms: Input Dependence of method, Model Access, Theoretical Approach, ``Generative": applicability to generation tasks.}}
    \label{tab:uq_llm_methods}
    \renewcommand{\arraystretch}{1.2}
    \begin{tabular}{
        @{}p{0.2\textwidth}
        p{0.1\textwidth}
        p{0.15\textwidth}
        p{0.1\textwidth}
        >{\centering\arraybackslash}p{0.1\textwidth}@{}
    }
    \toprule
    \textbf{Name \& Ref.} & \textbf{Model Access} & \textbf{Input Dependence} & \textbf{Theoretical Approach} & \textbf{Generative} \\
    \midrule
    SPUQ \cite{gao2024spuq} & Open-box & Logits & Ensemble & \cmark \\
    \midrule
    SE \cite{kuhn2023semanticuncertainty} & Open-box & Logits & Ensemble & \cmark \\
    \midrule
    CP \cite{kumar2023conformalllmmcqa} & Open-box & Logits & Exchangeability & \xmark \\
    \midrule
    DepUnc \cite{yoffe2024debunc} & Open-box & Logits & Ensemble & \xmark \\
    \midrule
    Active-Prompt \cite{diao2023activeprompt} & Open-box & Prompt & Bayesian & \smark \\
    \midrule
    LoFreeCP \cite{su2024lofreecp} & Closed-box & Text & Exchangeability & \smark \\
    \midrule
    KLE \cite{widmann2021multiclasscalkernel} & Closed-box & Text & Ensemble & \cmark \\
    \midrule
    SelfCheckGPT \cite{manakul2023selfcheckgpt} & Closed-box & Prompt & Ensemble & \cmark \\ 
    \midrule
    Deductive reasoning \cite{ling2024deductiveverification} & Closed-box & Prompt & Ensemble & \cmark \\
    \midrule
    SelfDetection \cite{zhao2023selfdetectionanswerclustering} & Closed-box & Prompt & Ensemble & \cmark \\
    \midrule
    Epistemic markers \cite{zhou2023navigating} & Closed-box & Text & Ensemble & \cmark \\
    \bottomrule
    \end{tabular}
\end{table*}

\reviewerOne{Across these methods, access to model internals ranges from fully closed-box approaches, where only outputs or textual interfaces are used, to open-box techniques exploiting logit information directly. SPUQ \cite{gao2024spuq} and SE \cite{kuhn2023semanticuncertainty} both require open-box access to model logits and internals, placing them in this category. Debates with Model Uncertainty Metrics likewise need direct access to logits for pairwise scoring \cite{yoffe2024debunc}. Conformal Prediction is also open-box, leveraging logits to calibrate confidence intervals \cite{kumar2023conformalllmmcqa}. LofreeCP, by contrast, avoids logits entirely and operates via text sampling, making it closed-box \cite{su2024lofreecp}. }

Next, we describe the key aspects and practices for LLMs revealed by our review.

\begin{table*}[!bt]
    \centering
    \caption{
    A summary of commonly used large models with their task and applied uncertainty methods. Abbreviations: UQ - Uncertainty Quantification, OSR - Open-Set Recognition, GP - Gaussian Process. 
    }
    \label{tab:review_methods_outline}
    \begin{tabular}{
    >{\arraybackslash}p{.2\textwidth}
    >{\centering\arraybackslash}p{.2\textwidth}
    >{\centering\arraybackslash}p{.1\textwidth}
    >{\centering\arraybackslash}p{.1\textwidth}
    >{\centering\arraybackslash}p{.2\textwidth}}
        \toprule
        \textbf{Methods \& Ref}  &
        \multicolumn{1}{c}{\textbf{\# Parameters}} & 
        \multicolumn{1}{c}{\textbf{Large model}} & 
        \multicolumn{1}{c}{\textbf{Dataset(s)}} &
        \multicolumn{1}{c}{\textbf{Task/Domain}} \\ 
        \midrule
        Semantic entropy \cite{kuhn2023semanticuncertainty}, ICL decomposition \cite{ling2024uqicl} & 2.7B/6.7B/13B/30B & OPT \cite{zhang2022opt} & TriviaQA, CoQA, AG\_News, SST2 & UQ \\
        \midrule
        DebUnc \cite{yoffe2024debunc}, Ensemble priors \cite{balabanov2024loraensemble} & 7B/123B & Mistral \cite{jiang2023mistral} & CoQA, MMLU, TruthfulQA, GSM8k, Arithmetic & UQ \\
        \midrule
        Plex: BatchEnsemble + GP \cite{tran2022plex} & 250B/1B & T5 \cite{raffel2020tfivettt} & MNLI, WikipediaTalk, NaLUE & UQ, Calibration, OSR \\
        \midrule
        BatchEnsemble \cite{wen2020batchensemble} & 65M/213M & Transformer & WMT14 & UQ \\
        \midrule
        Verbalized uncertainty \cite{lin2022teachingmodelstoexpress}, SelfCheckGPT\cite{manakul2023selfcheckgpt} &   125M/6.7B/175B & GPT-3 \cite{brown2020languagegpt3} & WikiBio, MMLU & UQ, Calibration \\
        \midrule
        Clarification ensemble \cite{hou2023decomposinguncertaintyinputc}, SelfCheck\cite{miao2023selfcheck}, Self-Refine\cite{madaan2024selfrefine} & 1.8T & GPT-4 \cite{achiam2023gpt4} & GSM8K, MathQA, MATH, NQ & UQ, Self-evaluation \\
        \midrule
        Deg, Ecc, EigV \cite{lin2023generatingwithconfblackbox}& 7B/13B/70B & Llama 2 \cite{touvron2023llama2} & CoQA, TriviaQA & UQ \\
        \bottomrule
    \end{tabular}
\end{table*}

\noindent{\bf Closed-box access.} \reviewerOne{Recently, many tech companies have restricted model access by securing their models behind the Application Programming Interfaces (APIs), making it more challenging to apply methods that require direct access \cite{huang2023lookbeforeyouleap, ulmer2024apricot}. The methods that calibrate these models are called closed-box methods. However, built-in (open-box) methods can be more informative than post-hoc methods that apply only to the last or penultimate layer, though which approach is better is still an open question \cite{wang2020transcal, xiong2023canllmexpress}. Integrating open-box methods into LLMs can be complex and computationally demanding \cite{lin2023generatingwithconfblackbox}. In contrast, closed-box methods are simpler to integrate and optimize for LLMs, especially as recent research increasingly focuses on verbalizing uncertainty using prompt engineering \cite{lin2022teachingmodelstoexpress, kadavath2022llmsknowmostly,wang2022selfconsistency, weng2022selfverification}. Several methods rely directly on logits, while KLE \cite{Nikitin2024KernelLE} extracts uncertainty from the diversity of generated prose, categorizing it as text-based \cite{Nikitin2024KernelLE}.}

\reviewerTwo{Conformal prediction (CP) is a model-agnostic UQ method in which the prediction set containing the correct answer is treated as a sample, and the set size is used as an uncertainty proxy; larger sets indicate higher uncertainty\cite{kumar2023conformalllmmcqa, quach2023conformallangmod}. However, CP methods for MCQA \cite{ye2024benchmarkingllmuq} have limited adaptability to generative tasks and still require access to model logits.}

\noindent{\bf Generative adaptability.} \reviewerOne{While most conformal and prompt-based classifiers focus on classification, SPUQ \cite{gao2024spuq}, SE \cite{kuhn2023semanticuncertainty}, KLE \cite{Nikitin2024KernelLE}, Deductive Reasoning \cite{{ling2024deductiveverification}}, SelfDetection \cite{zhao2023selfdetectionanswerclustering}, and Epistemic Markers \cite{zhou2023navigating} all extend to generative applications, providing uncertainty-aware text outputs. LofreeCP \cite{su2024lofreecp} applies primarily to classification but can generate text under strict exchangeability conditions, albeit with limited fidelity. On the other hand, exchangeability underpins CP and LofreeCP by assuming identically distributed data for coverage guarantees and Bayesian uncertainty estimation drives Active-Prompt \cite{diao2023activeprompt}. The remaining techniques harness ensemble concepts, running multiple perturbations, samples, debates, or reasoning chains, to approximate epistemic variance.}

\noindent{\bf Expressing uncertainty.} Prompting LLMs to verbalize uncertainty can efficiently elicit confidence signals and reduce user mistrust \cite{lin2022teachingmodelstoexpress, xu2024sayself}. In \cite{kim2024uncertain1st}, the authors compare first-person, generic, and non-LLM uncertainty expressions in user questionnaires and show that explicit uncertainty statements can reduce over-trust, although they may also produce overly generic responses. Uncertainty expression and chain-of-thought (CoT) confidence are also connected to explainable AI (XAI) \cite{becker2024cyclesofthought}. This is especially important in high-stakes domains such as healthcare, where practitioners must understand model reasoning to assess reliability \cite{chua2023tacklinguncert, wang2023uncertretinal}. \edit{An open question is how temperature settings influence verbalized uncertainty. Recent work emphasizes the importance of temperature selection \cite{liu2025metafaith}, while another open issue is whether models can reliably detect and report their own uncertainty \cite{joo2025blackboxhallucinationmetric}.}

\noindent{\bf Extracting semantics of sequences.} \edit{One crucial challenge in UQ research is the lack of a comprehensive evaluation metric that captures different types of uncertainty \cite{gawlikowski2023survey}. Because lexical similarity (LexSim) \cite{fomicheva2020lexsim} has limited ability to capture semantically equivalent outputs with diverse wording, semantic uncertainty is a necessary aspect of generation, and quantifying epistemic semantic uncertainty remains an open challenge \cite{kuhn2023semanticuncertainty, aichberger2025sdlg}. Existing works state the importance of the semantic entropy of the sentences for accurate UQ using natural language inference (NLI) since token entropy conflates different uncertainty kinds \cite{kuhn2023semanticuncertainty}. However, NLI was found to have limitations in detecting precise semantic dissimilarity between responses \cite{Qiu2024SemanticDU}. Recent work also samples LLM outputs to measure uncertainty, which increases generation cost \cite{Nikitin2024KernelLE, wang2022selfconsistency}. While existing methods use clustering of similar-meaning tokens to capture semantic uncertainty, more specific and complex semantic structures should be explored \cite{Nikitin2024KernelLE}.}

\noindent{\bf Quality assurance.} Empirical evaluation metrics can use quantified uncertainty to compare the performance of different AI models. Confidence scores can be used when returning an empty score array for abstaining from answering. Recent work proposes transferable uncertainty metrics for pre-trained models to distinguish certain from uncertain unseen examples using representation learning \cite{kirchhof2023url}. The limited ability of recent representation-learning methods to detect unseen samples also remains a challenge for anomaly detection \cite{hendrycks2021unsolved}.

\section{RQ2: Calibration in LLMs}
\label{sec:llm_calibration}
\subsection{Model Calibration}
\label{sec:modelcalibration}
Calibration, initially developed for forecasting~\cite{murphy1977reliabilityofforecasts, dawid1982wellcalibratedbayesian, degroot1983comparisonforecaster}, is a technique to assess and improve the quality of uncertainty estimates by ensuring that models provide more transparent and reliable predictive probabilities~\cite{hendrycks2021unsolved}. Since modern deep learning models frequently rely on softmax output probabilities to estimate uncertainty, improper calibration can lead to misleading confidence scores, which is particularly problematic in high-stakes decision-making tasks.
To achieve proper calibration, a model's predicted confidence level (e.g., 
the softmax score $\alpha$) should align with its actual accuracy~\cite{kumar2019verified}. 
In other words, among all instances where the model assigns a confidence of $\alpha$, the proportion of correct predictions should approximate $\alpha$. 
This alignment can be evaluated through techniques such as reliability diagrams, 
expected calibration error (ECE), and temperature scaling (TS) \cite{guo2017calibration}.
A classification model can be considered calibrated if:
\begin{equation}
    \Pr[f(X) = Y \mid \hat{p}_{f}(X) = \alpha] = \alpha,
    \label{eq:pref_cal}
\end{equation}
where $f(X)$ function that maps the feature space $X$ and label space $Y$, while $\hat{p}_{f}$ 
represents the predicted confidence $\alpha$ of $f$. 
To assess calibration, we can partition this confidence into intervals and construct 
reliability diagrams~\cite{niculescu2005predictingreldiag}, which visualize the 
relationship between predicted confidence and actual accuracy within each interval. While achieving perfect accuracy in real-world scenarios is often infeasible~\cite{si2022promptinggpt3reliable}, perfect calibration of uncertainty remains an attainable goal~\cite{nguyen2015posteriorcal, laves2019well}.

Post-hoc or post-training methods, such as TS \cite{guo2017calibration}, Platt scaling \cite{platt1999prbbplattscaling}, histogram binning \cite{zadrozny2001histogrambinning}, and training with label smoothing \cite{muller2019whendoeslabelsmoothinghelp}, are classical methods of calibration. Mixup training with learning the complexity of samples was proposed to calibrate the pre-trained language models \cite{park2022mixupcalibration}. We note that open-box methods of calibration can be complex and difficult to apply to LLMs, but they can be more informative of inner states than post-hoc methods~\cite{schuster2022confident}. 

\subsection{Calibration Methods}
\label{sec:advances}

\begin{figure*}[!bt]
    \centering
    \footnotesize
    \begin{forest}
        for tree={
            forked edges,
            grow'=0,
            rounded corners,
            node options={align=center},
            text width=3.0cm,
            s sep=4pt,
            l sep=8pt,
            calign=child edge, 
            calign child=(n_children()+1)/2,
        }
        [Methods of Uncertainty for LLMs, fill=gray!45, parent anchor=south, parent
            [Quantification, for tree={quantchild}
                [Open-Box, quantgrandchild
                    [Sampling-based, quantgrandchild
                        [MCD \cite{gal2016dropout, zablotskaia2023uncertaintybenchmark}; Batch ensemble \cite{wen2020batchensemble}; Model ensemble \cite{sun2022quantifyingmodelensemble}; R-U-SURE \cite{johnson2023rusure}; Clarification ensemble \cite{hou2023decomposinguncertaintyinputc}; SPUQ \cite{gao2024spuq};, style = referenceblock_l2]
                    ]
                    [Entropy-based \& UA, quantgrandchild
                        [SD \cite{Qiu2024SemanticDU}; UABS \cite{xiao2021hallucinationanduncertainty}; PE \cite{kadavath2022llmsknowmostly}; LNPE \cite{malinin2020uncertaintylengthnormalized}; SE \cite{kuhn2023semanticuncertainty, farquhar2024detecting}; SDLG \cite{aichberger2025sdlg}; Predictive variance \cite{xiao2019quantifyingunlp}; UA Self-correction \cite{yang2023improvingiclselfcorr}; ICL decomposition \cite{ling2024uqicl}; UAIT \cite{liu2024uait}; Uncertainty indices \cite{audrino2024uqindices}; SAR \cite{duan2024shiftingattentiontorelevance}; Claim-conditioned \cite{fadeeva2024ccp}; HUQ \cite{vazhentsev2023hybridue}; \reviewerTwo{WSE \cite{wang2025wordsequenceentropy}};, style = referenceblock_l2]
                    ]
                    [Conformal prediction, quantgrandchild
                        [\reviewerTwo{Non-exchangeable CP \cite{ulmer2024nonexchangeablecp}; DDCRP \cite{kaur2024ddcrp};}Exchangeability\cite{shafer2008tutorialconfpred, kumar2023conformalllmmcqa}; KnowNo \cite{ren2023robotsknowno}; Rejection/stopping rule \cite{quach2023conformallangmod};,
                        style = referenceblock_l2]
                    ]
                    [Language agents, quantgrandchild
                        [Reflexion \cite{shinn2024reflexion}; UALA \cite{han2024towardsuala}; CRITIC \cite{gou2023critic}; DebUnc \cite{yoffe2024debunc}; ReConcile \cite{chen2023reconcile}; MAD \cite{liang2023mid};,
                        style = referenceblock_l2]
                    ]
                    [Human uncertainty, quantgrandchild
                        [Fluency/Adecuacy \cite{miao2021preventoverconf}; Perceptual similarity \cite{peterson2019humancategoricalsimilarity};DistCE \cite{baan2022stopmeasurnigcaldisce}; LID \cite{Yin2024LID}; Active-Prompt\cite{diao2023activeprompt};, 
                        style = referenceblock_l2]
                    ]
                ]
                [Closed-Box, quantgrandchild
                    [Sampling-based,  quantgrandchild      
                        [\reviewerTwo{LoFreeCP\cite{su2024lofreecp}; ConU \cite{wang2024conu};}Prompt Ensemble \cite{si2022promptinggpt3reliable}; Unc-TTP \cite{huang2024uncttp}; LUQ-Ensemble \cite{zhang2024luq}; Probing/CoT uncertainty \cite{tanneru2024quantifyinguncertaintyllm}; LexSim \cite{fomicheva2020lexsim};, style = referenceblock_l2]
                    ]
                    [Semantic uncertainty, quantgrandchild
                        [KLE \cite{Nikitin2024KernelLE}; Eccentricity/Degree Matrix/EigV \cite{lin2023generatingwithconfblackbox};  Semantic embeddings \cite{grewal2024improvinguqllmsemanticembedding}; Rowen \cite{ding2024rowen};, style = referenceblock_l2]
                    ]
                    [Self-evaluation, quantgrandchild
                        [C-AUC \cite{ren2023selfeval}; Self-Verification \cite{weng2022selfverification}; SelfCheck \cite{miao2023selfcheck}; SelfCheckNLI \cite{manakul2023selfcheckgpt}; Self-Refine \cite{madaan2024selfrefine}; Deductive Reasoning \cite{ling2024deductiveverification}; Ref. Overlap \cite{Agrawal2023overlapestimation},
                        style = referenceblock_l2]
                    ]
                    [Self-detection, quantgrandchild
                        [Answer clustering \cite{zhao2023selfdetectionanswerclustering}; BSDetector \cite{chen2024quantifyingtrustworthiness}; Joint confidence \cite{li2024thinktwicebeforetrusting}; Self-reflection \cite{ji2023selfreflection};,
                        style = referenceblock_l2]
                    ]
                    [Verbalized uncertainty, quantgrandchild
                        [Expression \cite{lin2022teachingmodelstoexpress}; Epistemic markers \cite{zhou2023navigating}; Stable explanations \cite{becker2024cyclesofthought}; Demonstration uncertainty \cite{ling2023improvingdemonstuncert}; Red teaming \cite{ganguli2022redteamingllms}; Prompting \cite{shrivastava2023llamasknowwhatgptsdontshow}; Convex Hull \cite{catak2024uqoconvexhullarea};,
                        style = referenceblock_l2]
                    ]
                ]
            ]
            [Calibration, for tree={calibchild}
                [Bias reduction, calibgrandchild
                    [In-Context Learning, calibgrandchild
                        [Batch calibration \cite{zhou2023batchcalibration}; Domain-context \cite{roelofs2022domancontextcalib}; Prototypical  \cite{han2022prototypicalcalib}; Answer-level \cite{kumar2022answerlevecalib}; Contextual \cite{zhao2021contextualcalib}; CLAPS\cite{claps}; $\text{PMI}_{DC}$ \cite{holtzman2021dcpmi};,
                        style=referenceblock_l2]
                    ]
                    [Multi-calibration, calibgrandchild
                    [IGLB\cite{detommaso2024multicalibrationiglb}; IGHB \cite{hebert2018multicalibrationigbh}; Low-degree \cite{gopalan2022lowdegreemulticalib}; Loss Min. \cite{Blasiok2023LossMY}; Decision cal. \cite{zhao2021calibratingpredtodecisions};,
                        style=referenceblock_l2]
                    ]
                    [Selection bias, calibgrandchild
                        [PriDe \cite{zheng2023llmselbiaspride}; , 
                        style=referenceblock_l2]
                    ]
                ]
                [Correctness alignment, calibgrandchild
                    [Post-hoc, calibgrandchild
                        [Parametric, calibgrandchild
                            [Temperature scaling\cite{guo2017calibration}; Platt scaling\cite{platt1999prbbplattscaling}; Atypicality-aware \cite{yuksekgonul2024atypicalitycalib}; Joint pipeline \cite{dhuliawala2022jointcalibration}; PLEX \cite{tran2022plex};,
                            style=referenceblock_l3]
                        ]
                        [Non-parametric, calibgrandchild
                            [HB\cite{zadrozny2001histogrambinning}; Isotonic Regression\cite{zadrozny2002isotonicregression}; Localized \cite{luo2022lce};  Empirical RCE \cite{huang2024ulmrankcalib};,
                            style=referenceblock_l3]
                        ]
                        [Hybrid, calibgrandchild
                            [Scale-binning \cite{kumar2019verified},
                            style=referenceblock_l3]
                        ]
                    ]
                    [Fine-tuning \& Internal-state, calibgrandchild
                        [SLiC \cite{zhao2023calibratingsli}; SLiC-HF \cite{zhao2023slichf}; LoRA Ensemble \cite{balabanov2024loraensemble};
                        LS \cite{muller2019whendoeslabelsmoothinghelp, lu2022learningconftransformernmt, lee2022adaptivels}; R-tuning \cite{zhang2024rtuning}; LACIE \cite{stengel2024lacie}; LitCab \cite{liu2024litcab}; Decision-based RL \cite{band2024linguistic}; Laplace-LoRA \cite{Yang2023Laplacelora}; Sayself \cite{xu2024sayself}; Cal. tuning\cite{kapoor2024calibrationtuning}; ActCab \cite{liu2024activationconfcalibcodec}; Early exiting \cite{schuster2022confident}; CaliNet \cite{dong2022calinet};,
                        style = referenceblock_l2]
                    ]
                    [Closed-Box, calibgrandchild
                        [Linguistic, calibgrandchild
                            [Calibrator-controlled \cite{mielke2022reducingoverconfidence, lin2022teachingmodelstoexpress}; Human-LM \cite{zhou2024relyingonunreliable}; FaR \cite{zhao2024far}; Fidelity \cite{zhang2024calibratingbyelicitingfidelity}; Atomic \cite{zhang2024atomiccalib}; Calibration/Discrimination gap \cite{steyvers2024calibrationgap, steyvers2025llmexplanationconfdiscgap};, style=referenceblock_l3]
                        ]
                        [Self-consistency, calibgrandchild
                            [ SC \cite{wang2022selfconsistency, huang2024calibratinglongform}; Textual consistency \cite{schuster2022confident}; USC \cite{chen2023universalselfconsistency}; Descriptor consistency \cite{dai2023exploringmllmood}; SampleAvgDev \cite{rivera2024sampleavgdev}; Cross-model \cite{xue2025crossmodelconsistency}; \edit{CAI \cite{chen2025evaluatingcai}};, style=referenceblock_l3]
                        ]
                        [Ensemble, calibgrandchild 
                            [CAPE \cite{jiang2023cape}; Self-ensemble-SICL \cite{xiong2023canllmexpress, li2023selfensemblesicl};,
                            style = referenceblock_l3]
                        ]
                        [Auxiliary model, calibgrandchild
                            [Surrogate mixture \cite{shrivastava2023llamasknowwhatgptsdontshow}; Apricot\cite{ulmer2024apricot}; Majority examiner \cite{cohen2023lmvslm}; SAPLMA \cite{azaria2023saplma};,
                            style = referenceblock_l3]
                        ]
                    ]
                ]
            ]
        ]
    \end{forest}
    \caption{A classification of UQ and calibration methods for LLMs. Abbr.: Randomized Utility-driven Synthesis of Uncertain REgions (R-U-SURE), sampling with perturbation for UQ (SPUQ), Uncertainty-Aware Beam Search (UABS), Instruction Tuning (IT), Uncertainty-Aware Language Agent (UALA), Hybrid uncertainty quantification (HUQ), Self-Correcting with Tool-Interactive Critiquing (CRITIC), Human Distribution Calibration Error (DistCE), Length-normalized predictive entropy (LNPE), Predictive entropy (PE), Clustering and Pruning for Efficient Black-box Prompt Search (CLAPS), Iterative Grouped Histogram Binning (IGHB), Sequence Likelihoood Calibration (SLiC), Refusal-Aware Instruction Tuning (R-tuning), Listener-Aware Calibration for Implicit and Explicit confidence (LACIE), Statement Accuracy Prediction, based on Language Model Activations (SAPLMA), Self-consistency (SC), Universal self-consistency (USC), Label-smoothing (LS), Fact-and-Reflection (FaR), Uncertainty-aware Instruction Tuning (UAIT), Uncertainty Tripartite Testing Paradigm (Unc-TTP), Semantically Diverse Language Generation (SDLG), Kernel Language Entropy (KLE), Semantic Entropy (SE), Semantic Density (SD), Local Intrinsic Dimensions (LID), Multi-agent debate (MAD), \reviewerTwo{Word-Sequence Entropy (WSE), Distance Dependent Chinese Restaurant Process (DDCRP), Consistent-and-Inconsistent (CAI) Ratio}.}
    \label{fig:type-of-uncert-large}
    \vspace{-2.1ex}
\end{figure*}
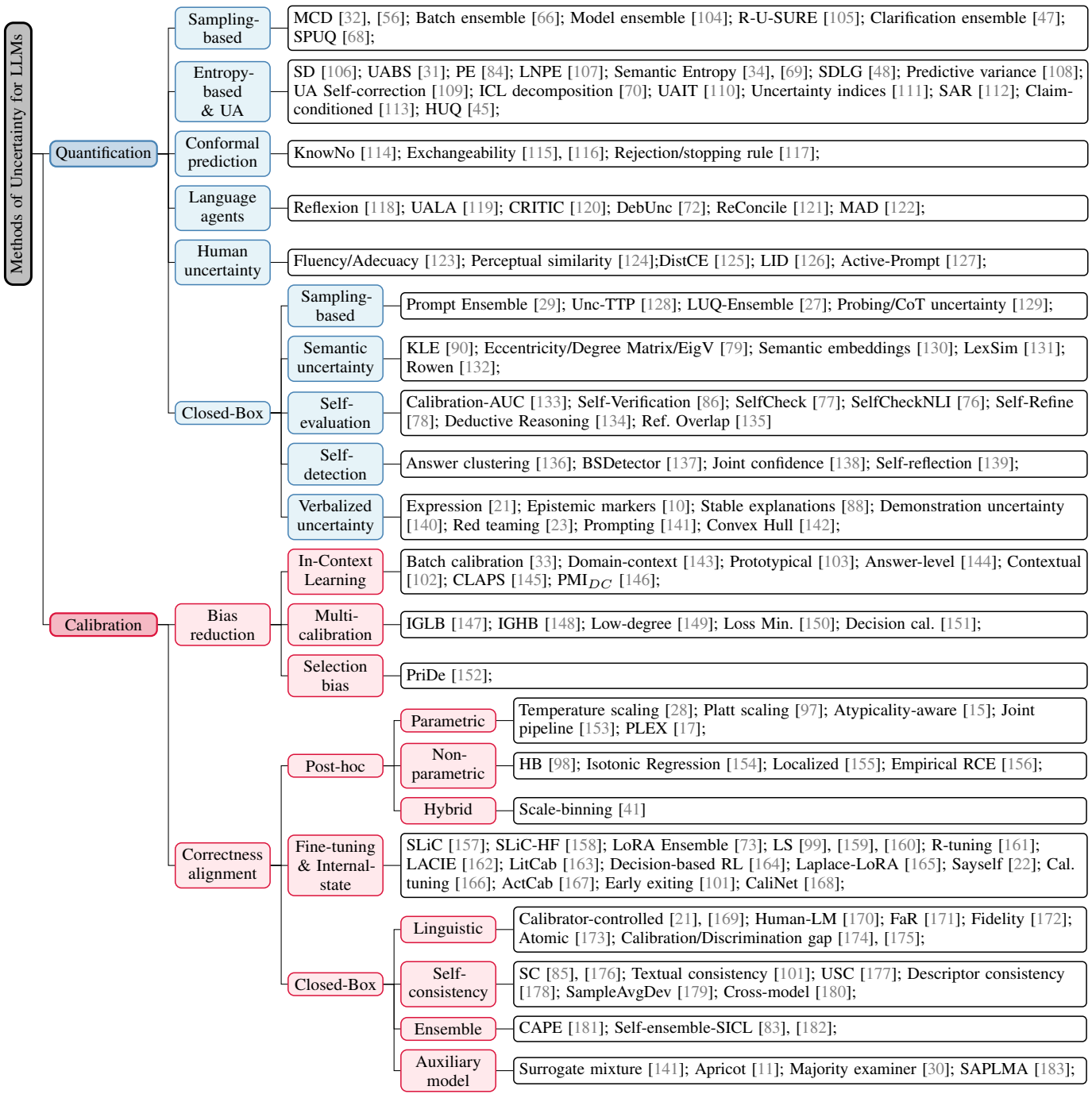

We explain the metrics and ways of calibration of uncertainty in LLMs \cite{fisch2022uncertaintyfornlptutorial}. Figure~\ref{fig:type-of-uncert-large} summarizes various methods to calibrate and quantify uncertainty for the setting of LLMs.

\subsection{Evaluation Metrics}
\label{sec:eval_metrics}

First, we provide a quick overview of the different evaluation metrics used in this work that can be used for language tasks. Appendix~A-C provides details and the mathematical formulation of evaluation metrics that we primarily used for experiments.


\noindent{\bf Expected calibration error (ECE).} ECE shows the calibratedness of the model inference, which is the average of the difference between the accuracy and confidence score of batched probabilistic outcomes specified within a pre-defined number of bins \cite{guo2017calibration}. The intuition behind this metric is that there should be an alignment between the accuracy and model confidence, which can be defined as the probability of predicted token \cite{zablotskaia2023uncertaintybenchmark}.

\noindent{\bf Uncertainty calibration error (UCE).} Since confidence scores in ECE can have a limited representation of calibration \cite{kuleshov2015calibrated}, the UCE metric considers the average bin-wise difference of model error and its uncertainty \cite{laves2019well}. UQ methods that we discussed previously can be used to obtain the uncertainty estimates \cite{krishnan2020improving}. The intuition behind UCE is that the more error our model has, the more we expect it to be uncertain about its prediction. 

\noindent{\bf Rank-Calibration Error (RCE).} RCE \cite{huang2024ulmrankcalib} is the average bin-wise difference from the given uncertainty value to the ideal expected accuracy of the regression function applied to the uncertainty, where lower uncertainty corresponds to higher accuracy. However, there is a lack of statistical evaluation of the metric.

\noindent{\bf Class-wise ECE/UCE.} According to \cite{kull2019beyond}, ECE revealed fundamental limitations because the binning method can show over-confidence and under-confidence between class categories. In addition, larger classes can influence calibration more than smaller classes. Hence, \cite{laves2020calibration} suggested the class-wise extension for the UCE metric, which is the average of UCE result per class and it is similar to the class-wise ECE (cECE) \cite{kull2019beyond}.

\noindent{\bf Weighted ECE.} Weighted ECE can be used for language tasks, such as neural machine translation (NMT), considering attention uncertainty in which whole probabilistic distribution is considered instead of top-scoring token for confidence \cite{kumar2019nmtcalibration}. Structured ECE is the extension of weighted ECE, where the expected BLEU score is used as confidence.

\noindent{\bf Adaptive Calibration Error (ACE).} ECE and UCE require a pre-defined number of bins and can not adapt the different-sized bins for accurate calibration assessment. ACE averages over bias-variance trade-off ranges of bin-wise data points \cite{nixon2019measuringcalibrationindl}. However, ranges can be larger if the confidence values are further away from each other.

\noindent{\bf SmoothECE.} SmoothECE is a recent extension of the ECE metric that addresses the binning problem in ECE by employing Gaussian kernel smoothing \cite{blasiok2023smoothece}. This approach replaces traditional binning with a smoothed ground-truth calibrator. However, it is a generic metric and does not capture different probability structures.

\noindent{\bf Sequence-level ECE.} Another extension of ECE is Sequence-level ECE, which evaluates the confidence of an entire sentence by calculating the product of the probabilities of each token within that sequence  \cite{zablotskaia2023uncertaintybenchmark}, similar in a sense to sentence BLEU \cite{ott2018analyzinguqnmt}.

\noindent{\bf Brier Score (BS).} Brier Score is defined as a squared error between prediction probability and ground-truth accuracy and can be used for quantifying reliability \cite{brier1950scoreverification, brocker2009reliabilitysufficiency, yin2023llmknowwhattheydont}. Because the Brier score only combines prediction accuracy and confidence, it can fail to detect the likelihoods of rare samples\cite{wang2020transcal}.

\noindent{\bf Bilingual evaluation understudy (BLEU) score.} BLEU score \cite{papineni2002bleu} can be used as a confidence score for language tasks, such as NMT \cite{miao2021preventoverconf}. Although this metric was originally developed for translation, it can be used to evaluate text generated by natural language processing models for different tasks. A better-calibrated model can reveal an improved BLEU score \cite{kumar2019nmtcalibration}.

\noindent{\bf Area Under the Receiver Operator Curve (AUROC).} AUROC metric is usually used for anomaly or OOD detection \cite{hendrycks2016aurocbaseline}. This metric can help to identify samples that the language model does not know in QA tasks when introduced with a different dataset \cite{becker2024cyclesofthought}. However, AUROC might be indifferent to the calibration of language models \cite{yin2023llmknowwhattheydont}.

\noindent{\bf Area Under the Risk-Coverage Curve (AURC).} \reviewerTwo{
AURC quantifies the tradeoff between model accuracy and the proportion of predictions retained based on confidence thresholds \cite{becker2024cyclesofthought}. As the coverage increases from the most confident predictions to the entire dataset, AURC summarizes how accuracy changes. 
When all predictions are included (full coverage), AURC reduces to the model's average accuracy.
}

\noindent{\bf Area Under Accuracy-Rejection Curve (AUARC).} \reviewerTwo{AUARC is a metric computed as average accuracy when we reject predictions based on the uncertainty of model prediction \cite{lin2023generatingwithconfblackbox}.}

\noindent{\bf Area Under Precision-Recall Curve (AUPRC).} \reviewerTwo{This metric quantifies the tradeoff between precision and recall retained based on different classification thresholds \cite{xiong2023canllmexpress}.}

\subsection{Types of Calibration}

\begin{table*}[!t]
    \centering
    \caption{A summary of commonly used calibration methods for LLMs. Key Terms: ``Cost-effective'': indicates the application cost, ``Generative": applicability to generation tasks. A possible failure case for each method in the given setting is marked as \xmark. $B$ is interval of bins $[\frac{1}{B}]$, $\alpha$ is confidence, $f(x)$: model output for input $x$, $p$ probability of prediction, $C$ is context, $g$ is group/category from Dataset, $\delta$ is hyperparameter for rank margin.}
    \label{tab:cal_llm_methods}
    \renewcommand{\arraystretch}{1.2}
    \begin{tabular}{
        @{}p{0.1\textwidth}
        p{0.09\textwidth}
        p{0.08\textwidth}
        p{0.4\textwidth}
        >{\centering}p{0.07\textwidth}
        >{\centering}p{0.04\textwidth}
        >{\centering\arraybackslash}p{0.08\textwidth}@{}
    }
    \toprule
    \textbf{Name \& Ref.} & \textbf{\reviewerTwo{Objective}} & \textbf{Approach} & \textbf{Formula} & \textbf{Complexity} & \textbf{Cost-effective} & \textbf{\reviewerTwo{Generative}} \\
    \midrule
    Batch Cal.\cite{zhou2023batchcalibration} & Bias reduction & ICL & $\hat{p} = \textbf{W}p + \textbf{b}, \textbf{W} = \mathds{I}, \textbf{b} = - \mathds{E}_x[p(y|x,C)]$ & Low & \cmark & \xmark \\
    \midrule
    IGLB \cite{detommaso2024multicalibrationiglb} & Bias reduction & Multi-calibration & $ f_{\text{round}+1}(x) = \text{TS}[f_\text{round}](x), x \in \{ f(x) = \alpha, g(x) = 1 \}$ & Medium & \smark & \xmark \\
    \midrule
    PriDe \cite{zheng2023llmselbiaspride} & Bias reduction & Selection bias & $ p(y_{\text{option}(i)} | q, x)=\frac{p(y_{\text{id}(i)}|q, x)}{p_\text{prior}(y_{\text{id}(i)})}, i \in \{1, 2, ..., n\} $ & Medium (with Cyclic permutation) & \smark & \xmark \\
    \midrule
    SLiC-HF \cite{zhao2023slichf} & Correctness alignment & Fine-tuning & $L^{\text{cal}}(w) = max(0, \delta - \text{log} P_w (y^+|x) + \text{log} P_w (y^- | x)) $ & High & \smark & \cmark \\
    \midrule
    Calibrator-controlled \cite{mielke2022reducingoverconfidence} & Correctness alignment & Linguistic & $p(y | x) = p(y | x + <\text{confidence token}> )$ & Low & \cmark & \cmark \\
    \midrule
    SC \cite{wang2022selfconsistency} & Correctness alignment & Self-consistency & $f(x) = argmax_x \frac{1}{k} \sum_{i \in k} \mathds{I} (y_i = f(x)) $ & Medium (many inference times) & \xmark & \cmark \\
    \midrule
    CAPE~\cite{jiang2023cape} & Correctness alignment & Ensemble & $ \hat{p}(y|x) = \sum_{i \in k} \frac{p(y|x_i)}{k} $ & Low & \smark & \cmark \\ 
    \midrule
    TS~\cite{guo2017calibration} & Correctness alignment & Parametric & $p(y | x) = \frac{\text{exp}(\frac{\text{logit}f(x)}{\textbf{T}})}{\sum_j \text{exp}(\frac{\text{logit}f_j(x)}{\textbf{T}})}$  & Low & \cmark & \smark \\
    \midrule
    HB \cite{zadrozny2001histogrambinning} & Correctness alignment & Non-parametric & $\hat{f}(x) = f(x) + \textbf{b}, \textbf{b} = \mathds{E}_x[Y - f(x) | f(x)=\alpha], x \in \{f(x) = \alpha\} $ & Low & \cmark & \smark \\
    \midrule
    Scale-bining \cite{kumar2019verified} & Correctness alignment & Hybrid & $ p(y | x) = \frac{1}{B_i}\text{HB}_{x_2}(g(x)), g(x) = \sigma_{x_3}(a \text{log} \frac{x}{1-x} + c)$ & Low & \cmark & \smark \\
    \bottomrule
    \end{tabular}
\end{table*}

We now describe the methods of calibration for LLMs. These methods can be divided into three parts for their treatment of LLM: open-box, post-hoc, and closed-box calibration methods. Table~\ref{tab:cal_llm_methods} shows the representative ones for LLMs, with their objective, approach, and formulations.

\noindent{\bf Open-box calibration.} Early methods of calibration were focused on fine-tuning or pre-training of LM by proposing a novel or extended loss function with a new metric \cite{muller2019whendoeslabelsmoothinghelp, kong-etal-2020-calibrated-lm-fine-tuning}. One example is label smoothing \cite{muller2019whendoeslabelsmoothinghelp} that can be integrated into the loss function objective or included in the parameter-efficient fine-tuning (PEFT) phase of LLMs, such as Low-rank adaptation (LoRA) \cite{hu2021lora}. While this method can make NMT and sequence models more calibrated, however, it can result in information loss in logits \cite{muller2019whendoeslabelsmoothinghelp, desai2020calibrationprtransformers}. 

Calibrated fine-tuning of LMs for multi-choice and extractive QA usually involves penalizing model confidence for incorrect answers and augmenting the model certainty for correct answers \cite{jiang2021qalmcalibration}. LitCab \cite{liu2024litcab} method tunes a single linear layer over the logits from the last layers of LMs using a max-margin objective function that uses prediction probabilities of generated negative samples and positive samples. This layer is then used to infer a bias term that can be used for modification of the LM's logits and, hence, the confidence of generation. Applying Laplace approximation to LoRA layers \cite{Yang2023Laplacelora} can provide better-calibrated confidence of LLMs.

SLiC-HF \cite{zhao2023slichf} considers negative and positive human preferences in the fine-tuning method of Sequence-likelihood calibration (SLiC) \cite{zhao2023calibratingsli}, which calibrates the confidence of token-ranking with the reinforcement learning (RL) reward model. Other lines of work propose fine-tuning datasets to make LLM aware of unsure responses \cite{zhang2024rtuning,xu2024sayself}. Fine-tuning-based methods were quite complicated to extend to further language models \cite{tian2023justaskforcalibration} and can be complex to apply to LLMs. In addition, instruction tuning was found to harm the calibration of Alpaca  \cite{zhu2023calibrationandalignment}.

\noindent{\bf Post-hoc calibration.} Perfect calibration is possible to achieve using the proposed methods, and it can be easier to achieve than perfect accuracy \cite{nguyen2015posteriorcal}. Calibration methods that can be applied after model training (post-hoc) can be divided into parametric and non-parametric methods \cite{fisch2022uncertaintyfornlptutorial}. TS \cite{guo2017calibration} is a parametric method applied after training or post-hoc to the rigid softmax outputs. Such a method requires assigning a temperature parameter $T$, which we divide into logits to make them softer or sharper for over-confident or under-confident predictions. We train the scaling parameter $T$ on the held-out dataset using an optimization function. However, temperature scaling may not capture all of the uncertainties, such as epistemic, and it does not capture point-wise uncertainty and calibrate locally \cite{ovadia2019can}. TS is also considered a simplified version of Platt scaling \cite{platt1999prbbplattscaling}, a sigmoid-based calibration method, where we optimize scale $a$ and bias $b$ to output calibrated probabilities using:

\begin{equation}
    p(y_i | x) = \sigma (a \text{logit}f_i(x) + b) = \frac{1}{1 + \text{exp}( - (a \text{logit}f_i(x) + b))}.
    \label{eq:platt_scale}
\end{equation}

TS is still suggested for a parametric method of calibration and is considered to be data-efficient and low-complexity calibration of probabilities \cite{xiao2022uqwithplm, geng2024surveyllmcal}. Temperature scaling does not affect the accuracy of the model as it does not affect the most confident prediction \cite{widmann2021multiclasscalkernel}. However, it can be hard to verify as it only considers the output probabilities of the last layer. It can also fail to calibrate epistemic uncertainty in data drifts \cite{ovadia2019can}.

Histogram binning (HB) \cite{zadrozny2001histogrambinning} is a non-parametric method that divides prediction probabilities in bins similar to ECE, however, we assign bin score $\theta$ to each bin and output this score if the probability falls into the bin of that score. The bin score is optimized using bin-wise squared error. Isotonic regression \cite{zadrozny2002isotonicregression} generalizes HB by optimizing both bin boundaries and bin scores. Bayesian Binning into Quantiles (BBQ) \cite{naeini2015bbq} also extends HB using the Bayesian model, averaging over possible binning boundary settings to get calibrated probabilities using the uniform and pre-defined prior of binning settings and scores. One limitation of non-parametric methods, as revealed by our preliminary experiments, is the reduction in model performance when the training dataset size is small. This finding aligns with the results reported by \cite{niculescu2005predictingreldiag}. Another notable limitation of non-parametric methods is their reliance on a considerable amount of data to achieve reduced variance, making them less effective in data-scarce scenarios. Non-parametric local recalibration (LoRe) method is a point-wise extension of histogram binning with a local calibration error (LCE) metric that considers bin-wise differences of confidence and accuracy of points with kernel similarity of features to balance the calibration of individual data points and similarly grouped data points \cite{luo2022lce}. 

Scale-binning calibrator is a hybrid method that combines parametric and non-parametric calibration approaches \cite{kumar2019verified}. In this method, we apply scaling on one part of the held-out dataset, $B$ similar bins on the second part based on the empirical distribution of the function, and HB on the third part. Scale binning can provide more stable measures; however, it only considers raw predictive scores

\noindent{\bf Closed-box calibration.} LLMs can be linguistically calibrated to mitigate the inaccurately expressed verbal confidence of the model \cite{mielke2022reducingoverconfidence}. A recent study \cite{xiong2023canllmexpress} also found that open-box methods have only a narrow gap for addressing uncertainty compared to closed-box linguistics. While it is unclear if CoT prompting improves verbalized calibration, a recent study revealed that linguistic calibration can be more accurate than the numerical methods \cite{tian2023justaskforcalibration}. \cite{becker2024cyclesofthought} stated that calibration metrics rely on the inductive hypothesis that training, test, and calibration data should be sampled from the same distribution, which is neither verifiable nor falsifiable at test time. ECE conflates the uncertainty and depends on accuracy and binning strategy \cite{becker2024cyclesofthought}. In \cite{chen2023plmcal}, authors consider measuring confidence in the wrong prediction for pre-trained language models. 

In \cite{zhang2024calibratingbyelicitingfidelity}, the authors measure the fidelity of the answer in a multi-choice setting toward other answers by replacing its expression with the remaining options. If the answer is chosen again after a chain of such modifications, then fidelity is considered high, and confidence is calibrated by multiplying to normalized uncertainty. Providing fact and reflective reasoning was also found to improve the calibration \cite{zhao2024far}. Self-consistency \cite{wang2022selfconsistency} prompts LLM multiple times and samples answers to score the most frequent ones. Combined with self-evaluation \cite{huang2024calibratinglongform}, self-consistency can calibrate long sequences for LLMs. Multiple prompt augmentation methods, such as paraphrasing, permutation of variants, and in-context samples, can be ensembled by averaging the probabilities to get calibrated confidence \cite{jiang2023cape}.

In the SAPLMA~\cite{azaria2023saplma} method, the author trains an auxiliary model to check if LLMs are lying on true-false statements based on the internal layers.  The layers of the model are taken from three hidden activation layers of LLMs, which are further trained three times based on the topic. Apricot \cite{ulmer2024apricot} trains a small auxiliary model on LLM outputs and cluster-wise accuracy for confidence elicitation from closed-box and open-box LLMs. Auxiliary calibrators are smaller in size and do not have many generations; hence can reduce inference costs.

\noindent{\bf Bias reduction.} Recent works pay attention to a novel objective of calibration, which is mitigating bias toward certain classes \cite{zhou2023batchcalibration, han2022prototypicalcalib, roelofs2022domancontextcalib, kumar2022answerlevecalib}. The root causes of such tendencies are majority label, recency, and common token biases \cite{zhao2021contextualcalib}. Prototypical calibration \cite{han2022prototypicalcalib} methods usually decide the boundary in output distribution for the final choice of class. Batch calibration (BC) \cite{zhou2023batchcalibration} quantifies the bias of context from a batch of size $M$ and divides log probabilities to average per-class prior obtained by the calibration layer. 

While the previously mentioned methods apply to binary classification tasks, multi-calibration \cite{hebert2018multicalibrationigbh, gopalan2022lowdegreemulticalib, detommaso2024multicalibrationiglb} methods reduce bias in multi-class settings. In multi-calibration \cite{detommaso2024multicalibrationiglb}, the objective is to reduce the bias conditioned on category by combining calibration methods, such as histogram binning\cite{zadrozny2001histogrambinning}, with a grouping strategy. PriDe~\cite{zheng2023llmselbiaspride} method addresses selection bias using prior bias estimation over cyclic permutations $n$ of option tokens and positions, assuming that observed probability can be divided into prior and de-biased ones.

\subsection{Applicability in LLMs}

While some of the calibration methods can be straightforwardly applied to LLMs, we outline important aspects and challenges to consider for better calibration.

\noindent{\bf Calibration and scale of model.} Recent studies have empirically shown that calibration in LLMs is influenced by model size \cite{kadavath2022llmsknowmostly, tran2022plex, chen2023plmcal}. Additionally, research has observed that as model size increases, generation uncertainty also tends to rise \cite{ye2024benchmarkingllmuq}. These findings highlight potential challenges in integrating standard calibration methods into large models, necessitating a deeper understanding of their implications.

\noindent{\bf Fine-tuning-based calibration.} The fine-tuning process also requires calibration, as a loss function objective \cite{muller2019whendoeslabelsmoothinghelp, lee2022adaptivels} or fine-tuning strategy, to provide better model inference. Bayes regularization can be used for fine-tuning calibration by applying it to the last layers with a deterministic attention layer or with a belief network~\cite{Fan2020BayesianAM, Zhang2021BayesianABN}. 

\noindent{\bf Calibration in multi-lingual and multi-modal large language models.} \edit{Multilingual language models face domain-shift and scaling challenges, and mitigating these issues through dictionaries and multilingual embedding similarity has gained attention \cite{gou2023critic, azaria2023saplma}. Recent advances such as Entropy2Vec \cite{irawan2025entropy2veccrosslinguallanguagemodeling} analyze the entropy of monolingual models applied to other languages to derive language vectors that capture cross-lingual typological relationships. Evaluating and improving faithful calibration of LLMs on non-English tasks presents an interesting challenge, as humans exhibit substantial differences in their use of linguistic uncertainty markers across cultures and languages \cite{liu2025metafaith}. Meanwhile, large vision-language models (LVLMs) exhibit input-related biases and uncertainties that are increasingly addressed through batch calibration and post-hoc calibration techniques \cite{zhou2023batchcalibration, dai2023exploringmllmood, upadhyay2023probvlm}. Fine-tuning remains a crucial step for producing reliable outputs in large language models (LLMs), as training from scratch is often prohibitively time-consuming. However, Parameter-Efficient Fine-Tuning (PEFT) methods such as LoRA \cite{hu2021lora} can sometimes degrade model quality in larger models. Understanding these fine-tuning trends is important because contributions from local data can reveal uncertainty gaps caused by slight domain shifts between training and test sets, where out-of-distribution (OOD) methods can help handle differences in samples and features \cite{yang2024generalizedoodsurvey}. Additionally, prompt-perturbation strategies, combined with evaluations of cross-lingual and cross-model consistency, have been used to assess semantic and multilingual inconsistencies, with mitigation supported by Retrieval-Augmented Generation (RAG) \cite{ding2024rowen}.}

\noindent{\bf Estimation and calibration for long sequences.} Calibration of longer sequence outputs of models is a relatively new concept \cite{muller2019whendoeslabelsmoothinghelp, liu2024litcab}, as previous studies were focused on short-form generations \cite{tian2023justaskforcalibration, zhou2023navigating}. Early investigations found that models tend to diffuse probability distribution over the sequence \cite{ott2018analyzinguqnmt}. Calibration for longer generations can be a harder problem. The model can have increased over-confidence or under-confidence at the end token probability of the sentence \cite{kumar2019nmtcalibration}. Token-level confidence also limits context to the current inference instead of representing the overall predictive uncertainty of a token \cite{xiao2021hallucinationanduncertainty}. Calibration of long-form generation gained attention in the recent works \cite{band2024linguistic, zhang2024luq}. Another recent study \cite{tian2023justaskforcalibration} found that verbalized confidences are usually more calibrated than the model's inference probabilities for recent LLMs, though temperature scaling can still be effective for the calibration of LLMs. 

\noindent{\bf Better binning strategy.} Measuring the original calibration error is impossible with a finite number of bins \cite{kumar2019verified}. One modification is choosing adaptive bin sizes instead of evenly sized bins according to data density \cite{nixon2019measuringcalibrationindl}. Using Kernels-wise calibration \cite{widmann2021multiclasscalkernel}, existing methods trade off the calibration for accuracy when applied on the likelihood of all sequences \cite{brocker2009reliabilitysufficiency} and reduce the categorical discriminative aspect of the model \cite{naeini2015bbq}. Individualized calibration of the categories of a dataset is an open challenge.

\noindent{\bf Selective calibration and abstention.} \reviewerTwo{Selective calibration and abstention techniques offer significant practical benefits by enabling models to selectively withhold answers when uncertainty is high, thus improving reliability. TS can be employed to calibrate large language models’ confidence thresholds for abstention decisions. Recent advances include SConU \cite{wang2025sconu}, which leverages selective conformal uncertainty to improve abstention with the exchangeability criterion, which calibrates the sample set. Surrogate Models provide another promising approach, simulating complex model behaviors to inform abstention \cite{shrivastava2023llamasknowwhatgptsdontshow}. Additionally, multi-LLM frameworks use cross-model competitive or collaborative feedback to decide when to abstain in QA, enhancing robustness through ensemble consensus \cite{feng2024don}. While gaining recent attention, these methods have increased inference cost and time. Inclusion of measurement of uncertainty as an intrinsic part of these methods can improve selective calibration and inference \cite{zhou2023navigating, zhou2024relyingonunreliable}.
}

\reviewerTwo{Better prompting techniques can improve selective calibration as well. Self-evaluation of LLM can refine abstention based on model confidence \cite{ren2023selfeval} while adding inference time. A notable method, AbstainPrompt \cite{madhusudhan2024abstainprompt},  investigates the abstention capabilities of LLMs using prompt engineering and ICL to guide selective answering. However, the method is only applicable to classification tasks and smaller models revealed less abstention.}

\section{RQ3: Evaluation}
\label{sec:evaluation}

\revision{Having reviewed UQ and calibration methods for LLMs, we now empirically evaluate how well representative methods perform in practice.} We conduct a rigorous benchmarking experiment and provide an analysis of the evaluation results.

\noindent{\bf Large Models.} 
Amongst open-box LLMs, we use instruction fine-tuned versions of Llama 3.1 8B \cite{dubey2024llama3.1}, Mistral v0.3 7B \cite{jiang2023mistral}, and Qwen2.5 7B \cite{yang2024qwen2.5}. We also include the recent DeepSeek-R1~\cite{guo2025deepseekrone}, which uses a reinforced reasoning method with distilled fine-tuning. For closed-box LLMs, we use GPT-4 \cite{achiam2023gpt4} and GPT-4o \cite{hurst2024gpt4o}, as OpenAI also recommends it over GPT-3.5-Turbo \cite{brown2020languagegpt3} for being faster and more capable.



%

\noindent{\bf Datasets.} For the calibration experiment on the generation task involving \edit{open-ended QA}, we use:
(i) {TriviaQA \cite{joshi2017triviaqa}}; 
(ii) {TruthfulQA  \cite{lin2021truthfulqa}}; \edit{and (iii) {Natural Questions (NQ) \cite{kwiatkowski-etal-2019-nq}}}. Further details of the datasets are provided in Appendix~A-A.




\noindent{\bf Metrics.}  For the calibration experiment on the \edit{open-ended generative QA} task, we use ECE, SmoothECE (smECE), BS, AUROC, AURC, and average BLEU score. ECE and SmoothECE are error metrics to evaluate the calibratedness of LLMs. The number of bins for the ECE metric is 10. BS is not the direct calibration metric, but it measures the amount of data variance the model can understand \cite{gopalan2022lowdegreemulticalib}. AUROC measure shows the indicativeness of model confidence for incorrect predictions. The average BLEU score measures the average confidence distance of generated answers to ground truth. We put a 0.3 threshold for the BLEU score. To observe the variance within the metric results, we sample 50 evaluation iterations.
Note that we provide a detailed description of the evaluation metrics used in this work in Section~\ref{sec:eval_metrics}. \edit{For NQ we also report the AURC metric that measures how well a model can balance prediction accuracy with coverage by quantifying the quality of uncertainty.}



\noindent{\bf Baselines.} We follow the setting for baselines as in \cite{ulmer2024apricot}, with some additional baselines explained as follows: normalized and average sequence likelihood with few-shot learning, few-shot learning and CoT tokens, and the recent method of verbalized qualitative uncertainty \cite{lin2022teachingmodelstoexpress, tian2023justaskforcalibration, zhou2023navigating, xiong2023canllmexpress} method to obtain confidence of the model. We ask verbally to provide confidence from ``very high'' to ``very low'' and then map this scale to numerical values using the scheme we describe in Appendix~A-B. In addition, we apply Platt Scaling \cite{platt1999prbbplattscaling} to average sequence likelihoods and TS \cite{guo2017calibration} to logit scores of generated sequences for obtaining scaled sequence likelihoods.

\begin{table*}[!ht]
    \centering
    \caption{Calibration results for Llama-3.1-8B-Instruct, Mistral-7B-Instruct-v0.3, Qwen2.5-7B-Instruct, GPT-4, and GPT-4o on TriviaQA and TruthfulQA.}
    \label{tab:calibration_results}
\begin{tabular}{
    >{\arraybackslash}p{0.13\textwidth}
    >{\centering\arraybackslash}p{0.05\textwidth}
    >{\centering\arraybackslash}p{0.06\textwidth}
    >{\centering\arraybackslash}p{0.04\textwidth}
    >{\centering\arraybackslash}p{0.07\textwidth}
    >{\centering\arraybackslash}p{0.08\textwidth}
    >{\centering\arraybackslash}p{0.05\textwidth}
    >{\centering\arraybackslash}p{0.06\textwidth}
    >{\centering\arraybackslash}p{0.04\textwidth}
    >{\centering\arraybackslash}p{0.07\textwidth}
    >{\centering\arraybackslash}p{0.08\textwidth}
}
\toprule
\textbf{Models \& Baselines} & 
\multicolumn{5}{c}{\textbf{TriviaQA}} & \multicolumn{5}{c}{\textbf{TruthfulQA}} \\
\cmidrule(lr){2-6} \cmidrule(lr){7-11}
& ECE ↓ & smECE ↓ & BS ↓ & AUROC ↑ & $\text{BLEU}_{avg}$ ↑ & ECE ↓ & smECE ↓ & BS ↓ & AUROC ↑ & $\text{BLEU}_{avg}$ ↑ \\
\midrule
\rowcolor[HTML]{EFEFEF} \multicolumn{11}{l}{\textbf{Llama-3.1-8B-Instruct}} \\

Few-shot            &	0.046 &	0.040 &	0.286 &	0.532 &	0.041 & 0.386 &	0.333 &	0.165 &	0.794 &	0.190 \\
CoT Few-shot        &	0.087 &	0.065 &	0.273 &	0.539 &	0.033 & 0.346 &	0.309 &	0.136 &	0.569 &	\bf 0.191	 \\
Verbal. Qual CoT    & 0.113 &	0.064 &	0.293 &	\bf 0.564 &	0.033 & 0.499 &	0.348 &	0.349 &	0.161 &	0.191  \\
Platt Scaling       & \bf 0.013 & \bf 0.013 &	\bf 0.249 &	0.495 &	0.041 &	 0.363 &	0.319 &	0.134 &	0.794 &	0.190 \\
Temp. Scaling       & 0.045 & 0.039 &	0.287 &	0.505 & \bf	0.041 &	\bf 0.154 & \bf	0.154 & \bf	0.032 & \bf	0.794 &	0.190 \\
\midrule

\rowcolor[HTML]{EFEFEF} \multicolumn{11}{l}{\textbf{Mistral-7B-Instruct-v0.3}} \\

Few-shot            & 0.044 &	0.043 &	0.265 &	0.621	& 0.228 & 0.241 &	0.233 &	0.075	& 0.554 &	0.426	 \\
CoT Few-shot        & 0.025 &	0.034 &	0.283 & \bf	0.702 &	\bf 0.310  & 0.219	& 0.214 &	0.062 &	0.261 &	0.386	\\
Verbal. Qual CoT    & 0.042 &	0.028 &	0.268 &	0.547 &	0.310 & 0.467 &	0.369 &	0.251 &	0.535 &	0.386 \\
Platt Scaling       & \bf 0.010	&  \bf 0.010 & \bf	0.250 & 0.621 & 0.228 & 0.176 &	0.175 &	0.034 &	0.554	& 0.426 \\
Temp. Scaling       & 0.042 &	0.039 &	0.262 &	0.621 &	0.228  & \bf 0.057 & \bf	0.058 &	\bf 0.007 & \bf	0.554 &	\bf 0.426 \\
\midrule

\rowcolor[HTML]{EFEFEF} \multicolumn{11}{l}{\textbf{Qwen2.5-7B-Instruct}}\\

Few-shot            & 0.310 &	0.286 &	0.508 &	0.547 & \bf	0.034 & 0.187	& 0.185 &	0.044 &	0.543 &	0.187 \\
CoT Few-shot        & 0.413 & 	0.350 &	0.483 &	0.631 &	0.012 & 0.180 &	0.180 &	0.040 & \bf	0.867 &	0.184	 \\
Verbal. Qual CoT    & 0.164 &	0.162 &	0.283 &	\bf 0.658 &	0.012 & 0.489 &	0.392 &	0.246 &	0.516 &	0.185 \\
Platt Scaling       & 0.142 &	0.141 &	0.208 &	0.417 &	0.012 & 0.363 &	0.320 &	0.135 &	0.543 &	0.187 \\
Temp. Scaling       & \bf 0.044 & \bf	0.037 & \bf	0.205 &	0.56 &	0.012 & \bf 0.066 & \bf	0.067 &	\bf 0.008 &	0.543 & \bf	0.187 \\
\midrule

\rowcolor[HTML]{EFEFEF} \textbf{GPT-4} & & & & & & & & & & \\

Few-shot            & \bf  0.101	& \bf 0.075 & \bf	0.151 &	0.789 &	0.721  & 0.325 &	0.232 &	0.222	& 0.833 &	0.347	\\
CoT Few-shot        & 0.217 &	0.210 &	0.297 &	0.572 &	0.051 & 0.635 &	0.349 &	0.664	& 0.367 &	0.181 \\
Verbal. Qual CoT    & 0.144 & 0.079 &	0.279 &	0.552 &	0.051 & 0.269 &	0.255 & \bf	0.088	& 0.539 &	0.181 \\
Platt Scaling       & 0.232	 & 0.225 &	0.206 &	0.789 &	0.721  & 0.352 & 0.313 &	0.136	& 0.914 &	0.347 \\
Temp. Scaling       & 0.105 &	0.077 &	0.153 & \bf	0.789 & \bf	0.721 & \bf 0.230 & \bf	0.145 &	0.185 & \bf 0.916 & \bf 0.347 \\
\midrule

\rowcolor[HTML]{EFEFEF} \textbf{GPT-4o} & & & & & & & & & & \\

Few-shot            &  0.090 & 0.079 &	\bf 0.163 &	0.765 &	0.653 &  0.277 &	0.259 &	0.113 &	0.408 &	0.301 \\
CoT Few-shot        & \bf 0.027 & \bf 0.034 &	0.263 &	0.613 &	0.042 & 0.488 &	0.345 &	0.314 &	0.285 &	0.147 \\
Verbal. Qual CoT    & 0.310 &	0.157 &	0.331 &	0.543 &	0.042 & 0.159 &	0.158 &	0.056 &	0.246 &	0.147 \\
Platt Scaling       & 0.232 &	0.225 &	0.216 &	0.765 &	0.653 & 0.366 &	0.321 &	0.137 & 0.408 & 0.301 \\
Temp. Scaling       & 0.096 &	0.085 &	0.165 &	\bf 0.765 &	\bf 0.653 & \bf 0.118 &	\bf 0.116 & \bf 0.027 & \bf	0.408 &	\bf 0.301\\
\midrule

\rowcolor[HTML]{EFEFEF} \textbf{GPT-5.2} & & & & & & & & & & \\
Verbal. Qual        & \bf 0.201 & \bf 0.101 & \bf 0.236 & 0.595 & \bf 0.458  & 0.156 & 0.154 & \bf 0.063 & \bf 0.672 & \bf 0.252 \\
Verbal. Qual CoT    & 0.311 & 0.161 & 0.345 & 0.595 & 0.031 & \bf 0.149 & \bf 0.142 & 0.064 & 0.290 & 0.144 \\
\midrule

\rowcolor[HTML]{EFEFEF} \textbf{DeepSeek-R1} & & & & & & & & & & \\

Verbal. Qual        & 0.067 & 0.061 & \bf 0.162 & \bf 0.594 & \bf 0.440 & 0.325 & 0.295 & 0.117 & 0.381 & \bf 0.225 \\
Verbal. Qual CoT    & \bf 0.047 & \bf 0.038 & 0.232 & 0.527 & 0.008 & \bf 0.314 & \bf 0.288 & \bf 0.108 & \bf 0.405 & 0.072 \\

\bottomrule
\end{tabular}
\end{table*}

\begin{table}[!t]
    \centering
    \caption{\edit{Evaluation results on NQ for Llama-3.1-8B, Mistral-7B-v0.3, Qwen2.5-7B, GPT-4o, GPT-5.2, and DeepSeek-R1.}}
    \label{tab:calibration_results_nq}
    \setlength{\tabcolsep}{3pt}
    \renewcommand{\arraystretch}{0.95}
    \footnotesize
\begin{tabular}{@{}lcccccc@{}}
\toprule
\textbf{Baseline} & ECE & smECE & BS & AUROC & AURC & BLEU \\
\midrule
\rowcolor[HTML]{EFEFEF} \multicolumn{7}{l}{\textbf{Llama-3.1-8B-Instruct}} \\
Few-shot            & 0.197 & \bf 0.126 & 0.417 & 0.514 & \bf 0.783 & 0.018 \\
Few-shot CoT        & 0.429 & 0.359 & 0.410 & 0.520 & 0.736 & 0.020 \\
Verbal.\ Qual CoT       & \bf 0.186 & 0.150 & \bf 0.302 & \bf 0.560 & 0.706 & \bf 0.020 \\
PS                  & 0.396 & 0.339 & 0.333 & 0.514 & 0.783 & 0.018 \\
TS                  & 0.437 & 0.363 & 0.437 & 0.514 & 0.783 & 0.018 \\
\midrule
\rowcolor[HTML]{EFEFEF} \multicolumn{7}{l}{\textbf{Mistral-7B-v0.3}} \\
Few-shot            & 0.328 & 0.297 & 0.442 & 0.633 & 0.631 & 0.080 \\
Few-shot CoT        & 0.215 & 0.210 & 0.438 & \bf 0.652 & 0.612 & 0.059 \\
Verbal.\ Qual CoT       & \bf 0.097 & \bf 0.065 & \bf 0.291 & 0.630 & \bf 0.631 & 0.059 \\
PS                  & 0.269 & 0.255 & 0.306 & 0.633 & 0.631 & 0.080 \\
TS                  & 0.239 & 0.231 & 0.351 & 0.633 & 0.631 & \bf 0.080 \\
\midrule
\rowcolor[HTML]{EFEFEF} \multicolumn{7}{l}{\textbf{Qwen2.5-7B-Instruct}} \\
Few-shot            & 0.539 & 0.415 & 0.550 & 0.617 & 0.776 & 0.015 \\
Few-shot CoT        & 0.487 & 0.390 & 0.565 & \bf 0.644 & 0.725 & 0.015 \\
Verbal.\ Qual CoT       & 0.294 & 0.272 & 0.298 & 0.615 & 0.753 & 0.015 \\
PS                  & 0.313 & 0.286 & 0.236 & 0.617 & 0.776 & 0.015 \\
TS                  & \bf 0.016 & \bf 0.016 & \bf 0.146 & 0.617 & \bf 0.776 & \bf 0.015 \\
\midrule
\rowcolor[HTML]{EFEFEF} \multicolumn{7}{l}{\textbf{GPT-4o}} \\
Few-shot            & 0.205 & 0.196 & 0.343 & 0.670 & 0.381 & 0.161 \\
Few-shot CoT        & \bf 0.117 & \bf 0.112 & 0.293 & 0.644 & 0.458 & 0.036 \\
Verbal.\ Qual CoT       & 0.365 & 0.204 & 0.427 & 0.602 & \bf 0.475 & 0.036 \\
PS                  & 0.185 & 0.183 & \bf 0.281 & 0.670 & 0.381 & 0.161 \\
TS                  & 0.201 & 0.195 & 0.339 & \bf 0.670 & 0.381 & \bf 0.161 \\
\midrule
\rowcolor[HTML]{EFEFEF} \multicolumn{7}{l}{\textbf{GPT-5.2}} \\
Verbal.\ Qual       & \bf 0.342 & \bf 0.188 & \bf 0.411 & \bf 0.648 & 0.492 & \bf 0.120 \\
Verbal.\ Qual CoT   & 0.391 & 0.215 & 0.442 & 0.611 & \bf 0.514 & 0.022 \\
\midrule
\rowcolor[HTML]{EFEFEF} \multicolumn{7}{l}{\textbf{DeepSeek-R1}} \\
Verbal.\ Qual       & 0.153 & 0.150 & 0.280 & 0.488 & \bf 0.541 & 0.003 \\
Verbal.\ Qual CoT       & \bf 0.073 & \bf 0.072 & \bf 0.239 & \bf 0.500 & 0.441 & \bf 0.005 \\
\bottomrule
\end{tabular}
\end{table}

\reviewerOne{Our results reveal that calibration success depends critically on model size, prompt sensitivity, and domain shift. Regarding model size, smaller models such as Qwen2.5 7B exhibit systematically worse calibration compared to larger models: GPT-4 and GPT-4o benefit from scale, achieving lower ECE and higher AUROC across benchmarks. However, scale alone does not determine calibration quality; DeepSeek-R1 achieves competitive calibration gains through its reinforcement-learning-based reasoning process rather than relying on massive parameter counts. Prompt sensitivity is another key factor: verbalized uncertainty extraction succeeds at different rates across models, and CoT prompting substantially improves calibration for some models (e.g., DeepSeek-R1) while yielding marginal gains for others. Few-shot prompt design also affects calibration stability, as different exemplar orderings and formats can shift confidence distributions. Under domain shift, our NQ results show a 15\% to 22\% AUROC drop compared to in-domain benchmarks, confirming that vocabulary mismatch and training domain divergence degrade calibration. Post-hoc methods such as TS and Platt scaling partially mitigate out-of-domain miscalibration for Qwen2.5 and Mistral v0.3, but we do not observe significant improvement for Llama 3.1, suggesting that the effectiveness of recalibration is model-dependent and influenced by the alignment between the held-out calibration set and the target domain.}

\reviewerTwo{The instruction tuning methodology also plays a confounding role in cross-model calibration comparison. Qwen2.5 and Mistral v0.3 employ different reinforcement learning from human feedback (RLHF) intensities and preference optimization strategies during their alignment phases, and recent literature suggests that aggressive RLHF can degrade calibration by training models to express uniformly high confidence regardless of actual correctness, an effect sometimes referred to as the ``alignment tax'' \cite{zhu2023calibrationandalignment}. DeepSeek-R1, by contrast, uses reinforcement-learning-based reasoning rather than standard RLHF, which yields qualitatively different calibration behavior: its reasoning traces provide a structured self-assessment that improves calibration without the flattening effect of preference-based alignment. GPT-4o's content filtering and safety tuning may further affect expressed confidence, as moderation layers can suppress or modify probability distributions for sensitive content. Overall, the style of instruction tuning (whether supervised fine-tuning (SFT), RLHF, or RL-based reasoning) constitutes a significant confounding factor that should be controlled for when interpreting cross-model calibration differences.}

\noindent{\bf Analyses.} In this section, we will discuss the suitability of methods and metrics for LLMs based on our experimental evaluation. In general, we notice that calibration methods can make models provide a more accurate probability of their predictions, hence making them more reliable, as noted in Table~\ref{tab:calibration_results}. \edit{For NQ dataset, we report the results of the experiment in Table~\ref{tab:calibration_results_nq}}. We bold the best results per dataset and model and underline those that are statistically significant compared to all other results after multiple experimental runs.

\reviewerTwo{On the more challenging NQ benchmark (Table~\ref{tab:calibration_results_nq}), combining chain-of-thought (CoT) prompting with verbalized uncertainty expression (Verbal.~CoT) often yields confidence estimates that are both better calibrated (lower ECE/smECE and BS) and more discriminative for detecting correct vs.~incorrect answers (higher AUROC / improved AURC). For Llama-3.1-8B, Verbal.~CoT achieves the lowest ECE (0.186) and the best AUROC (0.560) among its baselines, improving over Few-shot CoT (ECE 0.429, AUROC 0.520). For Mistral-v0.3, Verbal.~CoT sharply reduces calibration error (ECE 0.097, smECE 0.065) and improves BS (0.291) compared to Few-shot (ECE 0.328, BS 0.442), while keeping AUROC competitive (0.630 vs.~0.652 for Few-shot CoT). For DeepSeek-R1, adding CoT to verbalized uncertainty improves all reported metrics simultaneously (ECE 0.153$\rightarrow$0.073, smECE 0.150$\rightarrow$0.072, BS 0.280$\rightarrow$0.239, AUROC 0.488$\rightarrow$0.500). Overall, these results suggest that CoT provides a structured reasoning trace that helps the model internally assess evidence strength, and verbal uncertainty makes this self-assessment explicit, yielding a confidence signal that better correlates with correctness and supports more reliable error detection under ambiguity and distributional shift.}

\edit{For TriviaQA dataset, we provide reliability diagrams for Qwen2.5, Llama3.1, and Mistral v0.3 in Figure~\ref{fig:llama_trivia_qa_comparison}. We provide the remaining diagrams in Appendix~A-E. Our analysis reveals that parametric calibration methods make bin-wise confidences align with bin-wise accuracy, hence making the model more calibrated, though we observe that the Qwen2.5 model exhibits low calibration and accuracy.} As this model is a multi-lingual model and domain knowledge is also different, it can affect the accuracy of generation and the language mixing problem \cite{guo2025deepseekrone}. In this context, the calibration of LLMs across different domains, such as in scenarios involving dataset shift \cite{lin2022teachingmodelstoexpress}, remains an interesting and open problem for further investigation. 

TS can be implemented in two ways: (1) scaling the model confidences or log probabilities of candidate sequences \cite{jiang2021qalmcalibration, tian2023justaskforcalibration, ulmer2024apricot} or (2) scaling the logits before the softmax layer, which operates on the vocabulary size. In the latter approach, the temperature is tuned using the IDs of generated tokens as a reference \cite{liu2024litcab}. Our results indicate that Platt and temperature scaling applied on a confidence level in (1) is effective for out-of-domain calibration for both Qwen2.5 and Mistral v0.3, but we do not see significant improvement for Llama 3.1. We note that TS significantly improves model confidence on the TruthfulQA dataset, where the model is expected to be uncertain about questions with false premises.

We notice that domain knowledge, vocabulary, and model size highly affect the calibration since we see GPT-4 results are most calibrated with high accuracy on both TriviaQA and TruthfulQA datasets.  However, temperature scaling applied to LLM logits can have high memory requirements, given the large vocabulary sizes of modern LLMs and our limited budget to run such an evaluation. The ICL-based batch calibration~\cite{zhou2023batchcalibration} also applies at the logit level; hence, it might have the same limitation.  While recent Qwen2.5 \cite{yang2024qwen2.5} and Llama 3.1 \cite{dubey2024llama3.1} have even larger vocabularies of 152,064 and 128,256, respectively, GPT-4's tokenizer has unofficially been known to have a vocabulary size of 100,277 \cite{achiam2023gpt4}. Closed-box models like GPT-4 and GPT-4o do not publicly provide logits or vocabulary sizes, limiting researchers to using log probabilities of predicted tokens and the top 20 most likely tokens for each position. Open questions: How to determine the optimal layer for applying scaling to further improve calibration, e.g., should scaling be applied to the last layer or the penultimate layer for better results?

While the TS applied to model confidence shows promise, further research is necessary to explore additional parameters and techniques for optimal temperature scaling. In contrast, Platt scaling \cite{platt1999prbbplattscaling}, which employs a sigmoid function to adjust output probabilities, may not be suitable for capturing the complex functional relationships in modern LLMs. Our analysis also proves that Platt scaling is more effective when the held-out dataset size is small and when inference probability distortions closely follow a sigmoidal function \cite{niculescu2005predictingreldiag}.

\reviewerOne{For the verbalized uncertainty baseline, we also report the success rate of extraction of verbal uncertainty from the generated answers in figures for verbalized CoT. For DeepSeek-R1, the success rate was 100\% for most of the evaluations, hence we do not report in graphs.}

For GPT-4 and GPT-4o results in Figure~\ref{fig:gpt4_trivia_qa_comparison}, we note that CoT tokens can improve reasoning capabilities and contribute to calibration, combined with verbalized qualitative uncertainty. Also, GPT-4o is known to be smaller in size than GPT-4, and there is a difference in training domain \cite{hurst2024gpt4o}. GPT-4o also has the functionality to moderate and classify safe content from harmful and hazardous information while also filtering personal data \cite{hurst2024gpt4o}. We noticed some questions from the TriviaQA and TruthfulQA datasets belong to this category. We prompted the model to abstain from answering the question when the answer was against the content management policy. Because the model kept providing errors for some edge cases, we excluded those questions from both the training and testing phases. As the number of such questions ranges between 5 to 15, we believe it should not significantly affect the model's accuracy.

For the average BLEU score, we see that the metric is low mostly for baselines with CoT. This stems from the fact that the BLEU score is quite sensitive to sentence length, as it fails to capture structure similarity for wordy sequences.  We think the reason for the higher BLEU score of the Mistral v0.3 model is that its overall generations are quite compact and concise. Hence, we also compare the exact match (EM) alongside with BLEU score to judge the correctness of the generation. Overall, the Mistral v0.3 model revealed impressive accuracy and calibration, considering its relatively small size.

For DeepSeek-R1, we observe that the CoT baseline might improve the calibration and reduce ECE and SmoothECE scores. Reliability diagrams of DeepSeek-R1 on TriviaQA dataset are visualized in Figure~\ref{fig:dsr1_triviaqa}. Although we applied verbalized baselines on DeepSeek-R1, this model has improved calibration with the CoT baseline and comparable generation quality to closed-box models.

\reviewerTwo{As a sampling-based reference point, we also report Semantic Entropy (SE)~\cite{kuhn2023semanticuncertainty,farquhar2024detecting} in Appendix~A-D. On the out-of-domain NQ split SE attains the best ECE and smECE for both Llama-3.1 and Mistral-v0.3 (ECE $\approx$ 0.02-0.03), but does not consistently beat single-forward-pass baselines on AUROC or BLEU on in-domain TriviaQA, illustrating the estimation-quality-vs-cost trade-off discussed in Section~\ref{sec:future_research}.}

\reviewerOne{On the Natural Questions (NQ) dataset, models exhibit notable miscalibration, indicating that question hardness and ambiguity significantly contribute to increased model uncertainty. We observed a 15–22\% decrease in AUROC relative to in-domain benchmarks, along with elevated AURC values that reflect higher uncertainty under distributional shift. Question length also emerges as a relevant factor influencing uncertainty. In terms of model-specific trade-offs, GPT-4o shows the lowest relative increase in AURC, whereas Mistral v0.3 demonstrates a greater coverage error. Notably, verbalized uncertainty is found to be better calibrated when chain-of-thought reasoning is employed.}

\reviewerTwo{We also notice that verbal expressions of uncertainty are better with GPT-4 and GPT-4o, compared to the open-box models.}

\begin{figure*}[!t]
    \centering
    \subfloat[Llama 3.1]{
        \includegraphics[width=0.24\textwidth]{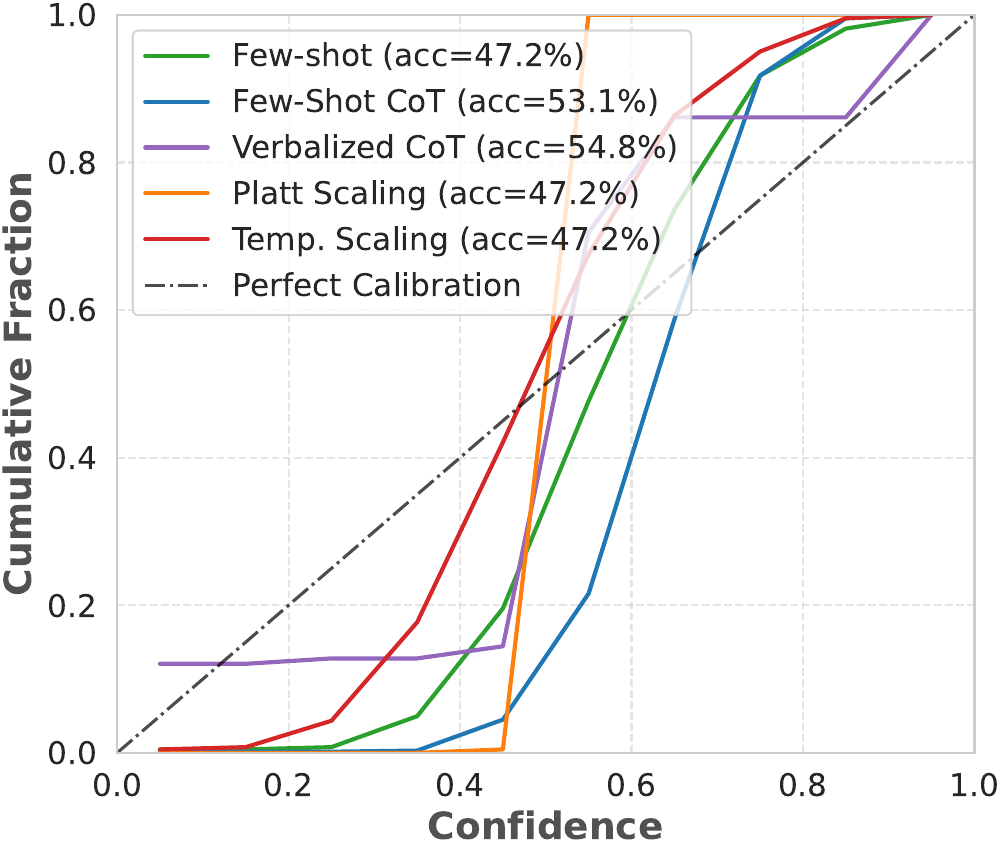}
        \label{fig:all_baselines_llama_triviaqa}}
    \subfloat[Qwen2.5]{
        \includegraphics[width=0.24\textwidth]{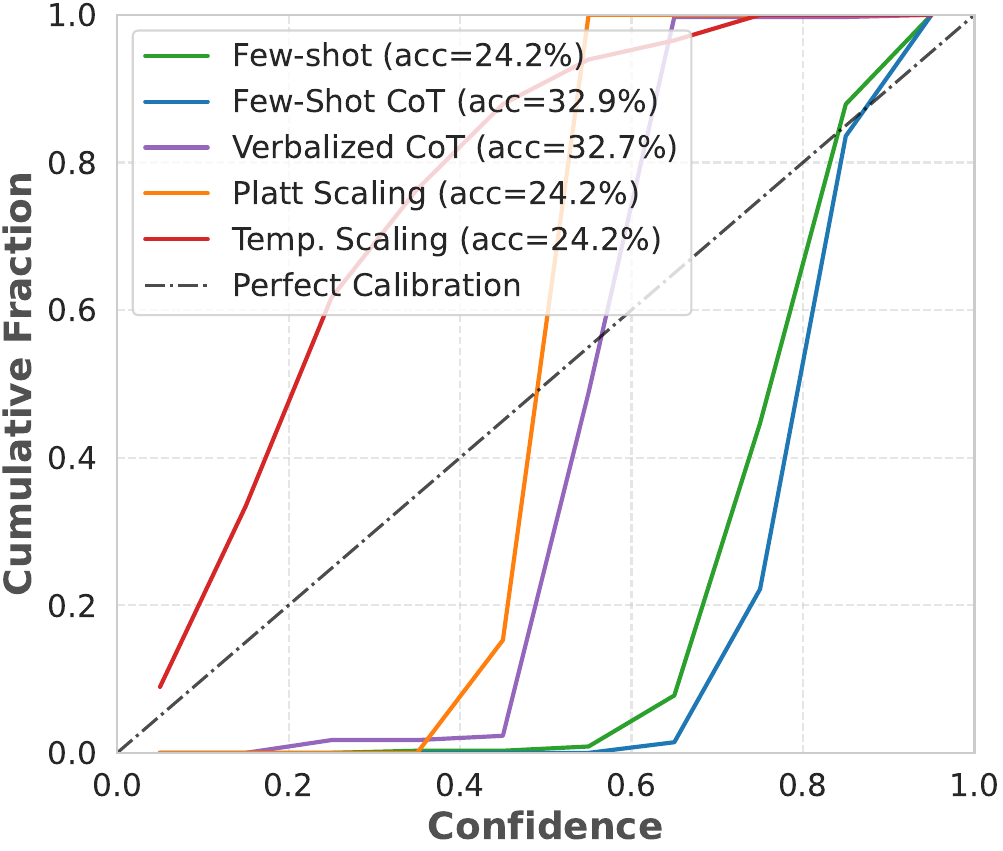}%
        \label{fig:all_baselines_qwen2.5_triviaqa}
    }
    \subfloat[Mistral v0.3]{
        \includegraphics[width=0.24\textwidth]{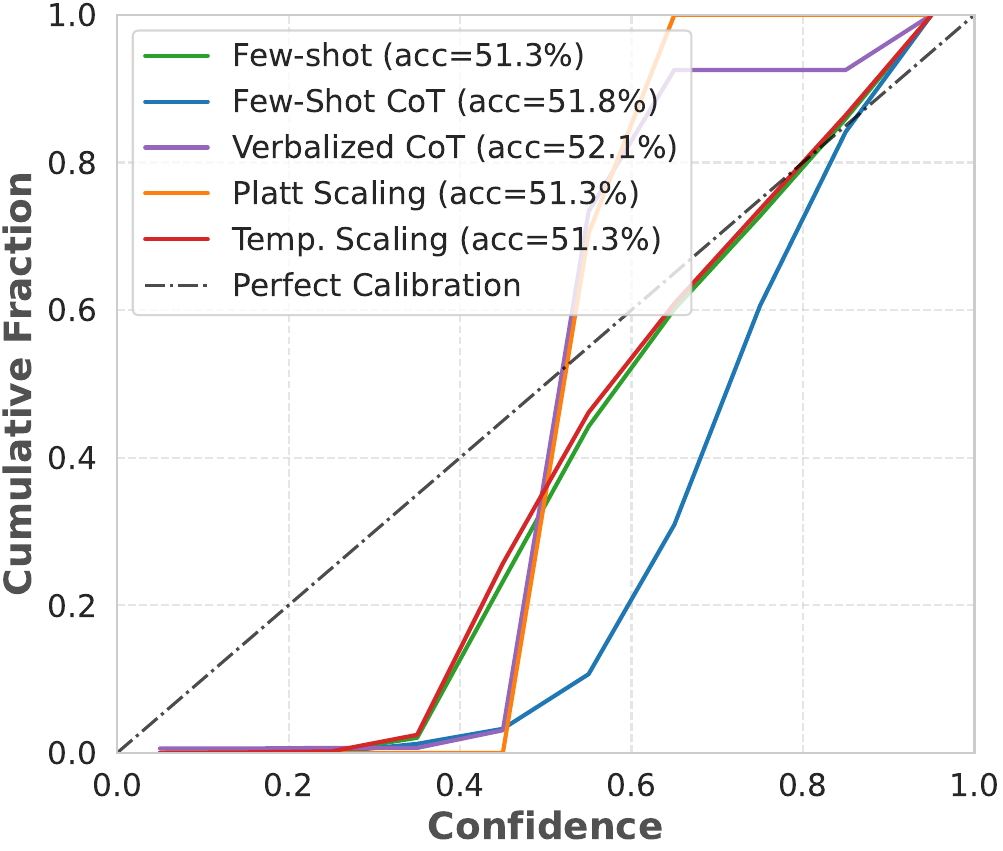}%
        \label{fig:verbalized_cot_qual_llama_triviaqa}
    }
    \\
    \subfloat[GPT-4]{
        \includegraphics[width=0.24\textwidth]{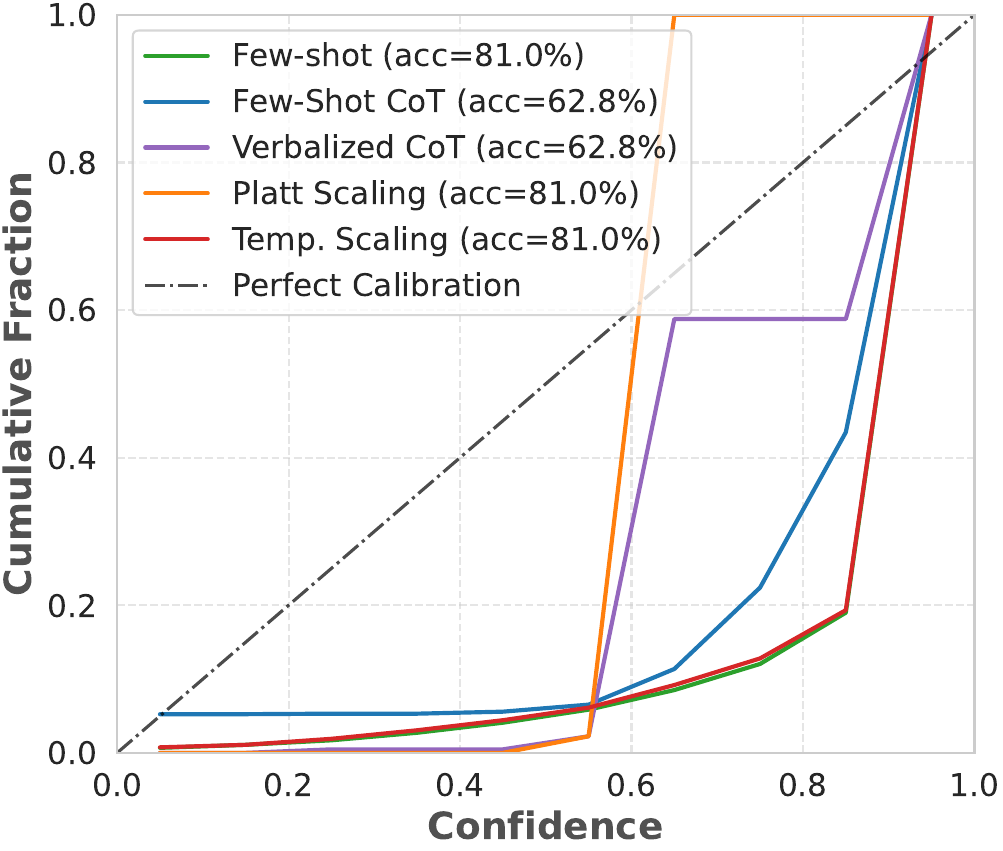}
        \label{fig:gpt4_triviaqa}
    }
    \subfloat[GPT-4o]{
        \includegraphics[width=0.24\textwidth]{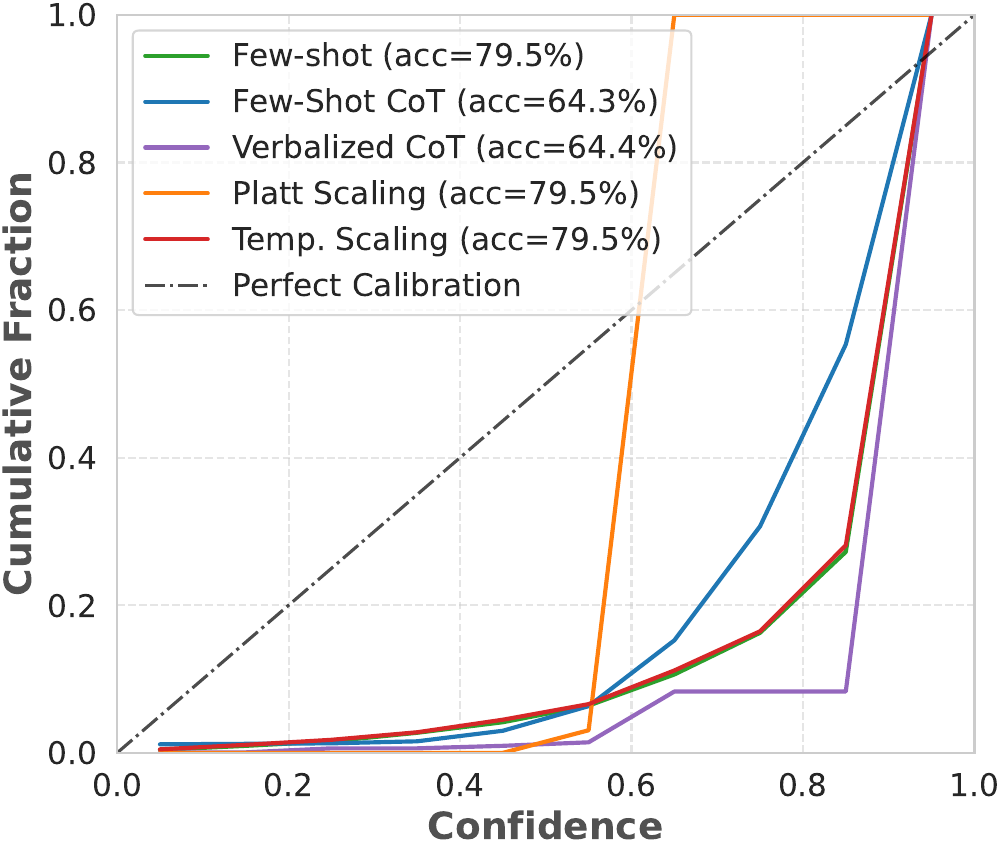}
        \label{fig:gpt4o_triviaqa}
    }
    \subfloat[GPT-5.2]{
        \includegraphics[width=0.24\textwidth]{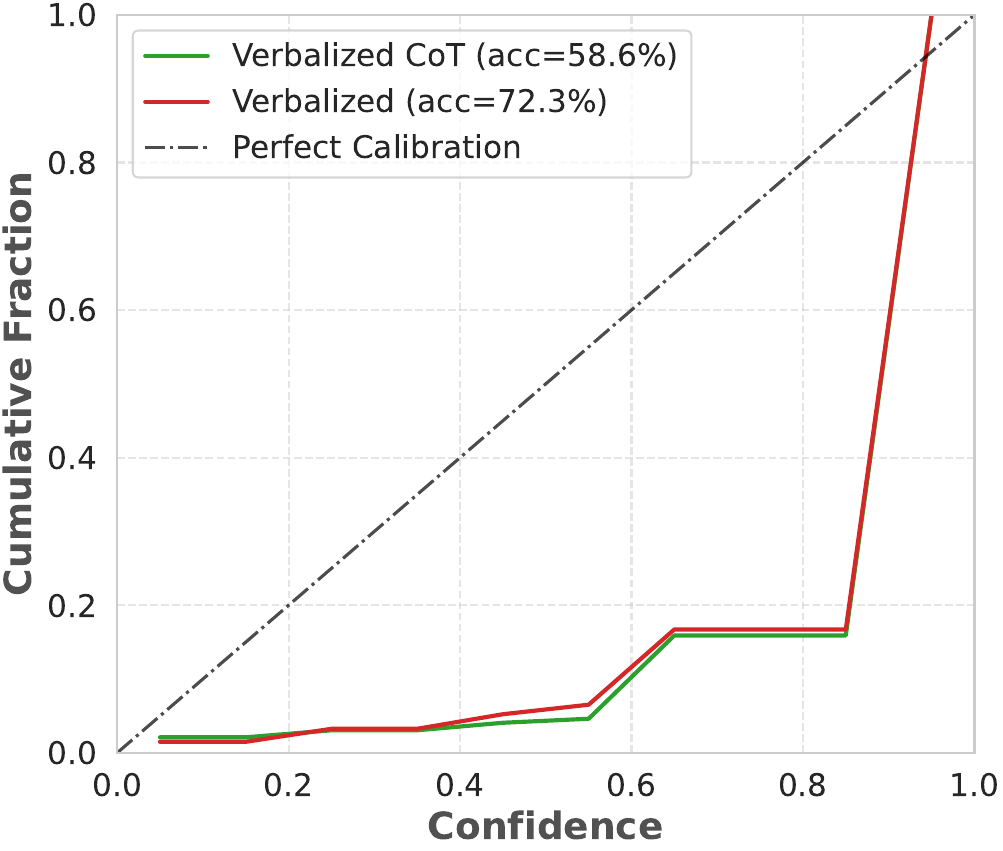}
        \label{fig:gpt5.2_triviaqa}
    }
    \subfloat[DeepSeek-R1]{
        \includegraphics[width=0.24\textwidth]{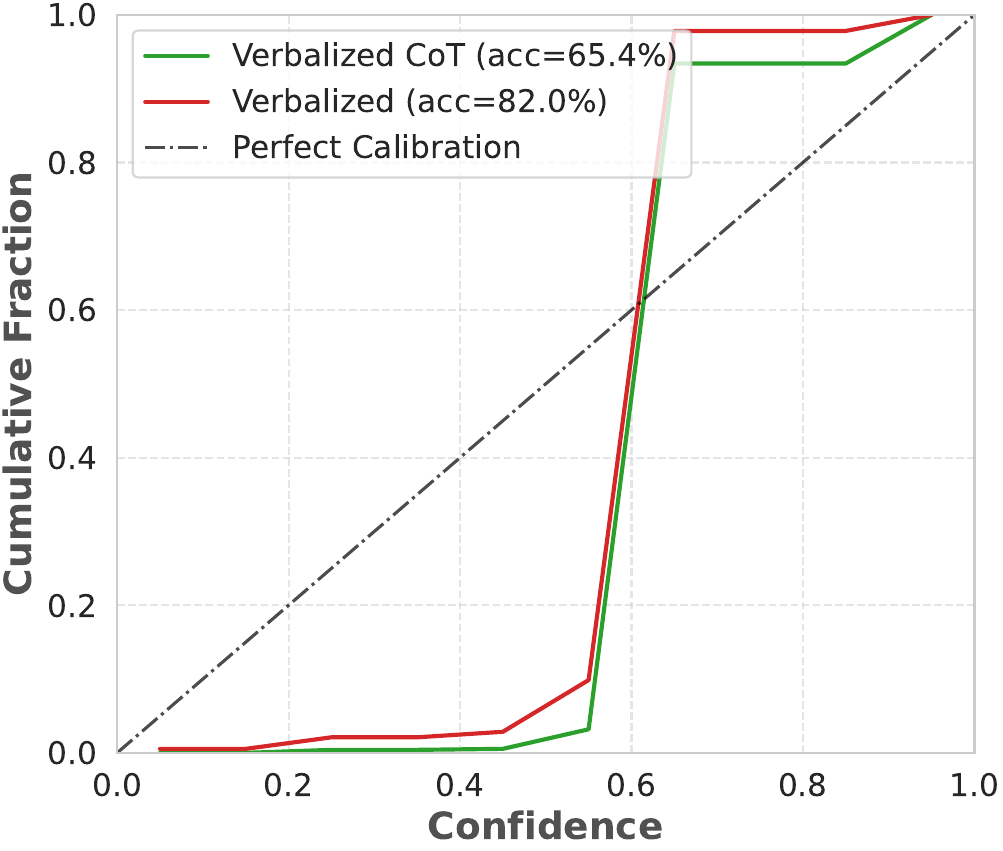}
        \label{fig:dsr1_triviaqa}
    }
    \caption{Reliability diagrams for calibration methods with 10 bins on TriviaQA. Top row: open-box models (Llama 3.1 8B, Qwen2.5 7B, Mistral v0.3 7B). Bottom row: closed-box models (GPT-4, GPT-4o, GPT-5.2, DeepSeek-R1).}
    \label{fig:llama_trivia_qa_comparison}
    \label{fig:gpt4_trivia_qa_comparison}
\end{figure*}

\section{RQ4: Research Challenges}
\label{sec:emerging_topics}

\subsection{Discussion}
\label{sec:discussion}

\reviewerOne{Calibration interacts with downstream reasoning: uncertainty-aware strategies such as selective abstention and confidence-gated generation reduce hallucination but introduce a trade-off between creativity and reliability depending on threshold choice. Our experiments suggest that pairing CoT prompting with verbalized uncertainty improves AUROC without degrading BLEU or exact-match, indicating that structured reasoning traces yield more reliable confidence signals. Optimal threshold selection and adaptive confidence gating based on task difficulty, domain, or user tolerance remain open practical challenges.}

We next discuss our evaluation's findings and review the limitations and relevance. 

\noindent{\bf Takeaways.} We observe that most existing methods for calibrating LLMs are designed primarily for textual and semantic classification tasks and lack scalability for generative tasks \cite{guo2017calibration, zhou2023batchcalibration}. Addressing the applicability of these methods to generative natural language tasks represents a significant research challenge. Recent studies have utilized small auxiliary models to elicit confidence and cluster questions and answers using sentence embedding models to achieve target accuracy \cite{ulmer2024apricot}. However, we contend that such auxiliary models may be both unreliable and computationally inefficient when applied to diverse tasks with LLMs. Instead, ranking questions and answers present a more practical and scalable alternative. Additionally, the effectiveness of eliciting and improving model confidence is significantly influenced by the quality and relevance of the questions, underscoring the importance of reasoning capabilities in LLMs for providing more reliable confidence assessments across various tasks \cite{guo2025deepseekrone}.

\noindent{\bf Lack of intuitive metric.} In addition, not all evaluation metrics are suitable for calibration tasks and have limitations in capturing different aspects of uncertainty and calibratedness of the model. The BLEU score cannot measure the semantics and syntactic correctness of the generation. It is also quite sensitive to the size of the generation and struggles to capture the validity of results that are derived from step-by-step expression as in CoT. As a reference-based metric \cite{zhao2023slichf}, it does not provide a measure of generation quality other than the provided reference answers. We think a robust evaluation metric is still an open question in the LLM literature. Selection bias can be mitigated using calibration \cite{zheng2023llmselbiaspride}; hence, proposing a metric for bias and including it in the supervised fine-tuning stage may be necessary for further research.

\noindent{\bf Better binning for ECE.} We observe that a smoothed version of ECE can provide a more concrete error value. As noted in previous studies \cite{becker2024cyclesofthought, ulmer2024apricot}, ECE has empirically been shown to measure general uncertainty that may not contain all uncertainty types and is quite sensitive toward pre-defined ranges or bins \cite{becker2024cyclesofthought}. We also observe that ECE fails to capture diverse types of uncertainty that LLM exhibits and can be an unstable metric \cite{huang2024ulmrankcalib}.  

\subsection{Future Research Directions}
\label{sec:future_research}

\reviewerOne{Scalable deployment of uncertainty-aware LLMs requires balancing estimation quality and cost. Sampling-based methods (e.g., deep ensembles, MC dropout) are prohibitive at LLM scale, while single-forward-pass alternatives such as verbalized uncertainty, conformal prediction, and post-hoc calibration add negligible latency. Online temperature-scaling updates can mitigate distribution drift, and hybrid architectures that route only low-confidence outputs to costlier verification (e.g., ensemble or consistency checks) help balance coverage, latency, and compute.}

\reviewerOne{\noindent{\bf Representing, updating, and reasoning about beliefs.} A broader open question is how LLMs can better represent, update, and reason about their own beliefs. Representations beyond miscalibrated next-token distributions, e.g., semantic-entropy aggregation~\cite{kuhn2023semanticuncertainty,farquhar2024detecting}, activation-level probes~\cite{azaria2023saplma}, and verbalized qualitative confidence~\cite{tian2023justaskforcalibration}, are promising. Belief updating calls for online or few-shot recalibration~\cite{li2024fewshotrecalibrator} and uncertainty-gated self-correction~\cite{huang2023llmselfcorrect,madaan2024selfrefine}, while reasoning-tuned models invite auditing trace faithfulness~\cite{xu2024sayself} and selective abstention~\cite{wang2025sconu,feng2024don}. Our NQ findings support CoT + verbalized qualitative confidence for ambiguous open-domain QA; a jointly faithful, calibrated, and deployment-time updatable intrinsic uncertainty signal remains open.}

In this section, we summarize research directions that can impact LLMs' confidence and are necessary to investigate further. We outline  avenues as follows:

\noindent{\bf Novel method, metrics, or loss components.} UQ methods can be used with LLMs to abstain from answer generation for questions or prompts of incorrect domain \cite{schuster2022confident}. Token-wise uncertainty quantification may not be efficient for longer sequences, as they only provide the likelihood of the next token without contextual insight \cite{xiong2023canllmexpress}. Entropy rates, which are obtained by averaging over the sequence, can be used to calibrate disagreements of long-term generations of LLMs \cite{Braverman2019CalibrationER}. Hence, preserving the semantic aspect of a sequence, calibration of longer sequences, and sequence-level confidence elicitation is an open research question for the future. Optimizing models with their uncertainty and accuracy can provide better-calibrated and improved confidence \cite{krishnan2020improving}. Such loss metrics can also be combined with TS to be robust for distributional shifts \cite{krishnan2020improving,yuksekgonul2024atypicalitycalib}. Following this venue, proposing an uncertainty-aware loss that can be used to fine-tune large or pre-train models is important. One example can be using the Kolmogorov-Smirnov (KS) similarity for recalibration \cite{gupta2021calibsplinefitting}. An uncertainty-aware loss function can also provide consistency of uncertainty estimation \cite{danruo2023uncertainty}. Focal loss can be adapted for fine-tuning of models, which can make LLM better calibrated and detect OOD samples accurately compared to TS \cite{mukhoti2020calibratingwithfocalloss}. Manifold smoothing can regularize LLMs to be more calibrated toward OOD samples \cite{kong-etal-2020-calibrated-lm-fine-tuning}. Open question: Is uncertainty toward OOD samples the same as epistemic uncertainty \cite{chua2023tacklinguncert}?

\noindent{\bf Advanced reasoning to calibrate user input ambiguity.} AI agents can converse for improved confidence and a better approach to uncertainty. Recent work \cite{ulmer2024apricot} found that the model could learn from self-criticism measured by confidence scores from the auxiliary model or from verbally expressed uncertainty to improve the ECE of LLMs. Meanwhile, input ambiguity provided by the user is another form of UQ, which is an open challenge \cite{han2024towardsuala}. Inputs in the form of short or obvious questions might be impolite for users, and calibration should consider that aspect \cite{stengel2024lacie}. Atypicality of LLM prompts and consistency-based methods can be useful to detect ambiguous user queries \cite{yuksekgonul2024atypicalitycalib, zhao2023selfdetectionanswerclustering}. Contrastive learning-based methods can be adapted to detect input ambiguity \cite{kirchhof2023probabilistic, he2023clur}.

\noindent{\bf Factuality and verification of LLMs.} Models show the limitation to generate consistently but incorrectly \cite{zhao2023selfdetectionanswerclustering, kuleshov2015calibrated}. Factual information that LLM internal-state contains can be calibrated with supporting pieces of evidence \cite{dong2022calinet, cohen2023lmvslm, liu2024activationconfcalibcodec, zhao2024far, fadeeva2024ccp}. Using external knowledge bases or verifier models for reflection can further improve LLM generations \cite{zhao2023selfdetectionanswerclustering}.

\noindent{\bf Faithfulness and uncertainty of reasoning text.} While modern LLMs contain advanced reasoning stages as part of their RL-based fine-tuning process \cite{guo2025deepseekrone}, faithful and hallucination-free reasoning steps are crucial\cite{xu2024sayself}. Hence, quantifying and calibrating reasoning text can also be as important \cite{liu2024uait}.

\noindent{\bf Self-correction strategy for improved performance of LLMs.} Although without access to external feedback, LLM self-correction gets degraded accuracy, and such feedback can be biased and misleading for the model \cite{huang2023llmselfcorrect}. One approach is using uncertainty-aware self-correction with ICL and an ensemble uncertainty score and self-correcting at test time \cite{yang2023improvingiclselfcorr}. In \cite{welleck2022generatingselfcorrect}, authors use the generator model and corrector with value pairing of generations for online training. LLMs can self-refine their generations with an iterative self-feedback \cite{madaan2024selfrefine}. Self-correction can also remove harmful content and bias of generation \cite{ganguli2023moralselfcorr}. \edit{In \cite{chensanitizingretrol}, they propose retrospective learning (RetroL) to improve self-correction of data annotation using the metric of consistency between student and teacher models \cite{chen2025evaluatingcai}}.

\noindent{\bf Generative calibration.} Recent methods focused on calibrating classification tasks, such as the multi-choice QA task \cite{becker2024cyclesofthought}. The application of calibration methods on text generation in the future should be further explored \cite{zhou2023batchcalibration} since calibration of the generation task for long sentence sequences can be hard for LLM to achieve by focusing on early output tokens. Contextual calibration methods, which are mostly applied to the semantic and text classification, lack the integration to the generative task \cite{zhou2023batchcalibration}, while scalability and efficiency to generative calibration is an important research direction. Majority consensus prompting for self-consistency for open-ended generation tasks is also not efficient for long-form generations \cite{chen2023universalselfconsistency}.

\noindent \textbf{Hallucination of LLMs}. \reviewerOne{One study revealed that \cite{xiao2021hallucinationanduncertainty} quantifying epistemic uncertainty can indicate hallucination better. Calibrated hallucination detectors \cite{guerreiro2022optimalhallucinationdet} that use Wasserstein distance to measure anomaly between generated and ground-truth sequences can help prevent hallucinating. Semantic entropy 
\cite{farquhar2024detecting} and iterative grouped linear binning \cite{detommaso2024multicalibrationiglb} can be used to detect hallucinations. Can we use cross-domain knowledge or different modalities for a textual explanation of output uncertainty?}

\noindent \textbf{Explainable uncertainty estimation}. Uncertainty estimation of probabilistic embeddings can be combined with large latent diffusion models, such as Stable Diffusion \cite{rombach2022stablediffusion}, to improve the interpretability of multi-modal outputs. Concept-relevant propagation (CRP) can also be adapted for LLMs to provide human-understandable explanations \cite{achtibat2023attribution}. In addition, mechanistic interpretability methods can be used to model uncertainty by examining how internal model activations are affected by MetaFaith and other prompting techniques \cite{liu2025metafaith, zelenka2023explainableue, kailkhura2019reliablex}. Can XAI methods be applied to predict uncertainty?

\noindent \textbf{Real-time epistemic uncertainty for LLMs.} Adapting existing UQ methods to detect epistemic semantic uncertainty \cite{kuhn2023semanticuncertainty, Nikitin2024KernelLE} in real-time for LLMs can prevent hazardous user experience \cite{kendallgal2017whatuncertaintiesweneed}.

\noindent \textbf{Ground-truth uncertainty estimates}. Ground-truth estimate of uncertainty might depend on the specified task: multi-choice questions, NMT, and generative tasks. It can be hard to obtain ground truths as a model can be miscalibrated according to different subsets of the model, or bias can be different for certain categories \cite{jiang2021qalmcalibration}. Is it necessary to have a human-defined threshold for the uncertainty of LFMs? 

\noindent \textbf{Classical approaches for LLMs}. 
Whether methods for numerical calibration are effective for linguistic calibration recently gained attention \cite{zhou2023navigating}. One common consideration for linguistic calibration is that NLP models can provide quite structured inference results \cite{nguyen2015posteriorcal}. Dirichlet calibration \cite{kull2019beyond, wang2023uncertretinal}, which is a multi-class extension of Beta calibration \cite{kull2017betacalib} that estimates Dirichlet distribution for classes, can be applied to LLMs for de-biasing their confidence toward categories or sensitive groups. Frameworks for prior networks, such as DPN \cite{malinin2018predictivepriornetworks}, can further be extended to LLMs and NLP tasks to capture distributional uncertainty that could exist between train and test datasets. In \cite{sankararaman2022bayesformer}, the authors proposed a dropout variational inference model for transformer architecture. In addition, adapting few-shot recalibration for RLHF and o1/R1-based models \cite{li2024fewshotrecalibrator} and Cloze distillation \cite{eisape2020clozedist} sounds like promising future work.

\noindent \textbf{Uncertainty of multi-modal large language models (MLLMs)}. \reviewerOne{Uncertainty estimation methods can estimate embedding distribution for deterministic LVLMs and, combined with calibration methods, can be applied to the selection of models and activate learning \cite{upadhyay2023probvlm}. Another important application for UQ and calibration methods in image-text matching (ITM), where uncertainty roots from unclear annotations and multiple matching images and captions \cite{chun2023pcmeplusplus}. Such textual uncertainty can also be used as a soft label for the loss function objective.}

\noindent \textbf{Calibration in Domain Adaptation (DA).} Calibration of DA methods has problems to address \cite{wang2020transcal}. One of the existing works for calibration on dataset shift, the TransCal method optimizes transferable calibration objective to get the parameters for calibration of model confidence for predictions on target domain \cite{wang2020transcal}. Calibration in OOD, domain shift, and distributional shift. Teaching models to express uncertainty with domain shift \cite{lin2022teachingmodelstoexpress}. Can we apply the pre-trained visual uncertainty method \cite{Kirchhof2024PretrainedVU} for language tasks?

\noindent \textbf{Reliable contribution incentivization and Blockchain.} Blockchain LLMs \cite{gai2023blockchainllm} as implemented as reward mechanism for preference optimization \cite{rafailov2023dpo}. Fine-tuning is an important stage for generating reasonable inferences from LLMs, as training from scratch might take a long time. While understanding the trending and important methods for fine-tuning LFMs, such as LoRA \cite{hu2021lora}, is crucial, the fine-tuning techniques of LFMs will be highly related to the contribution one can make via one’s local data. The reliability of non-learning methods is an interesting research avenue.

\noindent{\bf Code and Dataset Availability.} Lastly, reviewed studies hesitate to share their code and data availability. At the same time, such information can help to accelerate the research process and improve the implementation of prior methods \cite{gawlikowski2023survey}. Some datasets are proposed for truthful question answering and expression of unknown \cite{lin2021truthfulqa, yin2023llmknowwhattheydont}. We observe that there is a lack of calibration-specific datasets for unbiased reliability evaluation of LLMs. Proposing more representative datasets for benchmarking and reducing the uncertainty and hallucination of LLMs is important.

\section{Conclusion}
\label{sec:conclusion}

We reviewed a representative number of calibration methods and evaluation metrics for the reliability of large language models. We also evaluated seven top-level open-box and closed-box LLMs and showed which metrics better represent the quality of uncertainty for generative QA experiments on well-known datasets. We hope our work sheds light on the reliability of modern LLMs and methods to address their inherent uncertainty or provide better representation. We also highlighted and provided recommendations for future research directions and challenges for the community to facilitate.

\revision{Looking forward, we emphasize the practical Pareto framing developed in Section~\ref{sec:future_research}: reliable uncertainty at LLM scale is fundamentally a trade-off between estimation quality and deployment cost, and sampling-heavy techniques such as deep ensembles or MC dropout sit on the wrong side of that frontier. Single-forward-pass alternatives, verbalized qualitative confidence, conformal prediction, and post-hoc calibration, occupy the favorable region and can be augmented with online temperature-scaling updates to absorb distribution drift. Hybrid architectures that route only low-confidence outputs to costlier verifiers offer the clearest near-term path to production-grade uncertainty-aware LLMs.}

\section*{Acknowledgments}

We thank Prof. Li Dan from Tsinghua University and his researcher group for their supervision and support. Special thanks for Dr. Deng Xiang and Dr. Li Yan, engineers from Siemens Research China, for their valuable discussions and support.



\begingroup
\linespread{0.90}\selectfont
\bibliographystyle{IEEEtran}
\bibliography{ref}

@article{page2021prisma,
  title={PRISMA 2020 explanation and elaboration: updated guidance and exemplars for reporting systematic reviews},
  author={Page, Matthew J and Moher, David and Bossuyt, Patrick M and Boutron, Isabelle and Hoffmann, Tammy C and Mulrow, Cynthia D and Shamseer, Larissa and Tetzlaff, Jennifer M and Akl, Elie A and Brennan, Sue E and others},
  journal={bmj},
  volume={372},
  year={2021},
  publisher={British Medical Journal Publishing Group}
}

@article{kitchenham2004procedures,
  title={Procedures for performing systematic reviews},
  author={Kitchenham, Barbara},
  journal={Keele, UK, Keele University},
  volume={33},
  number={2004},
  pages={1--26},
  year={2004},
  publisher={Citeseer}
}

@article{wenchong2026asurveyonuqmethodsfordl,
author = {He, Wenchong and Jiang, Zhe and Xiao, Tingsong and Xu, Zelin and Li, Yukun},
title = {A Survey on Uncertainty Quantification Methods for Deep Learning},
year = {2026},
issue_date = {May 2026},
publisher = {Association for Computing Machinery},
address = {New York, NY, USA},
volume = {58},
number = {7},
issn = {0360-0300},
url = {https://doi.org/10.1145/3786319},
doi = {10.1145/3786319},
journal = {ACM Comput. Surv.},
month = feb,
articleno = {179},
numpages = {35}
}

@inproceedings{xia2025asurveyofuellms,
  title={A survey of uncertainty estimation methods on large language models},
  author={Xia, Zhiqiu and Xu, Jinxuan and Zhang, Yuqian and Liu, Hang},
  booktitle={Findings of the Association for Computational Linguistics: ACL 2025},
  pages={21381--21396},
  year={2025}
}

@article{shorinwa2025surveyuqllm,
  title={A survey on uncertainty quantification of large language models: Taxonomy, open research challenges, and future directions},
  author={Shorinwa, Ola and Mei, Zhiting and Lidard, Justin and Ren, Allen Z and Majumdar, Anirudha},
  journal={ACM Computing Surveys},
  year={2025},
  publisher={ACM New York, NY}
}

@article{yang2024generalizedoodsurvey,
  title={Generalized out-of-distribution detection: A survey},
  author={Yang, Jingkang and Zhou, Kaiyang and Li, Yixuan and Liu, Ziwei},
  journal={International Journal of Computer Vision},
  pages={1--28},
  year={2024},
  publisher={Springer}
}

@inproceedings{geng2024surveyllmcal,
  title={A Survey of Confidence Estimation and Calibration in Large Language Models},
  author={Geng, Jiahui and Cai, Fengyu and Wang, Yuxia and Koeppl, Heinz and Nakov, Preslav and Gurevych, Iryna},
  booktitle={Proceedings of the 2024 Conference of the North American Chapter of the Association for Computational Linguistics: Human Language Technologies (Volume 1: Long Papers)},
  pages={6577--6595},
  year={2024}
}

@article{chang2024surveyllmevaluation,
  title={A survey on evaluation of large language models},
  author={Chang, Yupeng and Wang, Xu and Wang, Jindong and Wu, Yuan and Yang, Linyi and Zhu, Kaijie and Chen, Hao and Yi, Xiaoyuan and Wang, Cunxiang and Wang, Yidong and others},
  journal={ACM Transactions on Intelligent Systems and Technology},
  volume={15},
  number={3},
  pages={1--45},
  year={2024},
  publisher={ACM New York, NY}
}

@article{ji2023surveyofhallucinationinnlg,
  title={Survey of hallucination in natural language generation},
  author={Ji, Ziwei and Lee, Nayeon and Frieske, Rita and Yu, Tiezheng and Su, Dan and Xu, Yan and Ishii, Etsuko and Bang, Ye Jin and Madotto, Andrea and Fung, Pascale},
  journal={ACM computing surveys},
  volume={55},
  number={12},
  pages={1--38},
  year={2023},
  publisher={ACM New York, NY}
}

@article{baan2023uncertaintyfromtheorytoapplications,
  title={Uncertainty in natural language generation: From theory to applications},
  author={Baan, Joris and Daheim, Nico and Ilia, Evgenia and Ulmer, Dennis and Li, Haau-Sing and Fern{\'a}ndez, Raquel and Plank, Barbara and Sennrich, Rico and Zerva, Chrysoula and Aziz, Wilker},
  journal={arXiv preprint arXiv:2307.15703},
  year={2023}
}

@article{liu2023trustworthy,
  title={Trustworthy LLMs: a Survey and Guideline for Evaluating Large Language Models' Alignment},
  author={Liu, Yang and Yao, Yuanshun and Ton, Jean-Francois and Zhang, Xiaoying and Cheng, Ruocheng Guo Hao and Klochkov, Yegor and Taufiq, Muhammad Faaiz and Li, Hang},
  journal={arXiv preprint arXiv:2308.05374},
  year={2023}
}

@article{gawlikowski2023survey,
  title={A survey of uncertainty in deep neural networks},
  author={Gawlikowski, Jakob and Tassi, Cedrique Rovile Njieutcheu and Ali, Mohsin and Lee, Jongseok and Humt, Matthias and Feng, Jianxiang and Kruspe, Anna and Triebel, Rudolph and Jung, Peter and Roscher, Ribana and others},
  journal={Artificial Intelligence Review},
  volume={56},
  number={Suppl 1},
  pages={1513--1589},
  year={2023},
  publisher={Springer}
}

@article{xiao2022uqwithplm,
  title={Uncertainty quantification with pre-trained language models: A large-scale empirical analysis},
  author={Xiao, Yuxin and Liang, Paul Pu and Bhatt, Umang and Neiswanger, Willie and Salakhutdinov, Ruslan and Morency, Louis-Philippe},
  journal={arXiv preprint arXiv:2210.04714},
  year={2022}
}

@article{zhang2021modern,
  title={Modern Monte Carlo methods for efficient uncertainty quantification and propagation: A survey},
  author={Zhang, Jiaxin},
  journal={Wiley Interdisciplinary Reviews: Computational Statistics},
  volume={13},
  number={5},
  pages={e1539},
  year={2021},
  publisher={Wiley Online Library}
}

@article{hendrycks2021unsolved,
  title={Unsolved problems in ml safety},
  author={Hendrycks, Dan and Carlini, Nicholas and Schulman, John and Steinhardt, Jacob},
  journal={arXiv preprint arXiv:2109.13916},
  year={2021}
}

@article{kendallgal2017whatuncertaintiesweneed,
  title={What uncertainties do we need in bayesian deep learning for computer vision?},
  author={Kendall, Alex and Gal, Yarin},
  journal={Advances in neural information processing systems},
  volume={30},
  year={2017}
}

@article{ye2024benchmarkingllmuq,
  title={Benchmarking llms via uncertainty quantification},
  author={Ye, Fanghua and Yang, Mingming and Pang, Jianhui and Wang, Longyue and Wong, Derek and Yilmaz, Emine and Shi, Shuming and Tu, Zhaopeng},
  journal={Advances in Neural Information Processing Systems},
  volume={37},
  pages={15356--15385},
  year={2024}
}

@article{wang2024subjectiveuqcal,
  title={On Subjective Uncertainty Quantification and Calibration in Natural Language Generation},
  author={Wang, Ziyu and Holmes, Chris},
  journal={arXiv preprint arXiv:2406.05213},
  year={2024}
}

@article{Kalai2023CalibratedLMmustHallucinate,
  title={Calibrated Language Models Must Hallucinate},
  author={Adam Tauman Kalai and Santosh S. Vempala},
  journal={Proceedings of the 56th Annual ACM Symposium on Theory of Computing},
  year={2023},
  url={https://api.semanticscholar.org/CorpusID:265445593}
}

@inproceedings{kim2024uncertain1st,
  title={" I'm Not Sure, But...": Examining the Impact of Large Language Models' Uncertainty Expression on User Reliance and Trust},
  author={Kim, Sunnie SY and Liao, Q Vera and Vorvoreanu, Mihaela and Ballard, Stephanie and Vaughan, Jennifer Wortman},
  booktitle={The 2024 ACM Conference on Fairness, Accountability, and Transparency},
  pages={822--835},
  year={2024}
}

@inproceedings{chen2023plmcal,
  title={A Close Look into the Calibration of Pre-trained Language Models},
  author={Chen, Yangyi and Yuan, Lifan and Cui, Ganqu and Liu, Zhiyuan and Ji, Heng},
  booktitle={Proceedings of the 61st Annual Meeting of the Association for Computational Linguistics (Volume 1: Long Papers)},
  pages={1343--1367},
  year={2023}
}

@article{chua2023tacklinguncert,
  title={Tackling prediction uncertainty in machine learning for healthcare},
  author={Chua, Michelle and Kim, Doyun and Choi, Jongmun and Lee, Nahyoung G and Deshpande, Vikram and Schwab, Joseph and Lev, Michael H and Gonzalez, Ramon G and Gee, Michael S and Do, Synho},
  journal={Nature Biomedical Engineering},
  volume={7},
  number={6},
  pages={711--718},
  year={2023},
  publisher={Nature Publishing Group UK London}
}

@article{huang2023lookbeforeyouleap,
  title={Look before you leap: An exploratory study of uncertainty measurement for large language models},
  author={Huang, Yuheng and Song, Jiayang and Wang, Zhijie and Zhao, Shengming and Chen, Huaming and Juefei-Xu, Felix and Ma, Lei},
  journal={arXiv preprint arXiv:2307.10236},
  year={2023}
}

@article{becker2024cyclesofthought,
  title={Cycles of Thought: Measuring LLM Confidence through Stable Explanations},
  author={Becker, Evan and Soatto, Stefano},
  journal={arXiv preprint arXiv:2406.03441},
  year={2024}
}

@article{xiong2023canllmexpress,
  title={Can llms express their uncertainty? an empirical evaluation of confidence elicitation in llms},
  author={Xiong, Miao and Hu, Zhiyuan and Lu, Xinyang and Li, Yifei and Fu, Jie and He, Junxian and Hooi, Bryan},
  journal={arXiv preprint arXiv:2306.13063},
  year={2023}
}

@inproceedings{feng2024don,
  title={Don’t Hallucinate, Abstain: Identifying LLM Knowledge Gaps via Multi-LLM Collaboration},
  author={Feng, Shangbin and Shi, Weijia and Wang, Yike and Ding, Wenxuan and Balachandran, Vidhisha and Tsvetkov, Yulia},
  booktitle={Proceedings of the 62nd Annual Meeting of the Association for Computational Linguistics (Volume 1: Long Papers)},
  pages={14664--14690},
  year={2024}
}

@article{madhusudhan2024abstainprompt,
  title={Do llms know when to not answer? investigating abstention abilities of large language models},
  author={Madhusudhan, Nishanth and Madhusudhan, Sathwik Tejaswi and Yadav, Vikas and Hashemi, Masoud},
  journal={arXiv preprint arXiv:2407.16221},
  year={2024}
}

@article{zablotskaia2023uncertaintybenchmark,
  title={On uncertainty calibration and selective generation in probabilistic neural summarization: A benchmark study},
  author={Zablotskaia, Polina and Phan, Du and Maynez, Joshua and Narayan, Shashi and Ren, Jie and Liu, Jeremiah},
  journal={arXiv preprint arXiv:2304.08653},
  year={2023}
}

@article{zhu2023calibrationandalignment,
  title={On the calibration of large language models and alignment},
  author={Zhu, Chiwei and Xu, Benfeng and Wang, Quan and Zhang, Yongdong and Mao, Zhendong},
  journal={arXiv preprint arXiv:2311.13240},
  year={2023}
}

@article{tian2023justaskforcalibration,
  title={Just ask for calibration: Strategies for eliciting calibrated confidence scores from language models fine-tuned with human feedback},
  author={Tian, Katherine and Mitchell, Eric and Zhou, Allan and Sharma, Archit and Rafailov, Rafael and Yao, Huaxiu and Finn, Chelsea and Manning, Christopher D},
  journal={arXiv preprint arXiv:2305.14975},
  year={2023}
}

@article{guerreiro2022optimalhallucinationdet,
  title={Optimal transport for unsupervised hallucination detection in neural machine translation},
  author={Guerreiro, Nuno M and Colombo, Pierre and Piantanida, Pablo and Martins, Andr{\'e} FT},
  journal={arXiv preprint arXiv:2212.09631},
  year={2022}
}

@article{chensanitizingretrol,
  title={Sanitizing LLMs: Retrospective Learning for Self-Correction of Inconsistent Samples via User Preferences},
  author={Chen, Cheng and Tsang, Ivor}
}

@article{ganguli2023moralselfcorr,
  title={The capacity for moral self-correction in large language models},
  author={Ganguli, Deep and Askell, Amanda and Schiefer, Nicholas and Liao, Thomas I and Luko{\v{s}}i{\=u}t{\.e}, Kamil{\.e} and Chen, Anna and Goldie, Anna and Mirhoseini, Azalia and Olsson, Catherine and Hernandez, Danny and others},
  journal={arXiv preprint arXiv:2302.07459},
  year={2023}
}

@article{welleck2022generatingselfcorrect,
  title={Generating sequences by learning to self-correct},
  author={Welleck, Sean and Lu, Ximing and West, Peter and Brahman, Faeze and Shen, Tianxiao and Khashabi, Daniel and Choi, Yejin},
  journal={arXiv preprint arXiv:2211.00053},
  year={2022}
}

@article{huang2023llmselfcorrect,
  title={Large language models cannot self-correct reasoning yet},
  author={Huang, Jie and Chen, Xinyun and Mishra, Swaroop and Zheng, Huaixiu Steven and Yu, Adams Wei and Song, Xinying and Zhou, Denny},
  journal={arXiv preprint arXiv:2310.01798},
  year={2023}
}

@inproceedings{widmann2021multiclasscalkernel,
  title={Calibration tests beyond classification},
  author={Widmann, David and Lindsten, Fredrik and Zachariah, Dave},
  booktitle={International Conference on Learning Representations, Virtual conference, May 3-May 7, 2021},
  pages={1--37},
  year={2021}
}

@article{jiang2021qalmcalibration,
  title={How can we know when language models know? on the calibration of language models for question answering},
  author={Jiang, Zhengbao and Araki, Jun and Ding, Haibo and Neubig, Graham},
  journal={Transactions of the Association for Computational Linguistics},
  volume={9},
  pages={962--977},
  year={2021},
  publisher={MIT Press One Rogers Street, Cambridge, MA 02142-1209, USA journals-info~…}
}

@article{desai2020calibrationprtransformers,
  title={Calibration of pre-trained transformers},
  author={Desai, Shrey and Durrett, Greg},
  journal={arXiv preprint arXiv:2003.07892},
  year={2020}
}

@inproceedings{nixon2019measuringcalibrationindl,
  title={Measuring Calibration in Deep Learning.},
  author={Nixon, Jeremy and Dusenberry, Michael W and Zhang, Linchuan and Jerfel, Ghassen and Tran, Dustin},
  booktitle={CVPR workshops},
  volume={2},
  number={7},
  year={2019}
}

@article{laves2019well,
  title={Well-calibrated model uncertainty with temperature scaling for dropout variational inference},
  author={Laves, Max-Heinrich and Ihler, Sontje and Kortmann, Karl-Philipp and Ortmaier, Tobias},
  journal={arXiv preprint arXiv:1909.13550},
  year={2019}
}

@article{laves2020calibration,
  title={Calibration of model uncertainty for dropout variational inference},
  author={Laves, Max-Heinrich and Ihler, Sontje and Kortmann, Karl-Philipp and Ortmaier, Tobias},
  journal={arXiv preprint arXiv:2006.11584},
  year={2020}
}

@article{ovadia2019can,
  title={Can you trust your model's uncertainty? evaluating predictive uncertainty under dataset shift},
  author={Ovadia, Yaniv and Fertig, Emily and Ren, Jie and Nado, Zachary and Sculley, David and Nowozin, Sebastian and Dillon, Joshua and Lakshminarayanan, Balaji and Snoek, Jasper},
  journal={Advances in neural information processing systems},
  volume={32},
  year={2019}
}

@article{nguyen2015posteriorcal,
  title={Posterior calibration and exploratory analysis for natural language processing models},
  author={Nguyen, Khanh and O'Connor, Brendan},
  journal={arXiv preprint arXiv:1508.05154},
  year={2015}
}

@article{walker2003defining,
  title={Defining uncertainty: a conceptual basis for uncertainty management in model-based decision support},
  author={Walker, Warren E and Harremo{\"e}s, Poul and Rotmans, Jan and Van Der Sluijs, Jeroen P and Van Asselt, Marjolein BA and Janssen, Peter and Krayer von Krauss, Martin P},
  journal={Integrated assessment},
  volume={4},
  number={1},
  pages={5--17},
  year={2003},
  publisher={Taylor \& Francis}
}

@inproceedings{niculescu2005predictingreldiag,
  title={Predicting good probabilities with supervised learning},
  author={Niculescu-Mizil, Alexandru and Caruana, Rich},
  booktitle={Proceedings of the 22nd international conference on Machine learning},
  pages={625--632},
  year={2005}
}

@article{li2023selfensemblesicl,
  title={On task performance and model calibration with supervised and self-ensembled in-context learning},
  author={Li, Chengzu and Zhou, Han and Glava{\v{s}}, Goran and Korhonen, Anna and Vuli{\'c}, Ivan},
  journal={arXiv preprint arXiv:2312.13772},
  year={2023}
}

@article{jiang2023cape,
  title={Calibrating language models via augmented prompt ensembles},
  author={Jiang, Mingjian and Ruan, Yangjun and Huang, Sicong and Liao, Saifei and Pitis, Silviu and Grosse, Roger Baker and Ba, Jimmy},
  year={2023}
}

@article{zheng2023llmselbiaspride,
  title={On Large Language Models' Selection Bias in Multi-Choice Questions},
  author={Zheng, Chujie and Zhou, Hao and Meng, Fandong and Zhou, Jie and Huang, Minlie},
  journal={arXiv preprint arXiv:2309.03882},
  year={2023}
}

@article{Blasiok2023LossMY,
  title={Loss minimization yields multicalibration for large neural networks},
  author={Jaroslaw Blasiok and Parikshit Gopalan and Lunjia Hu and Adam Tauman Kalai and Preetum Nakkiran},
  journal={ArXiv},
  year={2023},
  volume={abs/2304.09424},
  url={https://api.semanticscholar.org/CorpusID:258212539}
}

@inproceedings{hebert2018multicalibrationigbh,
  title={Multicalibration: Calibration for the (computationally-identifiable) masses},
  author={H{\'e}bert-Johnson, Ursula and Kim, Michael and Reingold, Omer and Rothblum, Guy},
  booktitle={International Conference on Machine Learning},
  pages={1939--1948},
  year={2018},
  organization={PMLR}
}

@inproceedings{gopalan2022lowdegreemulticalib,
  title={Low-degree multicalibration},
  author={Gopalan, Parikshit and Kim, Michael P and Singhal, Mihir A and Zhao, Shengjia},
  booktitle={Conference on Learning Theory},
  pages={3193--3234},
  year={2022},
  organization={PMLR}
}

@article{detommaso2024multicalibrationiglb,
  title={Multicalibration for confidence scoring in LLMs},
  author={Detommaso, Gianluca and Bertran, Martin and Fogliato, Riccardo and Roth, Aaron},
  journal={arXiv preprint arXiv:2404.04689},
  year={2024}
}

@article{holtzman2021dcpmi,
  title={Surface form competition: Why the highest probability answer isn't always right},
  author={Holtzman, Ari and West, Peter and Shwartz, Vered and Choi, Yejin and Zettlemoyer, Luke},
  journal={arXiv preprint arXiv:2104.08315},
  year={2021}
}

@article{zhou2023batchcalibration,
  title={Batch calibration: Rethinking calibration for in-context learning and prompt engineering},
  author={Zhou, Han and Wan, Xingchen and Proleev, Lev and Mincu, Diana and Chen, Jilin and Heller, Katherine and Roy, Subhrajit},
  journal={arXiv preprint arXiv:2309.17249},
  year={2023}
}

@inproceedings{roelofs2022domancontextcalib,
  title={Mitigating bias in calibration error estimation},
  author={Roelofs, Rebecca and Cain, Nicholas and Shlens, Jonathon and Mozer, Michael C},
  booktitle={International Conference on Artificial Intelligence and Statistics},
  pages={4036--4054},
  year={2022},
  organization={PMLR}
}

@article{han2022prototypicalcalib,
  title={Prototypical calibration for few-shot learning of language models},
  author={Han, Zhixiong and Hao, Yaru and Dong, Li and Sun, Yutao and Wei, Furu},
  journal={arXiv preprint arXiv:2205.10183},
  year={2022}
}

@inproceedings{kumar2022answerlevecalib,
  title={Answer-level calibration for free-form multiple choice question answering},
  author={Kumar, Sawan},
  booktitle={Proceedings of the 60th Annual Meeting of the Association for Computational Linguistics (Volume 1: Long Papers)},
  pages={665--679},
  year={2022}
}

@inproceedings{zhao2021contextualcalib,
  title={Calibrate before use: Improving few-shot performance of language models},
  author={Zhao, Zihao and Wallace, Eric and Feng, Shi and Klein, Dan and Singh, Sameer},
  booktitle={International conference on machine learning},
  pages={12697--12706},
  year={2021},
  organization={PMLR}
}

@article{claps,
  title={Survival of the most influential prompts: Efficient black-box prompt search via clustering and pruning},
  author={Zhou, Han and Wan, Xingchen and Vuli{\'c}, Ivan and Korhonen, Anna},
  journal={arXiv preprint arXiv:2310.12774},
  year={2023}
}

@inproceedings{xu2024sayself,
  title={Sayself: Teaching llms to express confidence with self-reflective rationales},
  author={Xu, Tianyang and Wu, Shujin and Diao, Shizhe and Liu, Xiaoze and Wang, Xingyao and Chen, Yangyi and Gao, Jing},
  booktitle={Proceedings of the 2024 Conference on Empirical Methods in Natural Language Processing},
  pages={5985--5998},
  year={2024}
}

@article{Yang2023Laplacelora,
  title={Bayesian low-rank adaptation for large language models},
  author={Adam X. Yang and Maxime Robeyns and Xi Wang and Laurence Aitchison},
  journal={ArXiv},
  year={2023},
  volume={abs/2308.13111},
  url={https://api.semanticscholar.org/CorpusID:261214713}
}

@inproceedings{eisape2020clozedist,
  title={Cloze distillation: Improving neural language models with human next-word prediction},
  author={Eisape, Tiwalayo and Zaslavsky, Noga and Levy, Roger},
  year={2020},
  organization={Association for Computational Linguistics (ACL)}
}

@inproceedings{band2024linguistic,
  title={Linguistic calibration of longform generations},
  author={Band, Neil and Li, Xuechen and Ma, Tengyu and Hashimoto, Tatsunori},
  booktitle={Forty-first International Conference on Machine Learning},
  year={2024}
}

@article{dong2022calinet,
  title={Calibrating factual knowledge in pretrained language models},
  author={Dong, Qingxiu and Dai, Damai and Song, Yifan and Xu, Jingjing and Sui, Zhifang and Li, Lei},
  journal={arXiv preprint arXiv:2210.03329},
  year={2022}
}

@inproceedings{kapoor2024calibrationtuning,
  title={Calibration-Tuning: Teaching Large Language Models to Know What They Don’t Know},
  author={Kapoor, Sanyam and Gruver, Nate and Roberts, Manley and Pal, Arka and Dooley, Samuel and Goldblum, Micah and Wilson, Andrew},
  booktitle={Proceedings of the 1st Workshop on Uncertainty-Aware NLP (UncertaiNLP 2024)},
  pages={1--14},
  year={2024}
}

@article{lee2022adaptivels,
  title={Adaptive label smoothing with self-knowledge in natural language generation},
  author={Lee, Dongkyu and Cheung, Ka Chun and Zhang, Nevin L},
  journal={arXiv preprint arXiv:2210.13459},
  year={2022}
}

@article{stengel2024lacie,
  title={LACIE: Listener-Aware Finetuning for Confidence Calibration in Large Language Models},
  author={Stengel-Eskin, Elias and Hase, Peter and Bansal, Mohit},
  journal={arXiv preprint arXiv:2405.21028},
  year={2024}
}

@inproceedings{zhang2024rtuning,
  title={R-Tuning: Instructing Large Language Models to Say ‘I Don’t Know’},
  author={Zhang, Hanning and Diao, Shizhe and Lin, Yong and Fung, Yi and Lian, Qing and Wang, Xingyao and Chen, Yangyi and Ji, Heng and Zhang, Tong},
  booktitle={Proceedings of the 2024 Conference of the North American Chapter of the Association for Computational Linguistics: Human Language Technologies (Volume 1: Long Papers)},
  pages={7106--7132},
  year={2024}
}

@article{zhao2021calibratingpredtodecisions,
  title={Calibrating predictions to decisions: A novel approach to multi-class calibration},
  author={Zhao, Shengjia and Kim, Michael and Sahoo, Roshni and Ma, Tengyu and Ermon, Stefano},
  journal={Advances in Neural Information Processing Systems},
  volume={34},
  pages={22313--22324},
  year={2021}
}

@article{schuster2022confident,
  title={Confident adaptive language modeling},
  author={Schuster, Tal and Fisch, Adam and Gupta, Jai and Dehghani, Mostafa and Bahri, Dara and Tran, Vinh and Tay, Yi and Metzler, Donald},
  journal={Advances in Neural Information Processing Systems},
  volume={35},
  pages={17456--17472},
  year={2022}
}

@article{balabanov2024loraensemble,
  title={Uncertainty quantification in fine-tuned LLMs using LoRA ensembles},
  author={Balabanov, Oleksandr and Linander, Hampus},
  journal={arXiv preprint arXiv:2402.12264},
  year={2024}
}

@article{lu2022learningconftransformernmt,
  title={Learning confidence for transformer-based neural machine translation},
  author={Lu, Yu and Zeng, Jiali and Zhang, Jiajun and Wu, Shuangzhi and Li, Mu},
  journal={arXiv preprint arXiv:2203.11413},
  year={2022}
}

@article{muller2019whendoeslabelsmoothinghelp,
  title={When does label smoothing help?},
  author={M{\"u}ller, Rafael and Kornblith, Simon and Hinton, Geoffrey E},
  journal={Advances in neural information processing systems},
  volume={32},
  year={2019}
}

@article{liu2024activationconfcalibcodec,
  title={Enhancing Language Model Factuality via Activation-Based Confidence Calibration and Guided Decoding},
  author={Liu, Xin and Bayat, Farima Fatahi and Wang, Lu},
  journal={arXiv preprint arXiv:2406.13230},
  year={2024}
}

@inproceedings{liu2024litcab,
  title={LitCab: Lightweight Language Model Calibration over Short-and Long-form Responses},
  author={Liu, Xin and Khalifa, Muhammad and Wang, Lu},
  booktitle={The Twelfth International Conference on Learning Representations},
  year={2024}
}

@article{zhao2023slichf,
  title={Slic-hf: Sequence likelihood calibration with human feedback},
  author={Zhao, Yao and Joshi, Rishabh and Liu, Tianqi and Khalman, Misha and Saleh, Mohammad and Liu, Peter J},
  journal={arXiv preprint arXiv:2305.10425},
  year={2023}
}

@inproceedings{ zhao2023calibratingsli,
 title={Calibrating Sequence likelihood Improves Conditional Language Generation},
 author={Yao Zhao and Mikhail Khalman and Rishabh Joshi and Shashi Narayan and Mohammad Saleh and Peter J Liu},
 booktitle={The Eleventh International Conference on Learning Representations },
 year={2023},
 url={https://openreview.net/forum?id=0qSOodKmJaN}
}

@article{azaria2023saplma,
  title={The internal state of an LLM knows when it's lying},
  author={Azaria, Amos and Mitchell, Tom},
  journal={arXiv preprint arXiv:2304.13734},
  year={2023}
}

@article{shrivastava2023llamasknowwhatgptsdontshow,
  title={Llamas Know What GPTs Don't Show: Surrogate Models for Confidence Estimation},
  author={Shrivastava, Vaishnavi and Liang, Percy and Kumar, Ananya},
  journal={arXiv preprint arXiv:2311.08877},
  year={2023}
}

@article{ulmer2024apricot,
  title={Calibrating Large Language Models Using Their Generations Only},
  author={Ulmer, Dennis and Gubri, Martin and Lee, Hwaran and Yun, Sangdoo and Oh, Seong Joon},
  journal={arXiv preprint arXiv:2403.05973},
  year={2024}
}

@article{cohen2023lmvslm,
  title={Lm vs lm: Detecting factual errors via cross examination},
  author={Cohen, Roi and Hamri, May and Geva, Mor and Globerson, Amir},
  journal={arXiv preprint arXiv:2305.13281},
  year={2023}
}

@article{chen2025evaluatingcai,
  title={Evaluating LLMs Without Oracle Feedback: Agentic Annotation Evaluation Through Unsupervised Consistency Signals},
  author={Chen, Cheng and Yin, Haiyan and Tsang, Ivor},
  journal={arXiv preprint arXiv:2509.08809},
  year={2025}
}

@article{xue2025crossmodelconsistency,
  title={Verify when Uncertain: Beyond Self-Consistency in Black Box Hallucination Detection},
  author={Xue, Yihao and Greenewald, Kristjan and Mroueh, Youssef and Mirzasoleiman, Baharan},
  journal={arXiv preprint arXiv:2502.15845},
  year={2025}
}

@article{rivera2024sampleavgdev,
  title={Combining confidence elicitation and sample-based methods for uncertainty quantification in misinformation mitigation},
  author={Rivera, Mauricio and Godbout, Jean-Fran{\c{c}}ois and Rabbany, Reihaneh and Pelrine, Kellin},
  journal={arXiv preprint arXiv:2401.08694},
  year={2024}
}

@article{chen2023universalselfconsistency,
  title={Universal self-consistency for large language model generation},
  author={Chen, Xinyun and Aksitov, Renat and Alon, Uri and Ren, Jie and Xiao, Kefan and Yin, Pengcheng and Prakash, Sushant and Sutton, Charles and Wang, Xuezhi and Zhou, Denny},
  journal={arXiv preprint arXiv:2311.17311},
  year={2023}
}

@article{huang2024calibratinglongform,
  title={Calibrating Long-form Generations from Large Language Models},
  author={Huang, Yukun and Liu, Yixin and Thirukovalluru, Raghuveer and Cohan, Arman and Dhingra, Bhuwan},
  journal={arXiv preprint arXiv:2402.06544},
  year={2024}
}

@article{dai2023exploringmllmood,
  title={Exploring large language models for multi-modal out-of-distribution detection},
  author={Dai, Yi and Lang, Hao and Zeng, Kaisheng and Huang, Fei and Li, Yongbin},
  journal={arXiv preprint arXiv:2310.08027},
  year={2023}
}

@article{wang2022selfconsistency,
  title={Self-consistency improves chain of thought reasoning in language models},
  author={Wang, Xuezhi and Wei, Jason and Schuurmans, Dale and Le, Quoc and Chi, Ed and Narang, Sharan and Chowdhery, Aakanksha and Zhou, Denny},
  journal={arXiv preprint arXiv:2203.11171},
  year={2022}
}

@article{liu2025metafaith,
  title={MetaFaith: Faithful Natural Language Uncertainty Expression in LLMs},
  author={Liu, Gabrielle Kaili-May and Yona, Gal and Caciularu, Avi and Szpektor, Idan and Rudner, Tim GJ and Cohan, Arman},
  journal={arXiv preprint arXiv:2505.24858},
  year={2025}
}

@article{steyvers2025llmexplanationconfdiscgap,
  title={What large language models know and what people think they know},
  author={Steyvers, Mark and Tejeda, Heliodoro and Kumar, Aakriti and Belem, Catarina and Karny, Sheer and Hu, Xinyue and Mayer, Lukas W and Smyth, Padhraic},
  journal={Nature Machine Intelligence},
  pages={1--11},
  year={2025},
  publisher={Nature Publishing Group UK London}
}

@article{steyvers2024calibrationgap,
  title={The calibration gap between model and human confidence in large language models},
  author={Steyvers, Mark and Tejeda, Heliodoro and Kumar, Aakriti and Belem, Catarina and Karny, Sheer and Hu, Xinyue and Mayer, Lukas and Smyth, Padhraic},
  journal={arXiv preprint arXiv:2401.13835},
  year={2024}
}

@article{zhang2024atomiccalib,
  title={Atomic Calibration of LLMs in Long-Form Generations},
  author={Zhang, Caiqi and Yang, Ruihan and Zhang, Zhisong and Huang, Xinting and Yang, Sen and Yu, Dong and Collier, Nigel},
  journal={arXiv preprint arXiv:2410.13246},
  year={2024}
}

@article{zhang2024calibratingbyelicitingfidelity,
  title={Calibrating the Confidence of Large Language Models by Eliciting Fidelity},
  author={Zhang, Mozhi and Huang, Mianqiu and Shi, Rundong and Guo, Linsen and Peng, Chong and Yan, Peng and Zhou, Yaqian and Qiu, Xipeng},
  journal={arXiv preprint arXiv:2404.02655},
  year={2024}
}

@article{zhao2024far,
  title={Fact-and-Reflection (FaR) Improves Confidence Calibration of Large Language Models},
  author={Zhao, Xinran and Zhang, Hongming and Pan, Xiaoman and Yao, Wenlin and Yu, Dong and Wu, Tongshuang and Chen, Jianshu},
  journal={arXiv preprint arXiv:2402.17124},
  year={2024}
}

@article{zhou2024relyingonunreliable,
  title={Relying on the Unreliable: The Impact of Language Models' Reluctance to Express Uncertainty},
  author={Zhou, Kaitlyn and Hwang, Jena D and Ren, Xiang and Sap, Maarten},
  journal={arXiv preprint arXiv:2401.06730},
  year={2024}
}

@article{mielke2022reducingoverconfidence,
  title={Reducing conversational agents’ overconfidence through linguistic calibration},
  author={Mielke, Sabrina J and Szlam, Arthur and Dinan, Emily and Boureau, Y-Lan},
  journal={Transactions of the Association for Computational Linguistics},
  volume={10},
  pages={857--872},
  year={2022},
  publisher={MIT Press One Broadway, 12th Floor, Cambridge, Massachusetts 02142, USA~…}
}

@article{huang2024ulmrankcalib,
  title={Uncertainty in language models: Assessment through rank-calibration},
  author={Huang, Xinmeng and Li, Shuo and Yu, Mengxin and Sesia, Matteo and Hassani, Hamed and Lee, Insup and Bastani, Osbert and Dobriban, Edgar},
  journal={arXiv preprint arXiv:2404.03163},
  year={2024}
}

@inproceedings{luo2022lce,
  title={Local calibration: Metrics and recalibration},
  author={Luo, Rachel and Bhatnagar, Aadyot and Bai, Yu and Zhao, Shengjia and Wang, Huan and Xiong, Caiming and Savarese, Silvio and Ermon, Stefano and Schmerling, Edward and Pavone, Marco},
  booktitle={Uncertainty in Artificial Intelligence},
  pages={1286--1295},
  year={2022},
  organization={PMLR}
}

@article{tran2022plex,
  title={Plex: Towards reliability using pretrained large model extensions},
  author={Tran, Dustin and Liu, Jeremiah and Dusenberry, Michael W and Phan, Du and Collier, Mark and Ren, Jie and Han, Kehang and Wang, Zi and Mariet, Zelda and Hu, Huiyi and others},
  journal={arXiv preprint arXiv:2207.07411},
  year={2022}
}

@article{kumar2019verified,
  title={Verified uncertainty calibration},
  author={Kumar, Ananya and Liang, Percy S and Ma, Tengyu},
  journal={Advances in Neural Information Processing Systems},
  volume={32},
  year={2019}
}

@inproceedings{zadrozny2002isotonicregression,
  title={Transforming classifier scores into accurate multiclass probability estimates},
  author={Zadrozny, Bianca and Elkan, Charles},
  booktitle={Proceedings of the eighth ACM SIGKDD international conference on Knowledge discovery and data mining},
  pages={694--699},
  year={2002}
}

@inproceedings{zadrozny2001histogrambinning,
  title={Obtaining calibrated probability estimates from decision trees and naive bayesian classifiers},
  author={Zadrozny, Bianca and Elkan, Charles},
  booktitle={Icml},
  volume={1},
  pages={609--616},
  year={2001}
}

@article{dhuliawala2022jointcalibration,
  title={Calibration of machine reading systems at scale},
  author={Dhuliawala, Shehzaad and Adolphs, Leonard and Das, Rajarshi and Sachan, Mrinmaya},
  journal={arXiv preprint arXiv:2203.10623},
  year={2022}
}

@article{yuksekgonul2024atypicalitycalib,
  title={Beyond confidence: Reliable models should also consider atypicality},
  author={Yuksekgonul, Mert and Zhang, Linjun and Zou, James Y and Guestrin, Carlos},
  journal={Advances in Neural Information Processing Systems},
  volume={36},
  year={2024}
}

@inproceedings{guo2017calibration,
  title={On calibration of modern neural networks},
  author={Guo, Chuan and Pleiss, Geoff and Sun, Yu and Weinberger, Kilian Q},
  booktitle={International conference on machine learning},
  pages={1321--1330},
  year={2017},
  organization={PMLR}
}

@article{platt1999prbbplattscaling,
  title={Probabilistic outputs for support vector machines and comparisons to regularized likelihood methods},
  author={Platt, John and others},
  journal={Advances in large margin classifiers},
  volume={10},
  number={3},
  pages={61--74},
  year={1999},
  publisher={Cambridge, MA}
}

@article{diao2023activeprompt,
  title={Active prompting with chain-of-thought for large language models},
  author={Diao, Shizhe and Wang, Pengcheng and Lin, Yong and Pan, Rui and Liu, Xiang and Zhang, Tong},
  journal={arXiv preprint arXiv:2302.12246},
  year={2023}
}

@article{Yin2024LID,
  title={Characterizing Truthfulness in Large Language Model Generations with Local Intrinsic Dimension},
  author={Fan Yin and Jayanth Srinivasa and Kai-Wei Chang},
  journal={ArXiv},
  year={2024},
  volume={abs/2402.18048},
  url={https://api.semanticscholar.org/CorpusID:268041551}
}

@article{baan2022stopmeasurnigcaldisce,
  title={Stop measuring calibration when humans disagree},
  author={Baan, Joris and Aziz, Wilker and Plank, Barbara and Fernandez, Raquel},
  journal={arXiv preprint arXiv:2210.16133},
  year={2022}
}

@inproceedings{peterson2019humancategoricalsimilarity,
  title={Human uncertainty makes classification more robust},
  author={Peterson, Joshua C and Battleday, Ruairidh M and Griffiths, Thomas L and Russakovsky, Olga},
  booktitle={Proceedings of the IEEE/CVF international conference on computer vision},
  pages={9617--9626},
  year={2019}
}

@inproceedings{miao2021preventoverconf,
  title={Prevent the Language Model from being Overconfident in Neural Machine Translation},
  author={Miao, Mengqi and Meng, Fandong and Liu, Yijin and Zhou, Xiao-Hua and Zhou, Jie},
  booktitle={Proceedings of the 59th Annual Meeting of the Association for Computational Linguistics and the 11th International Joint Conference on Natural Language Processing (Volume 1: Long Papers)},
  pages={3456--3468},
  year={2021}
}

@article{joo2025blackboxhallucinationmetric,
  title={Black-Box Hallucination Detection via Consistency Under the Uncertain Expression},
  author={Joo, Seongho and Min, Kyungmin and Koo, Jahyun and Jung, Kyomin},
  journal={arXiv preprint arXiv:2509.21999},
  year={2025}
}

@article{zhang2024luq,
  title={LUQ: Long-text Uncertainty Quantification for LLMs},
  author={Zhang, Caiqi and Liu, Fangyu and Basaldella, Marco and Collier, Nigel},
  journal={arXiv preprint arXiv:2403.20279},
  year={2024}
}

@article{huang2024uncttp,
  title={Unc-TTP: A Method for Classifying LLM Uncertainty to Improve In-Context Example Selection},
  author={Huang, Hsiu-Yuan and Wu, Zichen and Yang, Yutong and Zhang, Junzhao and Wu, Yunfang},
  journal={arXiv preprint arXiv:2408.09172},
  year={2024}
}

@article{lin2022teachingmodelstoexpress,
  title={Teaching models to express their uncertainty in words},
  author={Lin, Stephanie and Hilton, Jacob and Evans, Owain},
  journal={arXiv preprint arXiv:2205.14334},
  year={2022}
}

@article{si2022promptinggpt3reliable,
  title={Prompting gpt-3 to be reliable},
  author={Si, Chenglei and Gan, Zhe and Yang, Zhengyuan and Wang, Shuohang and Wang, Jianfeng and Boyd-Graber, Jordan and Wang, Lijuan},
  journal={arXiv preprint arXiv:2210.09150},
  year={2022}
}

@article{zhou2023navigating,
  title={Navigating the grey area: Expressions of overconfidence and uncertainty in language models},
  author={Zhou, Kaitlyn and Jurafsky, Dan and Hashimoto, Tatsunori},
  journal={arXiv e-prints},
  pages={arXiv--2302},
  year={2023}
}

@article{ganguli2022redteamingllms,
  title={Red teaming language models to reduce harms: Methods, scaling behaviors, and lessons learned},
  author={Ganguli, Deep and Lovitt, Liane and Kernion, Jackson and Askell, Amanda and Bai, Yuntao and Kadavath, Saurav and Mann, Ben and Perez, Ethan and Schiefer, Nicholas and Ndousse, Kamal and others},
  journal={arXiv preprint arXiv:2209.07858},
  year={2022}
}

@article{ling2023improvingdemonstuncert,
  title={Improving open information extraction with large language models: A study on demonstration uncertainty},
  author={Ling, Chen and Zhao, Xujiang and Zhang, Xuchao and Liu, Yanchi and Cheng, Wei and Wang, Haoyu and Chen, Zhengzhang and Osaki, Takao and Matsuda, Katsushi and Chen, Haifeng and others},
  journal={arXiv preprint arXiv:2309.03433},
  year={2023}
}

@article{catak2024uqoconvexhullarea,
  title={Uncertainty quantification in large language models through convex hull analysis},
  author={Catak, Ferhat Ozgur and Kuzlu, Murat},
  journal={Discover Artificial Intelligence},
  volume={4},
  number={1},
  pages={1--14},
  year={2024},
  publisher={Springer}
}

@misc{irawan2025entropy2veccrosslinguallanguagemodeling,
      title={Entropy2Vec: Crosslingual Language Modeling Entropy as End-to-End Learnable Language Representations}, 
      author={Patrick Amadeus Irawan and Ryandito Diandaru and Belati Jagad Bintang Syuhada and Randy Zakya Suchrady and Alham Fikri Aji and Genta Indra Winata and Fajri Koto and Samuel Cahyawijaya},
      year={2025},
      eprint={2509.05060},
      archivePrefix={arXiv},
      primaryClass={cs.CL},
      url={https://arxiv.org/abs/2509.05060}, 
}

@article{Nikitin2024KernelLE,
  title={Kernel Language Entropy: Fine-grained Uncertainty Quantification for LLMs from Semantic Similarities},
  author={Alexander Nikitin and Jannik Kossen and Yarin Gal and Pekka Marttinen},
  journal={ArXiv},
  year={2024},
  volume={abs/2405.20003},
  url={https://api.semanticscholar.org/CorpusID:270123445}
}

@article{ding2024rowen,
  title={Retrieve only when it needs: Adaptive retrieval augmentation for hallucination mitigation in large language models},
  author={Ding, Hanxing and Pang, Liang and Wei, Zihao and Shen, Huawei and Cheng, Xueqi},
  journal={arXiv preprint arXiv:2402.10612},
  year={2024}
}

@article{fomicheva2020lexsim,
  title={Unsupervised quality estimation for neural machine translation},
  author={Fomicheva, Marina and Sun, Shuo and Yankovskaya, Lisa and Blain, Fr{\'e}d{\'e}ric and Guzm{\'a}n, Francisco and Fishel, Mark and Aletras, Nikolaos and Chaudhary, Vishrav and Specia, Lucia},
  journal={Transactions of the Association for Computational Linguistics},
  volume={8},
  pages={539--555},
  year={2020},
  publisher={MIT Press One Rogers Street, Cambridge, MA 02142-1209, USA journals-info~…}
}

@article{grewal2024improvinguqllmsemanticembedding,
  title={Improving Uncertainty Quantification in Large Language Models via Semantic Embeddings},
  author={Grewal, Yashvir S and Bonilla, Edwin V and Bui, Thang D},
  journal={arXiv preprint arXiv:2410.22685},
  year={2024}
}

@article{lin2023generatingwithconfblackbox,
  title={Generating with confidence: Uncertainty quantification for black-box large language models},
  author={Lin, Zhen and Trivedi, Shubhendu and Sun, Jimeng},
  journal={arXiv preprint arXiv:2305.19187},
  year={2023}
}

@inproceedings{chen2024quantifyingtrustworthiness,
  title={Quantifying uncertainty in answers from any language model and enhancing their trustworthiness},
  author={Chen, Jiuhai and Mueller, Jonas},
  booktitle={Proceedings of the 62nd Annual Meeting of the Association for Computational Linguistics (Volume 1: Long Papers)},
  pages={5186--5200},
  year={2024}
}

@inproceedings{li2024thinktwicebeforetrusting,
  title={Think twice before trusting: Self-detection for large language models through comprehensive answer reflection},
  author={Li, Moxin and Wang, Wenjie and Feng, Fuli and Zhu, Fengbin and Wang, Qifan and Chua, Tat-Seng},
  booktitle={Findings of the Association for Computational Linguistics: EMNLP 2024},
  pages={11858--11875},
  year={2024}
}

@article{ji2023selfreflection,
  title={Towards mitigating hallucination in large language models via self-reflection},
  author={Ji, Ziwei and Yu, Tiezheng and Xu, Yan and Lee, Nayeon and Ishii, Etsuko and Fung, Pascale},
  journal={arXiv preprint arXiv:2310.06271},
  year={2023}
}

@article{zhao2023selfdetectionanswerclustering,
  title={Knowing what llms do not know: A simple yet effective self-detection method},
  author={Zhao, Yukun and Yan, Lingyong and Sun, Weiwei and Xing, Guoliang and Meng, Chong and Wang, Shuaiqiang and Cheng, Zhicong and Ren, Zhaochun and Yin, Dawei},
  journal={arXiv preprint arXiv:2310.17918},
  year={2023}
}

@article{Agrawal2023overlapestimation,
  title={Do Language Models Know When They’re Hallucinating References?},
  author={Ayush Kumar Agrawal and Lester W. Mackey and Adam Tauman Kalai},
  journal={ArXiv},
  year={2023},
  volume={abs/2305.18248},
  url={https://api.semanticscholar.org/CorpusID:258960346}
}

@article{ling2024deductiveverification,
  title={Deductive verification of chain-of-thought reasoning},
  author={Ling, Zhan and Fang, Yunhao and Li, Xuanlin and Huang, Zhiao and Lee, Mingu and Memisevic, Roland and Su, Hao},
  journal={Advances in Neural Information Processing Systems},
  volume={36},
  year={2024}
}

@article{madaan2024selfrefine,
  title={Self-refine: Iterative refinement with self-feedback},
  author={Madaan, Aman and Tandon, Niket and Gupta, Prakhar and Hallinan, Skyler and Gao, Luyu and Wiegreffe, Sarah and Alon, Uri and Dziri, Nouha and Prabhumoye, Shrimai and Yang, Yiming and others},
  journal={Advances in Neural Information Processing Systems},
  volume={36},
  year={2024}
}

@article{manakul2023selfcheckgpt,
  title={Selfcheckgpt: Zero-resource black-box hallucination detection for generative large language models},
  author={Manakul, Potsawee and Liusie, Adian and Gales, Mark JF},
  journal={arXiv preprint arXiv:2303.08896},
  year={2023}
}

@article{miao2023selfcheck,
  title={Selfcheck: Using llms to zero-shot check their own step-by-step reasoning},
  author={Miao, Ning and Teh, Yee Whye and Rainforth, Tom},
  journal={arXiv preprint arXiv:2308.00436},
  year={2023}
}

@article{weng2022selfverification,
  title={Large language models are better reasoners with self-verification},
  author={Weng, Yixuan and Zhu, Minjun and Xia, Fei and Li, Bin and He, Shizhu and Liu, Shengping and Sun, Bin and Liu, Kang and Zhao, Jun},
  journal={arXiv preprint arXiv:2212.09561},
  year={2022}
}

@inproceedings{ren2023selfeval,
  title={Self-evaluation improves selective generation in large language models},
  author={Ren, Jie and Zhao, Yao and Vu, Tu and Liu, Peter J and Lakshminarayanan, Balaji},
  booktitle={Proceedings on},
  pages={49--64},
  year={2023},
  organization={PMLR}
}

@article{fisch2022uncertaintyfornlptutorial,
  title={Uncertainty estimation for natural language processing},
  author={Fisch, Adam and Jia, Robin and Schuster, Tal},
  year={2022}
}

@article{liang2023mid,
  title={Encouraging divergent thinking in large language models through multi-agent debate},
  author={Liang, Tian and He, Zhiwei and Jiao, Wenxiang and Wang, Xing and Wang, Yan and Wang, Rui and Yang, Yujiu and Shi, Shuming and Tu, Zhaopeng},
  journal={arXiv preprint arXiv:2305.19118},
  year={2023}
}

@article{chen2023reconcile,
  title={Reconcile: Round-table conference improves reasoning via consensus among diverse llms},
  author={Chen, Justin Chih-Yao and Saha, Swarnadeep and Bansal, Mohit},
  journal={arXiv preprint arXiv:2309.13007},
  year={2023}
}

@article{yoffe2024debunc,
  title={DebUnc: mitigating hallucinations in large language model agent communication with uncertainty estimations},
  author={Yoffe, Luke and Amayuelas, Alfonso and Wang, William Yang},
  journal={arXiv preprint arXiv:2407.06426},
  year={2024}
}

@article{gou2023critic,
  title={Critic: Large language models can self-correct with tool-interactive critiquing},
  author={Gou, Zhibin and Shao, Zhihong and Gong, Yeyun and Shen, Yelong and Yang, Yujiu and Duan, Nan and Chen, Weizhu},
  journal={arXiv preprint arXiv:2305.11738},
  year={2023}
}

@article{han2024towardsuala,
  title={Towards Uncertainty-Aware Language Agent},
  author={Han, Jiuzhou and Buntine, Wray and Shareghi, Ehsan},
  journal={arXiv preprint arXiv:2401.14016},
  year={2024}
}

@article{shinn2024reflexion,
  title={Reflexion: Language agents with verbal reinforcement learning},
  author={Shinn, Noah and Cassano, Federico and Gopinath, Ashwin and Narasimhan, Karthik and Yao, Shunyu},
  journal={Advances in Neural Information Processing Systems},
  volume={36},
  year={2024}
}

@article{wang2025sconu,
  title={Sconu: Selective conformal uncertainty in large language models},
  author={Wang, Zhiyuan and Wang, Qingni and Zhang, Yue and Chen, Tianlong and Zhu, Xiaofeng and Shi, Xiaoshuang and Xu, Kaidi},
  journal={arXiv preprint arXiv:2504.14154},
  year={2025}
}

@article{kaur2024ddcrp,
  title={Addressing uncertainty in llms to enhance reliability in generative ai},
  author={Kaur, Ramneet and Samplawski, Colin and Cobb, Adam D and Roy, Anirban and Matejek, Brian and Acharya, Manoj and Elenius, Daniel and Berenbeim, Alexander M and Pavlik, John A and Bastian, Nathaniel D and others},
  journal={arXiv preprint arXiv:2411.02381},
  year={2024}
}

@article{su2024lofreecp,
  title={Api is enough: Conformal prediction for large language models without logit-access},
  author={Su, Jiayuan and Luo, Jing and Wang, Hongwei and Cheng, Lu},
  journal={arXiv preprint arXiv:2403.01216},
  year={2024}
}

@article{wang2024conu,
  title={Conu: Conformal uncertainty in large language models with correctness coverage guarantees},
  author={Wang, Zhiyuan and Duan, Jinhao and Cheng, Lu and Zhang, Yue and Wang, Qingni and Shi, Xiaoshuang and Xu, Kaidi and Shen, Hengtao and Zhu, Xiaofeng},
  journal={arXiv preprint arXiv:2407.00499},
  year={2024}
}

@article{ulmer2024nonexchangeablecp,
  title={Non-exchangeable conformal language generation with nearest neighbors},
  author={Ulmer, Dennis and Zerva, Chrysoula and Martins, Andr{\'e} FT},
  journal={arXiv preprint arXiv:2402.00707},
  year={2024}
}

@article{quach2023conformallangmod,
  title={Conformal language modeling},
  author={Quach, Victor and Fisch, Adam and Schuster, Tal and Yala, Adam and Sohn, Jae Ho and Jaakkola, Tommi S and Barzilay, Regina},
  journal={arXiv preprint arXiv:2306.10193},
  year={2023}
}

@article{ren2023robotsknowno,
  title={Robots that ask for help: Uncertainty alignment for large language model planners},
  author={Ren, Allen Z and Dixit, Anushri and Bodrova, Alexandra and Singh, Sumeet and Tu, Stephen and Brown, Noah and Xu, Peng and Takayama, Leila and Xia, Fei and Varley, Jake and others},
  journal={arXiv preprint arXiv:2307.01928},
  year={2023}
}

@article{kumar2023conformalllmmcqa,
  title={Conformal prediction with large language models for multi-choice question answering},
  author={Kumar, Bhawesh and Lu, Charlie and Gupta, Gauri and Palepu, Anil and Bellamy, David and Raskar, Ramesh and Beam, Andrew},
  journal={arXiv preprint arXiv:2305.18404},
  year={2023}
}

@article{shafer2008tutorialconfpred,
  title={A tutorial on conformal prediction.},
  author={Shafer, Glenn and Vovk, Vladimir},
  journal={Journal of Machine Learning Research},
  volume={9},
  number={3},
  year={2008}
}

@inproceedings{vazhentsev2023hybridue,
  title={Hybrid uncertainty quantification for selective text classification in ambiguous tasks},
  author={Vazhentsev, Artem and Kuzmin, Gleb and Tsvigun, Akim and Panchenko, Alexander and Panov, Maxim and Burtsev, Mikhail and Shelmanov, Artem},
  booktitle={Proceedings of the 61st Annual Meeting of the Association for Computational Linguistics (Volume 1: Long Papers)},
  pages={11659--11681},
  year={2023}
}

@article{Qiu2024SemanticDU,
  title={Semantic Density: Uncertainty Quantification in Semantic Space for Large Language Models},
  author={Xin Qiu and Risto Miikkulainen},
  journal={ArXiv},
  year={2024},
  volume={abs/2405.13845},
  url={https://api.semanticscholar.org/CorpusID:269983698}
}

@article{fadeeva2024ccp,
  title={Fact-checking the output of large language models via token-level uncertainty quantification},
  author={Fadeeva, Ekaterina and Rubashevskii, Aleksandr and Shelmanov, Artem and Petrakov, Sergey and Li, Haonan and Mubarak, Hamdy and Tsymbalov, Evgenii and Kuzmin, Gleb and Panchenko, Alexander and Baldwin, Timothy and others},
  journal={arXiv preprint arXiv:2403.04696},
  year={2024}
}

@inproceedings{duan2024shiftingattentiontorelevance,
  title={Shifting attention to relevance: Towards the predictive uncertainty quantification of free-form large language models},
  author={Duan, Jinhao and Cheng, Hao and Wang, Shiqi and Zavalny, Alex and Wang, Chenan and Xu, Renjing and Kailkhura, Bhavya and Xu, Kaidi},
  booktitle={Proceedings of the 62nd Annual Meeting of the Association for Computational Linguistics (Volume 1: Long Papers)},
  pages={5050--5063},
  year={2024}
}

@inproceedings{
aichberger2025sdlg,
title={Improving Uncertainty Estimation through Semantically Diverse Language Generation},
author={Lukas Aichberger and Kajetan Schweighofer and Mykyta Ielanskyi and Sepp Hochreiter},
booktitle={The Thirteenth International Conference on Learning Representations},
year={2025},
url={https://openreview.net/forum?id=HSi4VetQLj}
}

@article{wang2025wordsequenceentropy,
  title={Word-sequence entropy: Towards uncertainty estimation in free-form medical question answering applications and beyond},
  author={Wang, Zhiyuan and Duan, Jinhao and Yuan, Chenxi and Chen, Qingyu and Chen, Tianlong and Zhang, Yue and Wang, Ren and Shi, Xiaoshuang and Xu, Kaidi},
  journal={Engineering Applications of Artificial Intelligence},
  volume={139},
  pages={109553},
  year={2025},
  publisher={Elsevier}
}

@article{farquhar2024detecting,
  title={Detecting hallucinations in large language models using semantic entropy},
  author={Farquhar, Sebastian and Kossen, Jannik and Kuhn, Lorenz and Gal, Yarin},
  journal={Nature},
  volume={630},
  number={8017},
  pages={625--630},
  year={2024},
  publisher={Nature Publishing Group UK London}
}

@article{kuhn2023semanticuncertainty,
  title={Semantic uncertainty: Linguistic invariances for uncertainty estimation in natural language generation},
  author={Kuhn, Lorenz and Gal, Yarin and Farquhar, Sebastian},
  journal={arXiv preprint arXiv:2302.09664},
  year={2023}
}

@article{audrino2024uqindices,
  title={Quantifying uncertainty: a new era of measurement through large language models},
  author={Audrino, Francesco and Gentner, Jessica and Stalder, Simon},
  year={2024},
  publisher={Swiss National Bank Working Paper}
}

@inproceedings{liu2024uait,
  title={Can LLMs learn uncertainty on their own? expressing uncertainty effectively in a self-training manner},
  author={Liu, Shudong and Li, Zhaocong and Liu, Xuebo and Zhan, Runzhe and Wong, Derek and Chao, Lidia and Zhang, Min},
  booktitle={Proceedings of the 2024 Conference on Empirical Methods in Natural Language Processing},
  pages={21635--21645},
  year={2024}
}

@inproceedings{ling2024uqicl,
  title={Uncertainty Quantification for In-Context Learning of Large Language Models},
  author={Ling, Chen and Zhao, Xujiang and Zhang, Xuchao and Cheng, Wei and Liu, Yanchi and Sun, Yiyou and Oishi, Mika and Osaki, Takao and Matsuda, Katsushi and Ji, Jie and others},
  booktitle={Proceedings of the 2024 Conference of the North American Chapter of the Association for Computational Linguistics: Human Language Technologies (Volume 1: Long Papers)},
  pages={3357--3370},
  year={2024}
}

@article{malinin2020uncertaintylengthnormalized,
  title={Uncertainty estimation in autoregressive structured prediction},
  author={Malinin, Andrey and Gales, Mark},
  journal={arXiv preprint arXiv:2002.07650},
  year={2020}
}

@inproceedings{xiao2019quantifyingunlp,
  title={Quantifying uncertainties in natural language processing tasks},
  author={Xiao, Yijun and Wang, William Yang},
  booktitle={Proceedings of the AAAI conference on artificial intelligence},
  volume={33},
  number={01},
  pages={7322--7329},
  year={2019}
}

@article{yang2023improvingiclselfcorr,
  title={Improving the Reliability of Large Language Models by Leveraging Uncertainty-Aware In-Context Learning},
  author={Yang, Yuchen and Li, Houqiang and Wang, Yanfeng and Wang, Yu},
  journal={arXiv preprint arXiv:2310.04782},
  year={2023}
}

@article{xiao2021hallucinationanduncertainty,
  title={On hallucination and predictive uncertainty in conditional language generation},
  author={Xiao, Yijun and Wang, William Yang},
  journal={arXiv preprint arXiv:2103.15025},
  year={2021}
}

@article{kadavath2022llmsknowmostly,
  title={Language models (mostly) know what they know},
  author={Kadavath, Saurav and Conerly, Tom and Askell, Amanda and Henighan, Tom and Drain, Dawn and Perez, Ethan and Schiefer, Nicholas and Hatfield-Dodds, Zac and DasSarma, Nova and Tran-Johnson, Eli and others},
  journal={arXiv preprint arXiv:2207.05221},
  year={2022}
}

@inproceedings{tanneru2024quantifyinguncertaintyllm,
  title={Quantifying uncertainty in natural language explanations of large language models},
  author={Tanneru, Sree Harsha and Agarwal, Chirag and Lakkaraju, Himabindu},
  booktitle={International Conference on Artificial Intelligence and Statistics},
  pages={1072--1080},
  year={2024},
  organization={PMLR}
}

@article{gao2024spuq,
  title={SPUQ: Perturbation-Based Uncertainty Quantification for Large Language Models},
  author={Gao, Xiang and Zhang, Jiaxin and Mouatadid, Lalla and Das, Kamalika},
  journal={arXiv preprint arXiv:2403.02509},
  year={2024}
}

@article{hou2023decomposinguncertaintyinputc,
  title={Decomposing uncertainty for large language models through input clarification ensembling},
  author={Hou, Bairu and Liu, Yujian and Qian, Kaizhi and Andreas, Jacob and Chang, Shiyu and Zhang, Yang},
  journal={arXiv preprint arXiv:2311.08718},
  year={2023}
}

@article{johnson2023rusure,
  title={RU-SURE? uncertainty-aware code suggestions by maximizing utility across random user intents},
  author={Johnson, Daniel D and Tarlow, Daniel and Walder, Christian},
  journal={arXiv preprint arXiv:2303.00732},
  year={2023}
}

@inproceedings{sun2022quantifyingmodelensemble,
  title={Quantifying uncertainty in foundation models via ensembles},
  author={Sun, Meiqi and Yan, Wilson and Abbeel, Pieter and Mordatch, Igor},
  booktitle={NeurIPS 2022 Workshop on Robustness in Sequence Modeling},
  year={2022}
}

@inproceedings{gal2016dropout,
  title={Dropout as a bayesian approximation: Representing model uncertainty in deep learning},
  author={Gal, Yarin and Ghahramani, Zoubin},
  booktitle={international conference on machine learning},
  pages={1050--1059},
  year={2016},
  organization={PMLR}
}

@article{loquercio2020generaluedl,
  title={A general framework for uncertainty estimation in deep learning},
  author={Loquercio, Antonio and Segu, Mattia and Scaramuzza, Davide},
  journal={IEEE Robotics and Automation Letters},
  volume={5},
  number={2},
  pages={3153--3160},
  year={2020},
  publisher={IEEE}
}

@article{wen2020batchensemble,
  title={Batchensemble: an alternative approach to efficient ensemble and lifelong learning},
  author={Wen, Yeming and Tran, Dustin and Ba, Jimmy},
  journal={arXiv preprint arXiv:2002.06715},
  year={2020}
}

@article{degroot1983comparisonforecaster,
  title={The comparison and evaluation of forecasters},
  author={DeGroot, Morris H and Fienberg, Stephen E},
  journal={Journal of the Royal Statistical Society: Series D (The Statistician)},
  volume={32},
  number={1-2},
  pages={12--22},
  year={1983},
  publisher={Wiley Online Library}
}

@article{dawid1982wellcalibratedbayesian,
  title={The well-calibrated Bayesian},
  author={Dawid, A Philip},
  journal={Journal of the American Statistical Association},
  volume={77},
  number={379},
  pages={605--610},
  year={1982},
  publisher={Taylor \& Francis}
}

@article{murphy1977reliabilityofforecasts,
  title={Reliability of subjective probability forecasts of precipitation and temperature},
  author={Murphy, Allan H and Winkler, Robert L},
  journal={Journal of the Royal Statistical Society Series C: Applied Statistics},
  volume={26},
  number={1},
  pages={41--47},
  year={1977},
  publisher={Oxford University Press}
}

@article{deng2023agreementscore,
  title={Great models think alike: improving model reliability via inter-model latent agreement},
  author={Deng, Ailin and Xiong, Miao and Hooi, Bryan},
  journal={arXiv preprint arXiv:2305.01481},
  year={2023}
}

@inproceedings{he2023clur,
  title={CLUR: uncertainty estimation for few-shot text classification with contrastive learning},
  author={He, Jianfeng and Zhang, Xuchao and Lei, Shuo and Alhamadani, Abdulaziz and Chen, Fanglan and Xiao, Bei and Lu, Chang-Tien},
  booktitle={Proceedings of the 29th ACM SIGKDD Conference on Knowledge Discovery and Data Mining},
  pages={698--710},
  year={2023}
}

@misc{phasetransition2025,
       title={Phase Transitions in Large Language Models and the $O(N)$ Model}, 
       author={Youran Sun and Babak Haghighat},
       year={2025},
       eprint={2501.16241},
       archivePrefix={arXiv},
       primaryClass={cs.LG},
       url={https://arxiv.org/abs/2501.16241}, 
}

@article{wang2023uncertretinal,
  title={Uncertainty-inspired open set learning for retinal anomaly identification},
  author={Wang, Meng and Lin, Tian and Wang, Lianyu and Lin, Aidi and Zou, Ke and Xu, Xinxing and Zhou, Yi and Peng, Yuanyuan and Meng, Qingquan and Qian, Yiming and others},
  journal={Nature Communications},
  volume={14},
  number={1},
  pages={6757},
  year={2023},
  publisher={Nature Publishing Group UK London}
}

@inproceedings{bengs2023secondorderloss,
  title={On second-order scoring rules for epistemic uncertainty quantification},
  author={Bengs, Viktor and H{\"u}llermeier, Eyke and Waegeman, Willem},
  booktitle={International Conference on Machine Learning},
  pages={2078--2091},
  year={2023},
  organization={PMLR}
}

@inproceedings{yan2023distortion,
  title={Distortion and uncertainty aware loss for panoramic depth completion},
  author={Yan, Zhiqiang and Li, Xiang and Wang, Kun and Chen, Shuo and Li, Jun and Yang, Jian},
  booktitle={International Conference on Machine Learning},
  pages={39099--39109},
  year={2023},
  organization={PMLR}
}

@article{danruo2023uncertainty,
  title={Uncertainty Estimation by Fisher Information-based Evidential Deep Learning},
  author={Danruo, DENG and Chen, Guangyong and Yang, YU and Liu, Furui and Heng, Pheng-Ann},
  year={2023}
}

@article{Kirchhof2024PretrainedVU,
  title={Pretrained Visual Uncertainties},
  author={Michael Kirchhof and Mark Collier and Seong Joon Oh and Enkelejda Kasneci},
  journal={ArXiv},
  year={2024},
  volume={abs/2402.16569}
}

@article{kirchhof2023probabilistic,
  title={Probabilistic contrastive learning recovers the correct aleatoric uncertainty of ambiguous inputs},
  author={Kirchhof, Michael and Kasneci, Enkelejda and Oh, Seong Joon},
  journal={arXiv preprint arXiv:2302.02865},
  year={2023}
}

@inproceedings{Daxberger2021laplaceredux,
  title={Laplace Redux - Effortless Bayesian Deep Learning},
  author={Erik A. Daxberger and Agustinus Kristiadi and Alexander Immer and Runa Eschenhagen and M. Bauer and Philipp Hennig},
  booktitle={Neural Information Processing Systems},
  year={2021},
  url={https://api.semanticscholar.org/CorpusID:235658031}
}

@article{Cinquin2021PathologiesIP,
  title={Pathologies in priors and inference for Bayesian transformers},
  author={Tristan Cinquin and Alexander Immer and Max Horn and Vincent Fortuin},
  journal={ArXiv},
  year={2021},
  volume={abs/2110.04020},
  url={https://api.semanticscholar.org/CorpusID:238531503}
}

@article{sankararaman2022bayesformer,
  title={Bayesformer: Transformer with uncertainty estimation},
  author={Sankararaman, Karthik Abinav and Wang, Sinong and Fang, Han},
  journal={arXiv preprint arXiv:2206.00826},
  year={2022}
}

@article{Tran2018BayesianLA,
  title={Bayesian Layers: A Module for Neural Network Uncertainty},
  author={Dustin Tran and Michael W. Dusenberry and Mark van der Wilk and Danijar Hafner},
  journal={ArXiv},
  year={2018},
  volume={abs/1812.03973},
  url={https://api.semanticscholar.org/CorpusID:54461378}
}

@article{Zhang2019CyclicalSG,
  title={Cyclical Stochastic Gradient MCMC for Bayesian Deep Learning},
  author={Ruqi Zhang and Chunyuan Li and Jianyi Zhang and Changyou Chen and Andrew Gordon Wilson},
  journal={ArXiv},
  year={2019},
  volume={abs/1902.03932},
  url={https://api.semanticscholar.org/CorpusID:60440536}
}

@inproceedings{blundell2015bayesbybackprop,
  title={Weight uncertainty in neural network},
  author={Blundell, Charles and Cornebise, Julien and Kavukcuoglu, Koray and Wierstra, Daan},
  booktitle={International conference on machine learning},
  pages={1613--1622},
  year={2015},
  organization={PMLR}
}

@inproceedings{upadhyay2022bayescap,
  title={Bayescap: Bayesian identity cap for calibrated uncertainty in frozen neural networks},
  author={Upadhyay, Uddeshya and Karthik, Shyamgopal and Chen, Yanbei and Mancini, Massimiliano and Akata, Zeynep},
  booktitle={European Conference on Computer Vision},
  pages={299--317},
  year={2022},
  organization={Springer}
}

@article{buntine1991bayesian,
  title={Bayesian backpropagation},
  author={Buntine, Wray L},
  journal={Complex systems},
  volume={5},
  pages={603--643},
  year={1991}
}

@article{malinin2018predictivepriornetworks,
  title={Predictive uncertainty estimation via prior networks},
  author={Malinin, Andrey and Gales, Mark},
  journal={Advances in neural information processing systems},
  volume={31},
  year={2018}
}

@article{lakshminarayanan2017simple,
  title={Simple and scalable predictive uncertainty estimation using deep ensembles},
  author={Lakshminarayanan, Balaji and Pritzel, Alexander and Blundell, Charles},
  journal={Advances in neural information processing systems},
  volume={30},
  year={2017}
}

@article{dietterich2000experimental,
  title={An experimental comparison of three methods for constructing ensembles of decision trees: Bagging, boosting, and randomization},
  author={Dietterich, Thomas G},
  journal={Machine learning},
  volume={40},
  pages={139--157},
  year={2000},
  publisher={Springer}
}

@inproceedings{nix1994estimating,
  title={Estimating the mean and variance of the target probability distribution},
  author={Nix, David A and Weigend, Andreas S},
  booktitle={Proceedings of 1994 ieee international conference on neural networks (ICNN'94)},
  volume={1},
  pages={55--60},
  year={1994},
  organization={IEEE}
}

@inproceedings{zelenka2023explainableue,
  title={A Simple and Explainable Method for Uncertainty Estimation using Attribute Prototype Networks},
  author={Zelenka, Claudius and G{\"o}hring, Andrea and Kazempour, Daniyal and H{\"u}nem{\"o}rder, Maximilian and Schmarje, Lars and Kr{\"o}ger, Peer},
  booktitle={Proceedings of the IEEE/CVF International Conference on Computer Vision},
  pages={4570--4579},
  year={2023}
}

@article{achtibat2023attribution,
  title={From attribution maps to human-understandable explanations through Concept Relevance Propagation},
  author={Achtibat, Reduan and Dreyer, Maximilian and Eisenbraun, Ilona and Bosse, Sebastian and Wiegand, Thomas and Samek, Wojciech and Lapuschkin, Sebastian},
  journal={Nature Machine Intelligence},
  volume={5},
  number={9},
  pages={1006--1019},
  year={2023},
  publisher={Nature Publishing Group UK London}
}

@inproceedings{upadhyay2023probvlm,
  title={Probvlm: Probabilistic adapter for frozen vison-language models},
  author={Upadhyay, Uddeshya and Karthik, Shyamgopal and Mancini, Massimiliano and Akata, Zeynep},
  booktitle={Proceedings of the IEEE/CVF International Conference on Computer Vision},
  pages={1899--1910},
  year={2023}
}

@article{chun2023pcmeplusplus,
  title={Improved probabilistic image-text representations},
  author={Chun, Sanghyuk},
  journal={arXiv preprint arXiv:2305.18171},
  year={2023}
}

@inproceedings{rombach2022stablediffusion,
  title={High-resolution image synthesis with latent diffusion models},
  author={Rombach, Robin and Blattmann, Andreas and Lorenz, Dominik and Esser, Patrick and Ommer, Bj{\"o}rn},
  booktitle={Proceedings of the IEEE/CVF conference on computer vision and pattern recognition},
  pages={10684--10695},
  year={2022}
}

@article{kailkhura2019reliablex,
  title={Reliable and explainable machine-learning methods for accelerated material discovery},
  author={Kailkhura, Bhavya and Gallagher, Brian and Kim, Sookyung and Hiszpanski, Anna and Han, T Yong-Jin},
  journal={npj Computational Materials},
  volume={5},
  number={1},
  pages={108},
  year={2019},
  publisher={Nature Publishing Group UK London}
}

@article{park2022mixupcalibration,
  title={On the calibration of pre-trained language models using mixup guided by area under the margin and saliency},
  author={Park, Seo Yeon and Caragea, Cornelia},
  journal={arXiv preprint arXiv:2203.07559},
  year={2022}
}

@article{He2023PreservingPF,
  title={Preserving Pre-trained Features Helps Calibrate Fine-tuned Language Models},
  author={Guande He and Jianfei Chen and Jun Zhu},
  journal={ArXiv},
  year={2023},
  volume={abs/2305.19249},
  url={https://api.semanticscholar.org/CorpusID:258967945}
}

@article{Zhang2021BayesianABN,
  title={Bayesian Attention Belief Networks},
  author={Shujian Zhang and Xinjie Fan and Bo Chen and Mingyuan Zhou},
  journal={ArXiv},
  year={2021},
  volume={abs/2106.05251},
  url={https://api.semanticscholar.org/CorpusID:235377150}
}

@article{Fan2020BayesianAM,
  title={Bayesian Attention Modules},
  author={Xinjie Fan and Shujian Zhang and Bo Chen and Mingyuan Zhou},
  journal={ArXiv},
  year={2020},
  volume={abs/2010.10604},
  url={https://api.semanticscholar.org/CorpusID:224814357}
}

@article{li2024fewshotrecalibrator,
  title={Few-Shot Recalibration of Language Models},
  author={Li, Xiang Lisa and Khandelwal, Urvashi and Guu, Kelvin},
  journal={arXiv preprint arXiv:2403.18286},
  year={2024}
}

@article{kumar2019nmtcalibration,
  title={Calibration of encoder decoder models for neural machine translation},
  author={Kumar, Aviral and Sarawagi, Sunita},
  journal={arXiv preprint arXiv:1903.00802},
  year={2019}
}

@inproceedings{kong-etal-2020-calibrated-lm-fine-tuning,
    title = "Calibrated Language Model Fine-Tuning for In- and Out-of-Distribution Data",
    author = "Kong, Lingkai  and
      Jiang, Haoming  and
      Zhuang, Yuchen  and
      Lyu, Jie  and
      Zhao, Tuo  and
      Zhang, Chao",
    editor = "Webber, Bonnie  and
      Cohn, Trevor  and
      He, Yulan  and
      Liu, Yang",
    booktitle = "Proceedings of the 2020 Conference on Empirical Methods in Natural Language Processing (EMNLP)",
    month = nov,
    year = "2020",
    address = "Online",
    publisher = "Association for Computational Linguistics",
    url = "https://aclanthology.org/2020.emnlp-main.102",
    doi = "10.18653/v1/2020.emnlp-main.102",
    pages = "1326--1340",
}

@article{mukhoti2020calibratingwithfocalloss,
  title={Calibrating deep neural networks using focal loss},
  author={Mukhoti, Jishnu and Kulharia, Viveka and Sanyal, Amartya and Golodetz, Stuart and Torr, Philip and Dokania, Puneet},
  journal={Advances in Neural Information Processing Systems},
  volume={33},
  pages={15288--15299},
  year={2020}
}

@misc{gupta2021calibsplinefitting,
      title={Calibration of Neural Networks using Splines}, 
      author={Kartik Gupta and Amir Rahimi and Thalaiyasingam Ajanthan and Thomas Mensink and Cristian Sminchisescu and Richard Hartley},
      year={2021},
      eprint={2006.12800},
      archivePrefix={arXiv},
      primaryClass={cs.LG},
      url={https://arxiv.org/abs/2006.12800}, 
}

@article{wang2020transcal,
  title={Transferable calibration with lower bias and variance in domain adaptation},
  author={Wang, Ximei and Long, Mingsheng and Wang, Jianmin and Jordan, Michael},
  journal={Advances in Neural Information Processing Systems},
  volume={33},
  pages={19212--19223},
  year={2020}
}

@article{krishnan2020improving,
  title={Improving model calibration with accuracy versus uncertainty optimization},
  author={Krishnan, Ranganath and Tickoo, Omesh},
  journal={Advances in Neural Information Processing Systems},
  volume={33},
  pages={18237--18248},
  year={2020}
}

@inproceedings{Braverman2019CalibrationER,
  title={Calibration, Entropy Rates, and Memory in Language Models},
  author={Mark Braverman and Xinyi Chen and Sham M. Kakade and Karthik Narasimhan and Cyril Zhang and Yi Zhang},
  booktitle={International Conference on Machine Learning},
  year={2019},
  url={https://api.semanticscholar.org/CorpusID:189762078}
}

@article{kull2019beyond,
  title={Beyond temperature scaling: Obtaining well-calibrated multi-class probabilities with dirichlet calibration},
  author={Kull, Meelis and Perello Nieto, Miquel and K{\"a}ngsepp, Markus and Silva Filho, Telmo and Song, Hao and Flach, Peter},
  journal={Advances in neural information processing systems},
  volume={32},
  year={2019}
}

@article{kull2017betacalib,
  title={Beyond sigmoids: How to obtain well-calibrated probabilities from binary classifiers with beta calibration},
  author={Kull, Meelis and Silva Filho, Telmo M and Flach, Peter},
  year={2017}
}

@inproceedings{naeini2015bbq,
  title={Obtaining well calibrated probabilities using bayesian binning},
  author={Naeini, Mahdi Pakdaman and Cooper, Gregory and Hauskrecht, Milos},
  booktitle={Proceedings of the AAAI conference on artificial intelligence},
  volume={29},
  number={1},
  year={2015}
}

@article{kuleshov2015calibrated,
  title={Calibrated structured prediction},
  author={Kuleshov, Volodymyr and Liang, Percy S},
  journal={Advances in Neural Information Processing Systems},
  volume={28},
  year={2015}
}

@article{gai2023blockchainllm,
  title={Blockchain large language models},
  author={Gai, Yu and Zhou, Liyi and Qin, Kaihua and Song, Dawn and Gervais, Arthur},
  journal={arXiv preprint arXiv:2304.12749},
  year={2023}
}

@article{blasiok2023smoothece,
  title={Smooth ECE: Principled reliability diagrams via kernel smoothing},
  author={B{\l}asiok, Jaros{\l}aw and Nakkiran, Preetum},
  journal={arXiv preprint arXiv:2309.12236},
  year={2023}
}

@article{kirchhof2023url,
  title={URL: A Representation Learning Benchmark for Transferable Uncertainty Estimates},
  author={Kirchhof, Michael and Mucs{\'a}nyi, B{\'a}lint and Oh, Seong Joon and Kasneci, Enkelejda},
  journal={arXiv preprint arXiv:2307.03810},
  year={2023}
}

@article{hendrycks2016aurocbaseline,
  title={A baseline for detecting misclassified and out-of-distribution examples in neural networks},
  author={Hendrycks, Dan and Gimpel, Kevin},
  journal={arXiv preprint arXiv:1610.02136},
  year={2016}
}

@inproceedings{ott2018analyzinguqnmt,
  title={Analyzing uncertainty in neural machine translation},
  author={Ott, Myle and Auli, Michael and Grangier, David and Ranzato, Marc’Aurelio},
  booktitle={International Conference on Machine Learning},
  pages={3956--3965},
  year={2018},
  organization={PMLR}
}

@inproceedings{papineni2002bleu,
author = {Papineni, Kishore and Roukos, Salim and Ward, Todd and Zhu, Wei-Jing},
title = {BLEU: a method for automatic evaluation of machine translation},
year = {2002},
publisher = {Association for Computational Linguistics},
address = {USA},
url = {https://doi.org/10.3115/1073083.1073135},
doi = {10.3115/1073083.1073135},
booktitle = {Proceedings of the 40th Annual Meeting on Association for Computational Linguistics},
pages = {311–318},
numpages = {8},
location = {Philadelphia, Pennsylvania},
series = {ACL '02}
}

@article{brocker2009reliabilitysufficiency,
  title={Reliability, sufficiency, and the decomposition of proper scores},
  author={Br{\"o}cker, Jochen},
  journal={Quarterly Journal of the Royal Meteorological Society: A journal of the atmospheric sciences, applied meteorology and physical oceanography},
  volume={135},
  number={643},
  pages={1512--1519},
  year={2009},
  publisher={Wiley Online Library}
}

@article{brier1950scoreverification,
  title={Verification of forecasts expressed in terms of probability},
  author={Brier, Glenn W},
  journal={Monthly weather review},
  volume={78},
  number={1},
  pages={1--3},
  year={1950}
}

@article{lin2021truthfulqa,
  title={Truthfulqa: Measuring how models mimic human falsehoods},
  author={Lin, Stephanie and Hilton, Jacob and Evans, Owain},
  journal={arXiv preprint arXiv:2109.07958},
  year={2021}
}

@article{kwiatkowski-etal-2019-nq,
    title = "Natural Questions: A Benchmark for Question Answering Research",
    author = "Kwiatkowski, Tom  and
      Palomaki, Jennimaria  and
      Redfield, Olivia  and
      Collins, Michael  and
      Parikh, Ankur  and
      Alberti, Chris  and
      Epstein, Danielle  and
      Polosukhin, Illia  and
      Devlin, Jacob  and
      Lee, Kenton  and
      Toutanova, Kristina  and
      Jones, Llion  and
      Kelcey, Matthew  and
      Chang, Ming-Wei  and
      Dai, Andrew M.  and
      Uszkoreit, Jakob  and
      Le, Quoc  and
      Petrov, Slav",
    editor = "Lee, Lillian  and
      Johnson, Mark  and
      Roark, Brian  and
      Nenkova, Ani",
    journal = "Transactions of the Association for Computational Linguistics",
    volume = "7",
    year = "2019",
    address = "Cambridge, MA",
    publisher = "MIT Press",
    url = "https://aclanthology.org/Q19-1026/",
    doi = "10.1162/tacl_a_00276",
    pages = "452--466"
}

@article{yin2023llmknowwhattheydont,
  title={Do Large Language Models Know What They Don't Know?},
  author={Yin, Zhangyue and Sun, Qiushi and Guo, Qipeng and Wu, Jiawen and Qiu, Xipeng and Huang, Xuanjing},
  journal={arXiv preprint arXiv:2305.18153},
  year={2023}
}

@article{joshi2017triviaqa,
  title={Triviaqa: A large scale distantly supervised challenge dataset for reading comprehension},
  author={Joshi, Mandar and Choi, Eunsol and Weld, Daniel S and Zettlemoyer, Luke},
  journal={arXiv preprint arXiv:1705.03551},
  year={2017}
}

@article{guo2025deepseekrone,
  title={Deepseek-r1: Incentivizing reasoning capability in llms via reinforcement learning},
  author={Guo, Daya and Yang, Dejian and Zhang, Haowei and Song, Junxiao and Zhang, Ruoyu and Xu, Runxin and Zhu, Qihao and Ma, Shirong and Wang, Peiyi and Bi, Xiao and others},
  journal={arXiv preprint arXiv:2501.12948},
  year={2025}
}

@article{dubey2024llama3.1,
  title={The llama 3 herd of models},
  author={Dubey, Abhimanyu and Jauhri, Abhinav and Pandey, Abhinav and Kadian, Abhishek and Al-Dahle, Ahmad and Letman, Aiesha and Mathur, Akhil and Schelten, Alan and Yang, Amy and Fan, Angela and others},
  journal={arXiv preprint arXiv:2407.21783},
  year={2024}
}

@article{touvron2023llama2,
  title={Llama 2: Open foundation and fine-tuned chat models},
  author={Touvron, Hugo and Martin, Louis and Stone, Kevin and Albert, Peter and Almahairi, Amjad and Babaei, Yasmine and Bashlykov, Nikolay and Batra, Soumya and Bhargava, Prajjwal and Bhosale, Shruti and others},
  journal={arXiv preprint arXiv:2307.09288},
  year={2023}
}

@article{yang2024qwen2.5,
  title={Qwen2. 5 Technical Report},
  author={Yang, An and Yang, Baosong and Zhang, Beichen and Hui, Binyuan and Zheng, Bo and Yu, Bowen and Li, Chengyuan and Liu, Dayiheng and Huang, Fei and Wei, Haoran and others},
  journal={arXiv preprint arXiv:2412.15115},
  year={2024}
}

@article{jiang2023mistral,
  title={Mistral 7B},
  author={Jiang, Albert Q and Sablayrolles, Alexandre and Mensch, Arthur and Bamford, Chris and Chaplot, Devendra Singh and Casas, Diego de las and Bressand, Florian and Lengyel, Gianna and Lample, Guillaume and Saulnier, Lucile and others},
  journal={arXiv preprint arXiv:2310.06825},
  year={2023}
}

@article{zhang2022opt,
  title={Opt: Open pre-trained transformer language models},
  author={Zhang, Susan and Roller, Stephen and Goyal, Naman and Artetxe, Mikel and Chen, Moya and Chen, Shuohui and Dewan, Christopher and Diab, Mona and Li, Xian and Lin, Xi Victoria and others},
  journal={arXiv preprint arXiv:2205.01068},
  year={2022}
}

@article{raffel2020tfivettt,
  title={Exploring the limits of transfer learning with a unified text-to-text transformer},
  author={Raffel, Colin and Shazeer, Noam and Roberts, Adam and Lee, Katherine and Narang, Sharan and Matena, Michael and Zhou, Yanqi and Li, Wei and Liu, Peter J},
  journal={Journal of machine learning research},
  volume={21},
  number={140},
  pages={1--67},
  year={2020}
}

@article{hurst2024gpt4o,
  title={Gpt-4o system card},
  author={Hurst, Aaron and Lerer, Adam and Goucher, Adam P and Perelman, Adam and Ramesh, Aditya and Clark, Aidan and Ostrow, AJ and Welihinda, Akila and Hayes, Alan and Radford, Alec and others},
  journal={arXiv preprint arXiv:2410.21276},
  year={2024}
}

@article{achiam2023gpt4,
  title={Gpt-4 technical report},
  author={Achiam, Josh and Adler, Steven and Agarwal, Sandhini and Ahmad, Lama and Akkaya, Ilge and Aleman, Florencia Leoni and Almeida, Diogo and Altenschmidt, Janko and Altman, Sam and Anadkat, Shyamal and others},
  journal={arXiv preprint arXiv:2303.08774},
  year={2023}
}

@article{brown2020languagegpt3,
  title={Language models are few-shot learners},
  author={Brown, Tom and Mann, Benjamin and Ryder, Nick and Subbiah, Melanie and Kaplan, Jared D and Dhariwal, Prafulla and Neelakantan, Arvind and Shyam, Pranav and Sastry, Girish and Askell, Amanda and others},
  journal={Advances in neural information processing systems},
  volume={33},
  pages={1877--1901},
  year={2020}
}

@article{rafailov2023dpo,
  title={Direct preference optimization: Your language model is secretly a reward model},
  author={Rafailov, Rafael and Sharma, Archit and Mitchell, Eric and Manning, Christopher D and Ermon, Stefano and Finn, Chelsea},
  journal={Advances in Neural Information Processing Systems},
  volume={36},
  pages={53728--53741},
  year={2023}
}

@article{hu2021lora,
  title={Lora: Low-rank adaptation of large language models},
  author={Hu, Edward J and Shen, Yelong and Wallis, Phillip and Allen-Zhu, Zeyuan and Li, Yuanzhi and Wang, Shean and Wang, Lu and Chen, Weizhu},
  journal={arXiv preprint arXiv:2106.09685},
  year={2021}
}

@article{dong2025confcalibsurvey,
  title={A Survey on Confidence Calibration of Deep Learning-Based Classification Models Under Class Imbalance Data},
  author={Dong, Jinzong and Jiang, Zhaohui and Pan, Dong and Chen, Zhiwen and Guan, Qingyi and Zhang, Hongbin and Gui, Gui and Gui, Weihua},
  journal={IEEE Transactions on Neural Networks and Learning Systems},
  volume={36},
  number={9},
  pages={15664--15684},
  year={2025},
  publisher={IEEE},
  doi={10.1109/TNNLS.2025.3565159}
}

@article{wang2025conformalkeypoint,
  title={Satellite Pose Set Estimation by Uncertainty-Guided Conformal Keypoint Detection},
  author={Wang, Jinghao and Li, Zhang and Sun, Cong and Guo, Yulan and Wang, Zi and Yu, Qifeng},
  journal={IEEE Transactions on Neural Networks and Learning Systems},
  volume={36},
  number={12},
  pages={20120--20132},
  year={2025},
  publisher={IEEE},
  doi={10.1109/TNNLS.2025.3598481}
}

@article{zhao2024oodcrossclass,
  title={Out-of-Distribution Detection by Cross-Class Vicinity Distribution of In-Distribution Data},
  author={Zhao, Zhilin and Cao, Longbing and Lin, Kun-Yu},
  journal={IEEE Transactions on Neural Networks and Learning Systems},
  volume={35},
  number={10},
  pages={13777--13788},
  year={2024},
  publisher={IEEE},
  doi={10.1109/TNNLS.2023.3271856}
}
\endgroup


\appendices

\section{Additional Experimental Settings and Results}
\label{appendix:exp}

\begin{figure*}[!ht]
    \centering
    \subfloat[Few-shot]{
        \includegraphics[width=0.188\textwidth]{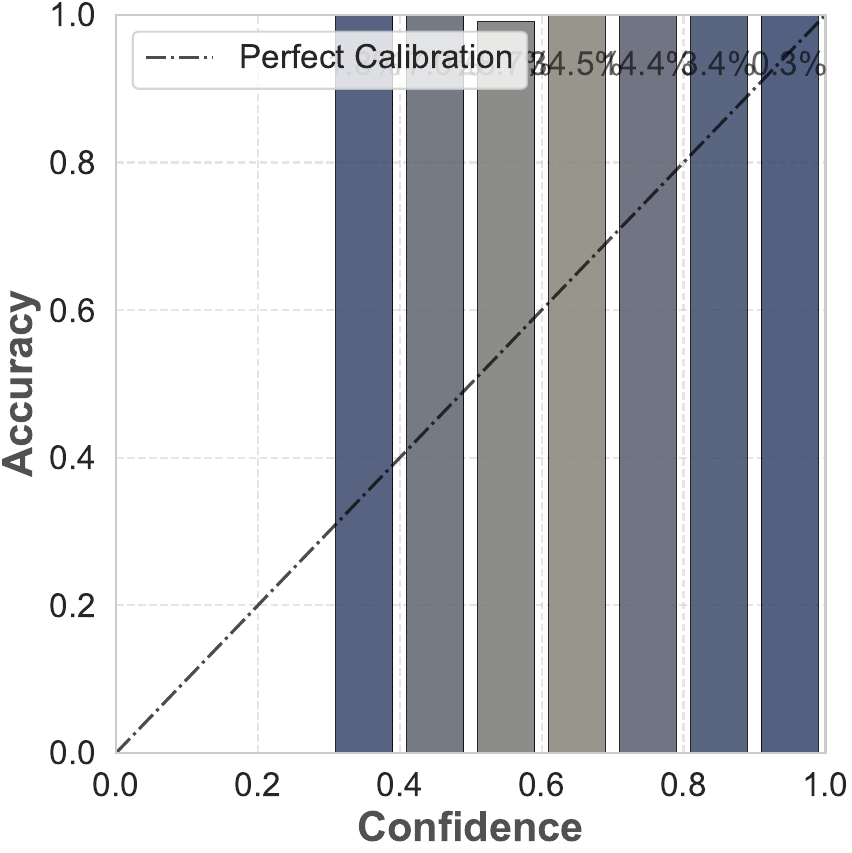}
        \label{fig:seq_likelihood_llama_truthful_qa}}
    \subfloat[Few-Shot CoT]{
        \includegraphics[width=0.188\textwidth]{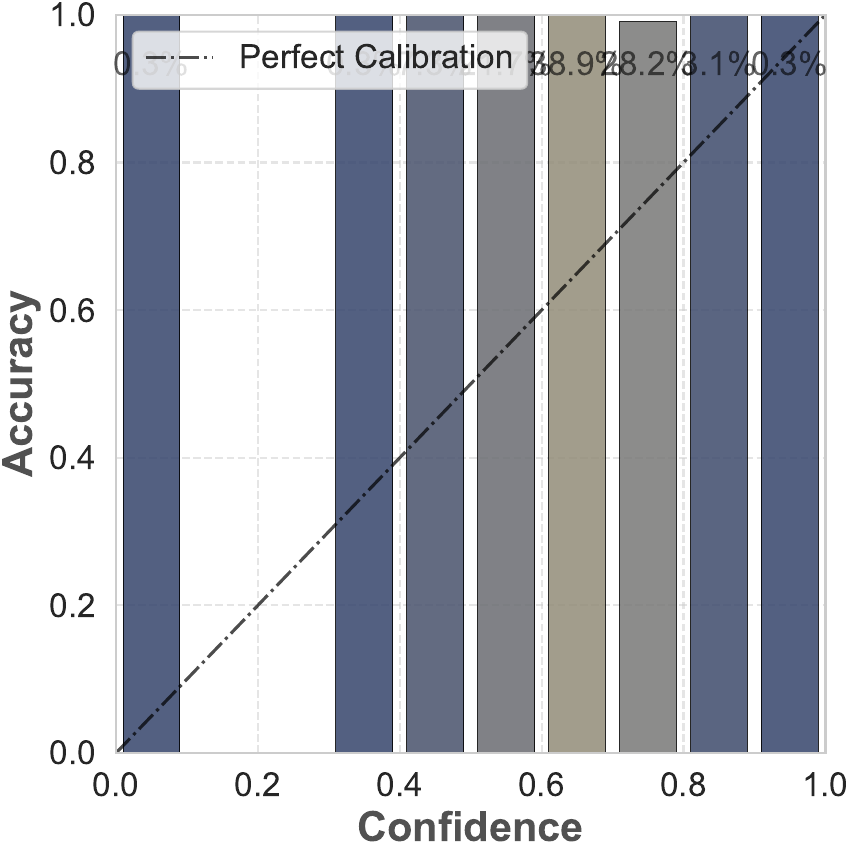}%
        \label{fig:cot_seq_likelihood_llama_truthful_qa}
    }
    \subfloat[Verbalized CoT]{
        \includegraphics[width=0.188\textwidth]{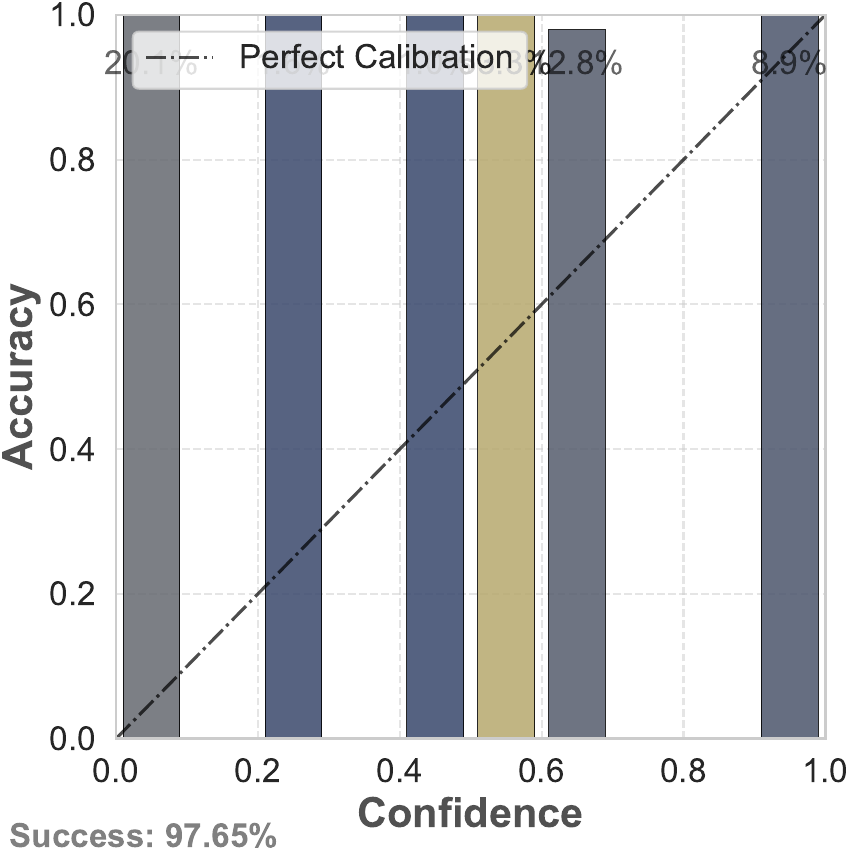}%
        \label{fig:verbalized_cot_qual_llama_truthful_qa}
    }
    \subfloat[Platt Scaling]{
        \includegraphics[width=0.188\textwidth]{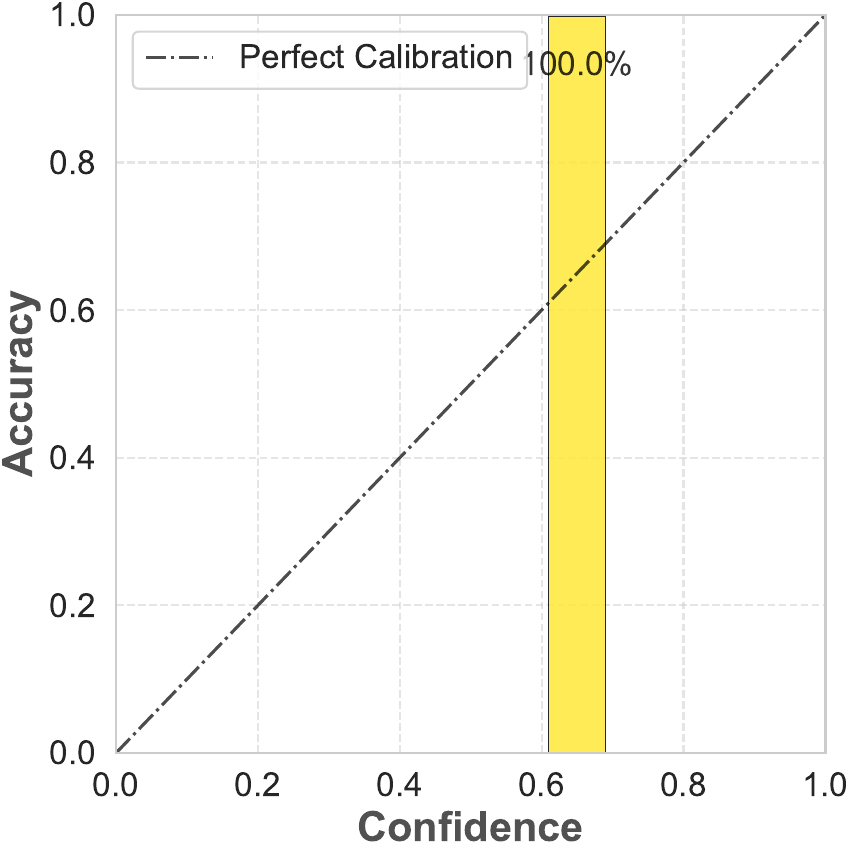}
        \label{fig:ps_seq_likelihood_llama_truthful_qa}
    }
    \subfloat[Temp. Scaling]{
        \includegraphics[width=0.188\textwidth]{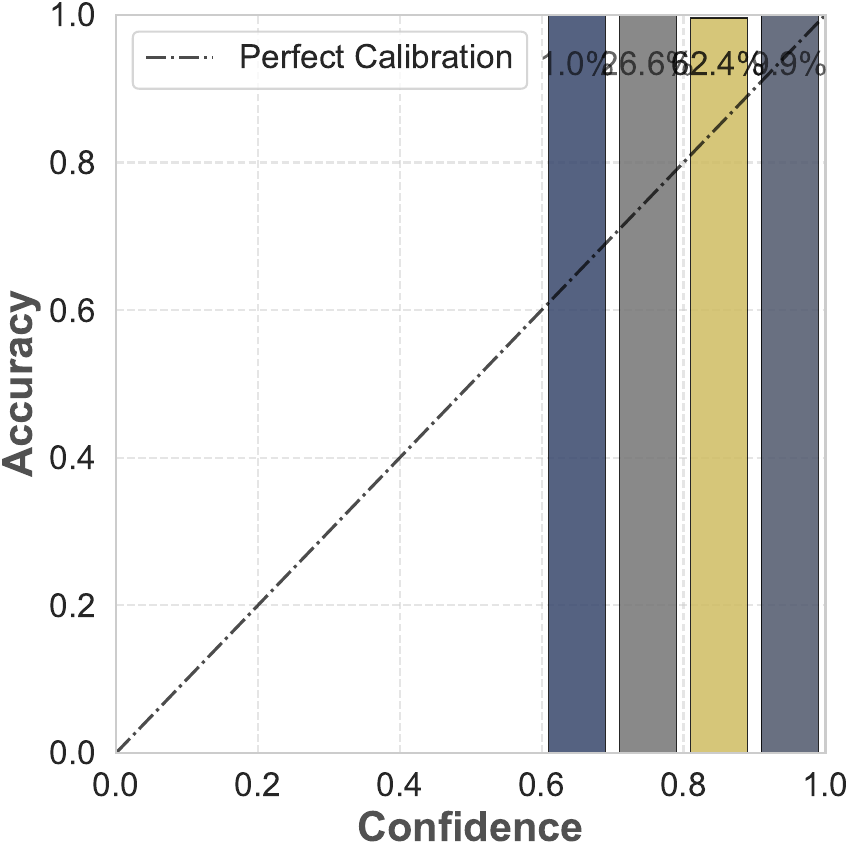}
        \label{fig:ts_seq_likelihood_llama_truthful_qa}
    }
    \caption{Reliability diagrams for different calibration methods with 10 bins using Llama 3.1 8B on TruthfulQA. The number of diagram bins is 10. The color and the percentage number on each bar indicate the proportion of total points contained in each bin. }
    \label{fig:llama_truthful_qa_comparison}
\end{figure*}

\begin{figure*}[!ht]
    \centering
    \subfloat[Few-shot]{
        \includegraphics[width=0.188\textwidth]{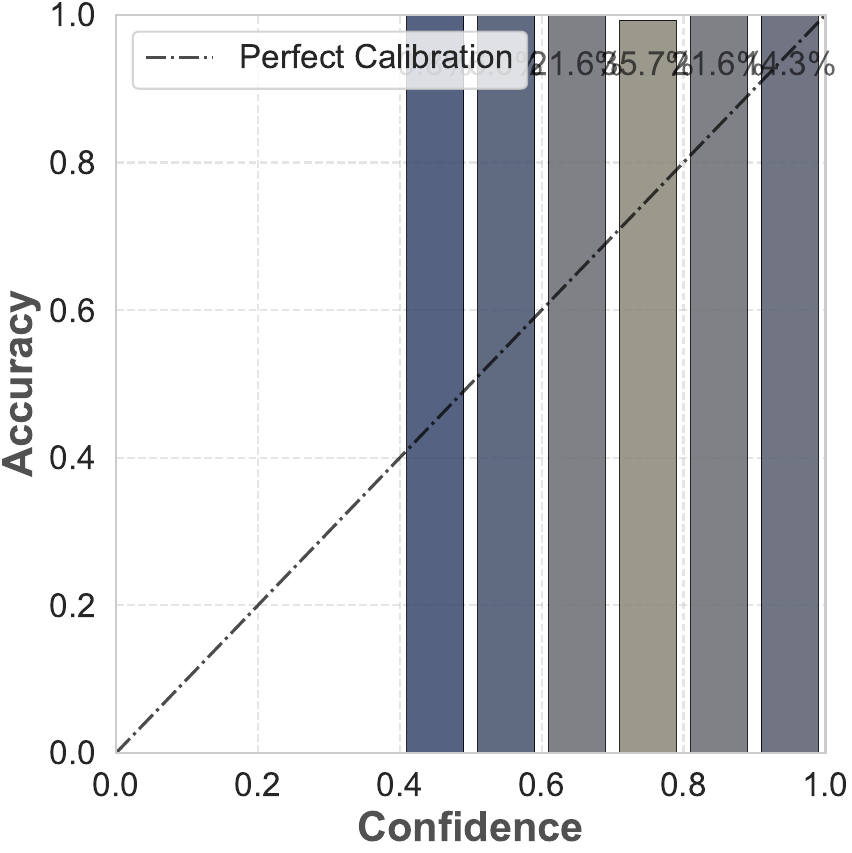}
        \label{fig:seq_likelihood_mistral_truthful_qa}}
    \subfloat[Few-Shot CoT]{
        \includegraphics[width=0.188\textwidth]{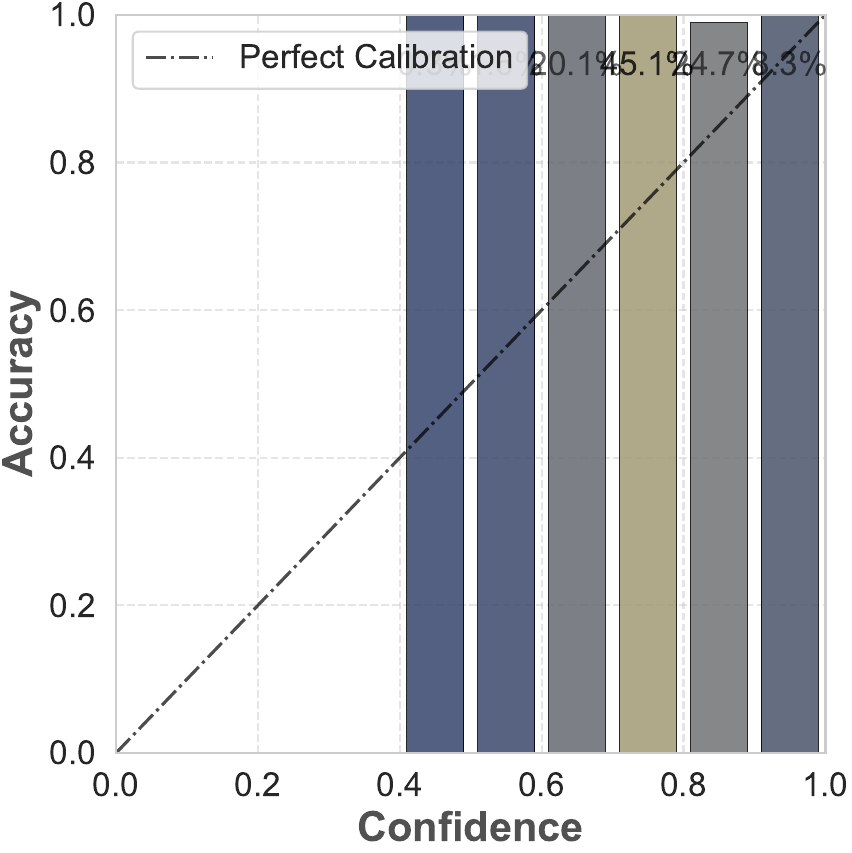}%
        \label{fig:cot_seq_likelihood_mistral_truthful_qa}
    }
    \subfloat[Verbalized CoT]{
        \includegraphics[width=0.188\textwidth]{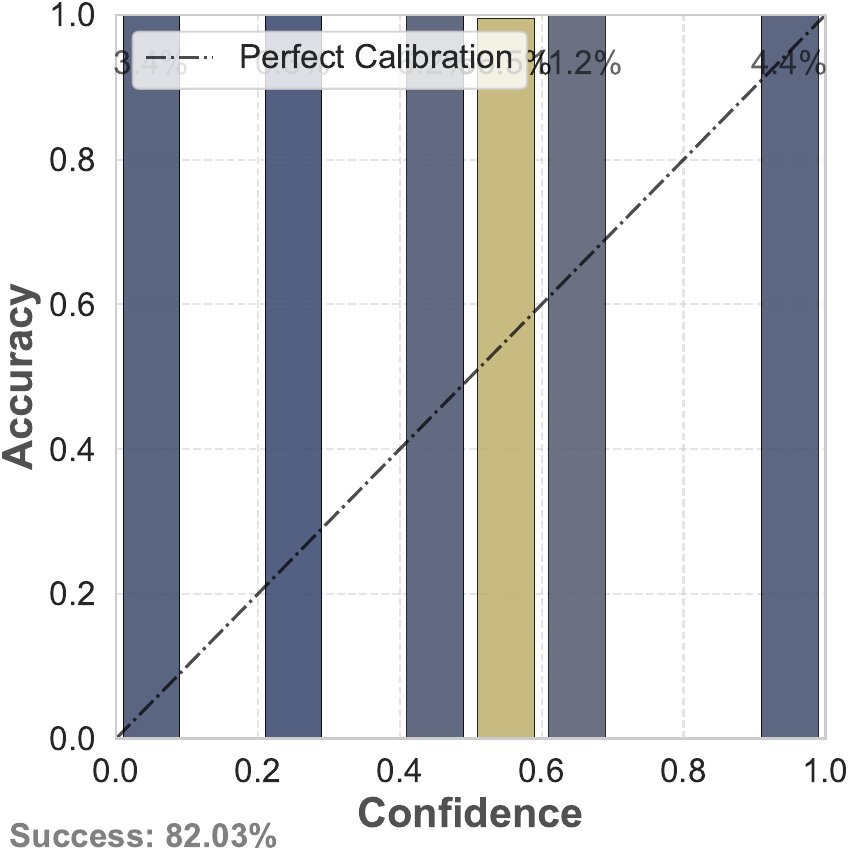}%
        \label{fig:verbalized_cot_qual_mistral_truthful_qa}
    }
    \subfloat[Platt Scaling]{
        \includegraphics[width=0.188\textwidth]{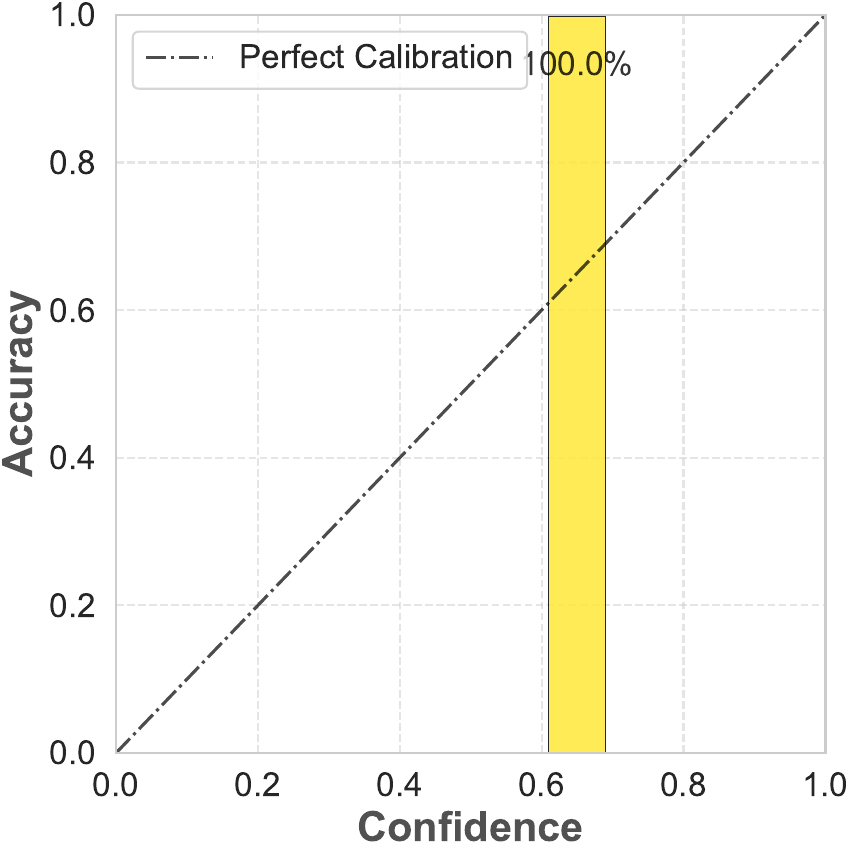}
        \label{fig:ps_seq_likelihood_mistral_truthful_qa}
    }
    \subfloat[Temp. Scaling]{
        \includegraphics[width=0.188\textwidth]{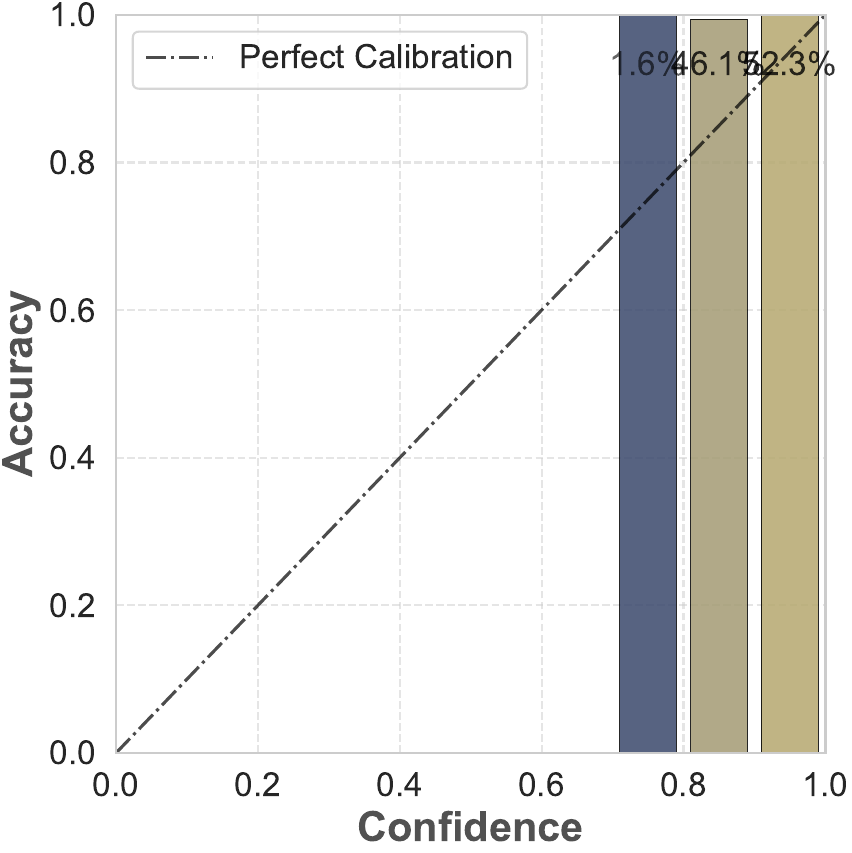}
        \label{fig:ts_seq_likelihood_mistral_truthful_qa}
    }
    \caption{Reliability diagrams for different calibration methods with 10 bins using Mistral v0.3 7B on TruthfulQA. The number of diagram bins is 10. The color and the percentage number on each bar indicate the proportion of total points contained in each bin. }
    \label{fig:mistral_truthful_qa_comparison}
\end{figure*}

\begin{figure*}[!ht]
    \centering
    \subfloat[Few-shot]{
        \includegraphics[width=0.188\textwidth]{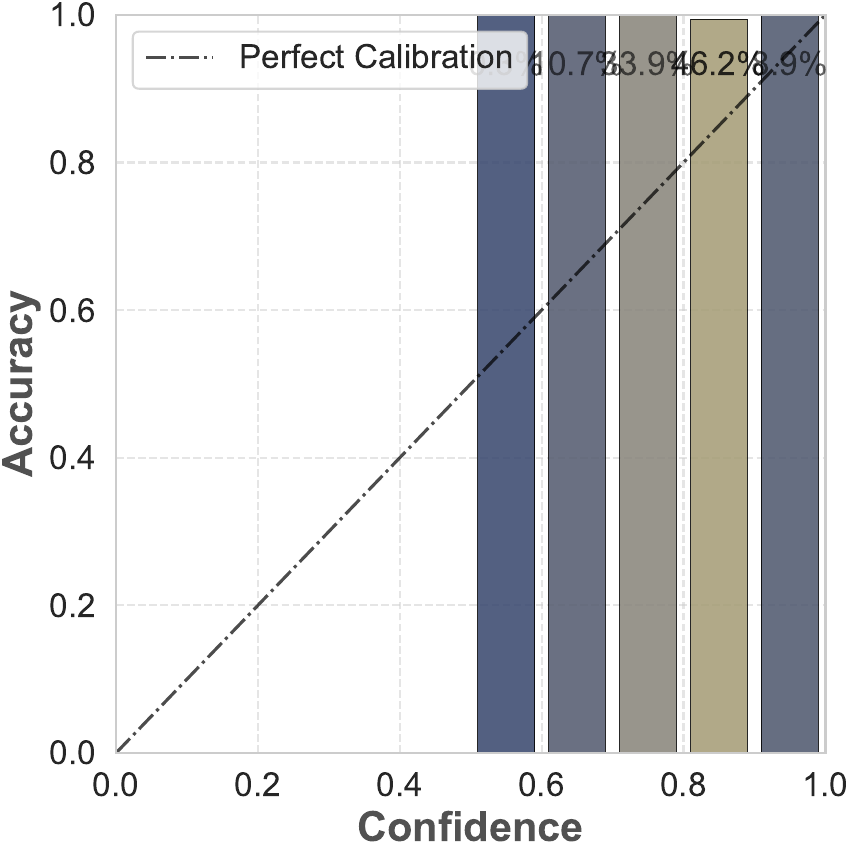}
        \label{fig:seq_likelihood_qwen2.5_truthful_qa}
    }
    \subfloat[Few-Shot CoT]{
        \includegraphics[width=0.188\textwidth]{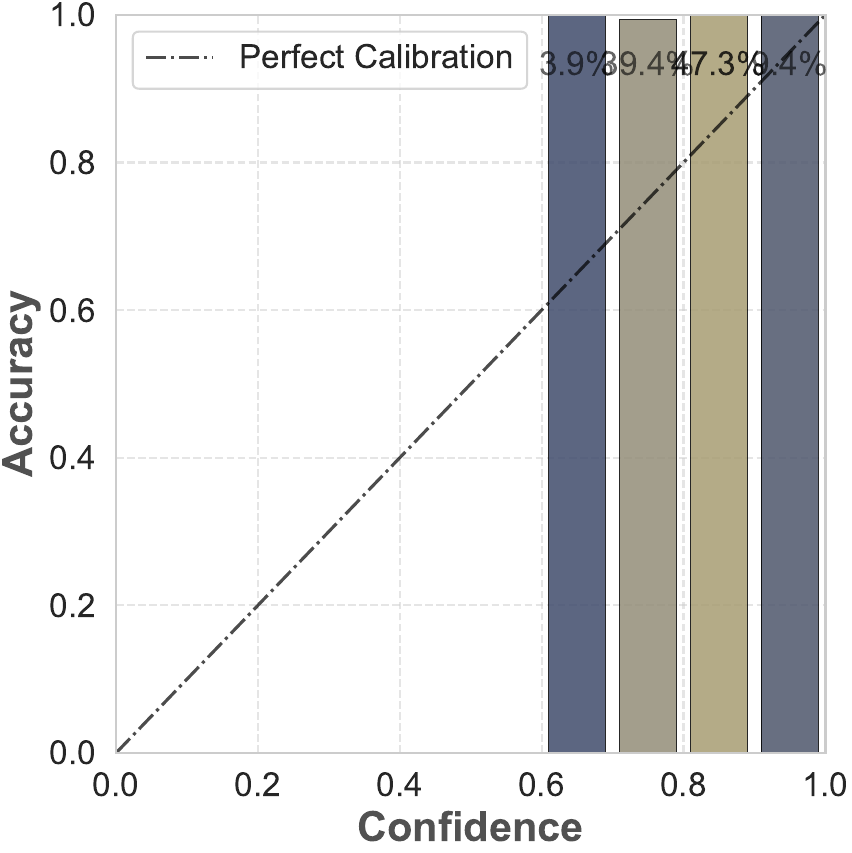}
        \label{fig:cot_seq_likelihood_qwen2.5_truthful_qa}
    }
    \subfloat[Verbalized CoT]{
        \includegraphics[width=0.188\textwidth]{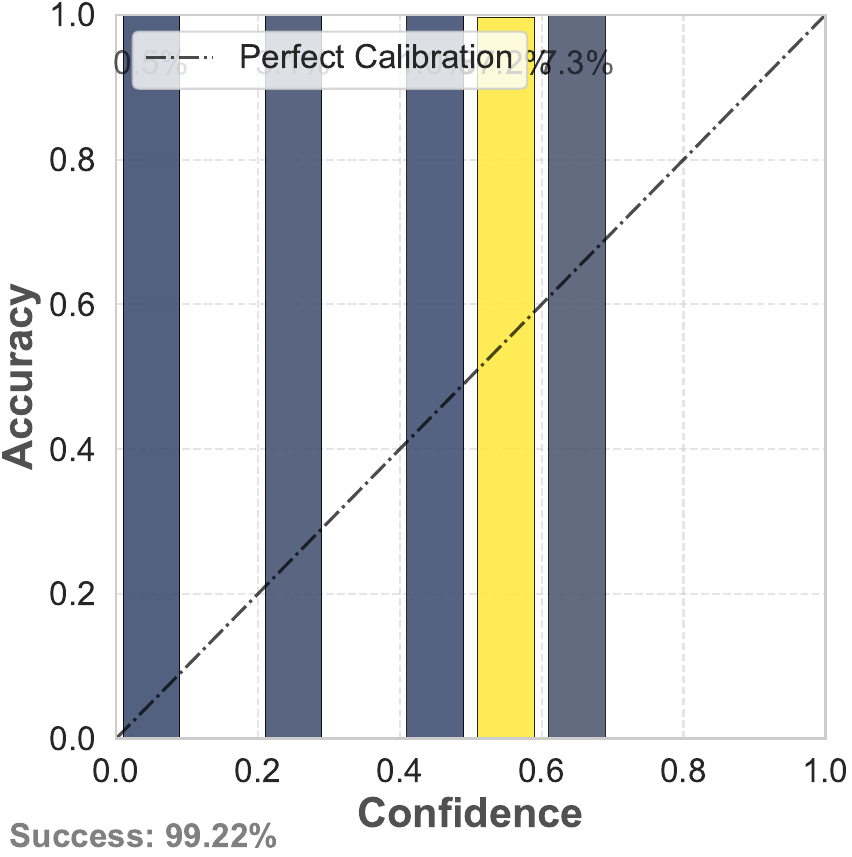}
        \label{fig:verbalized_cot_qual_qwen2.5_truthful_qa}
    }
    \subfloat[Platt Scaling]{
        \includegraphics[width=0.188\textwidth]{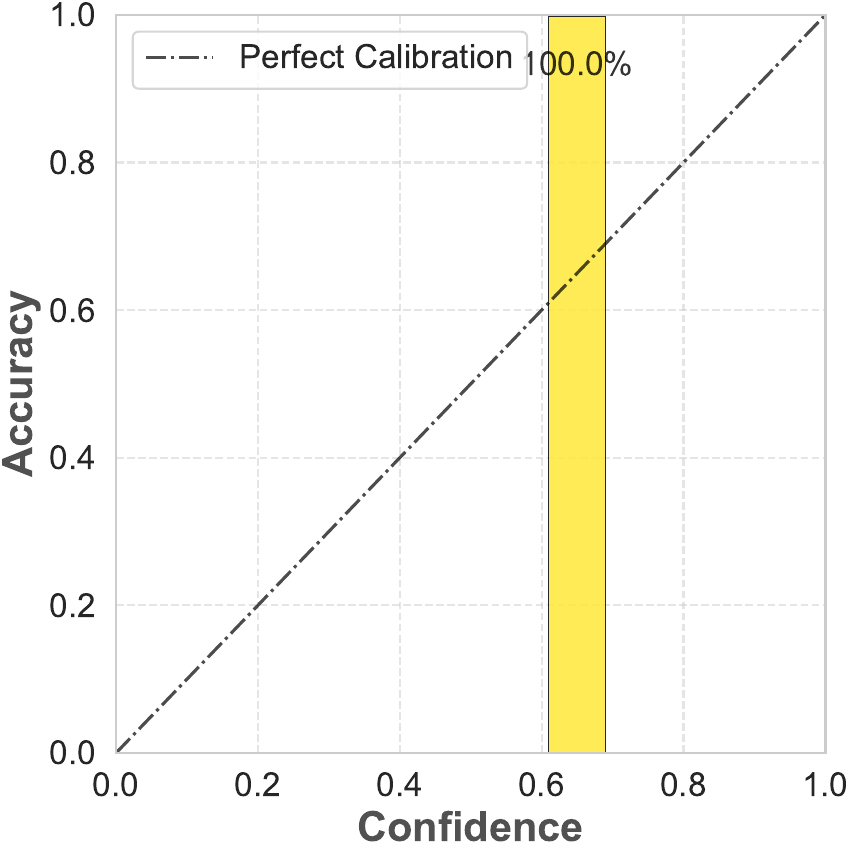}
        \label{fig:ps_seq_likelihood_qwen2.5_truthful_qa}
    }
    \subfloat[Temp. Scaling]{
        \includegraphics[width=0.188\textwidth]{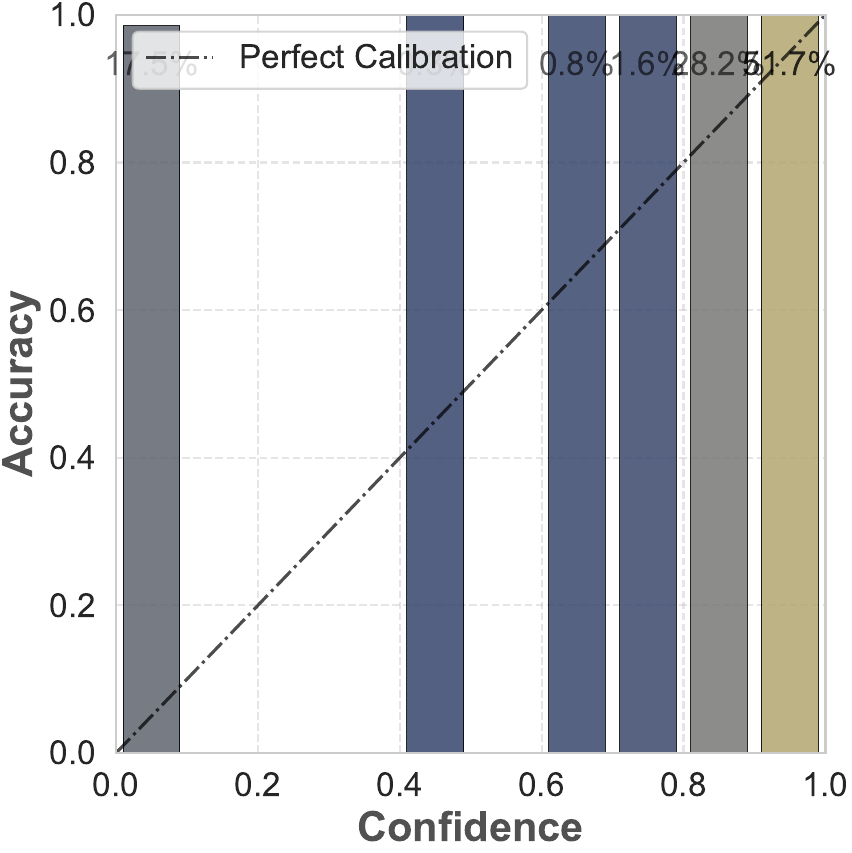}
        \label{fig:ts_seq_likelihood_qwen2.5_truthful_qa}
    }
    \caption{Reliability diagrams for calibration methods using Qwen2.5 7B on TruthfulQA. The number of diagram bins is 10. The color and the percentage number within each bar indicate the proportion of total points contained in each bin.}
    \label{fig:qwen2.5_truthful_qa_comparison}
\end{figure*}

\begin{figure*}[!ht]
    \centering
    \subfloat[Few-shot]{
        \includegraphics[width=0.188\textwidth]{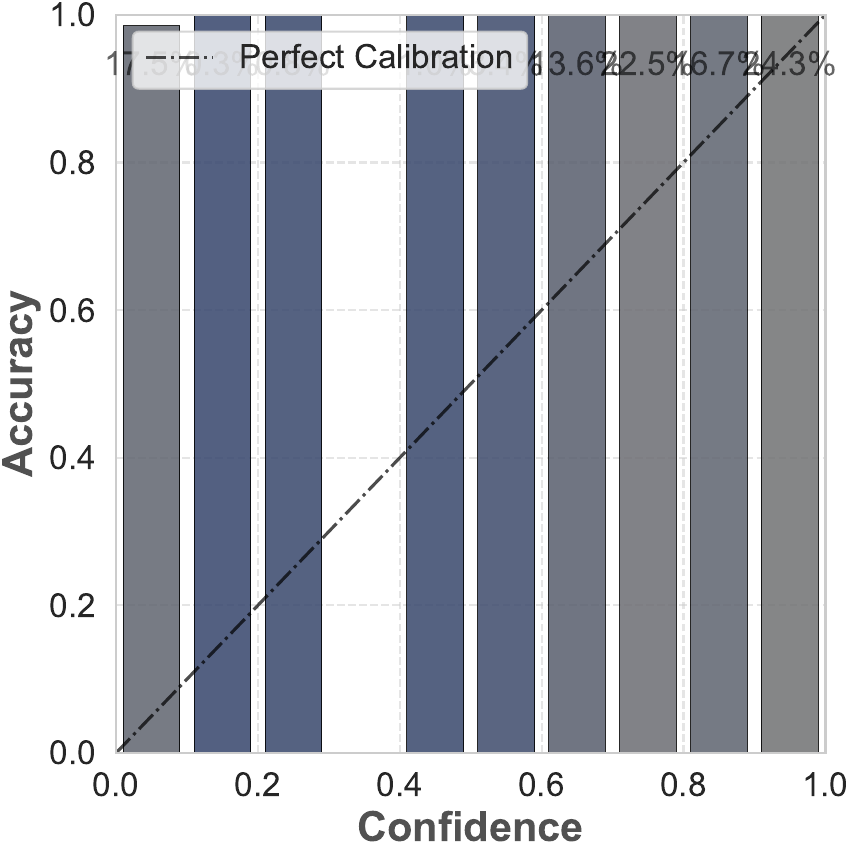}
        \label{fig:seq_likelihood_gpt4_truthful_qa}
    }
    \subfloat[Few-Shot CoT]{
        \includegraphics[width=0.188\textwidth]{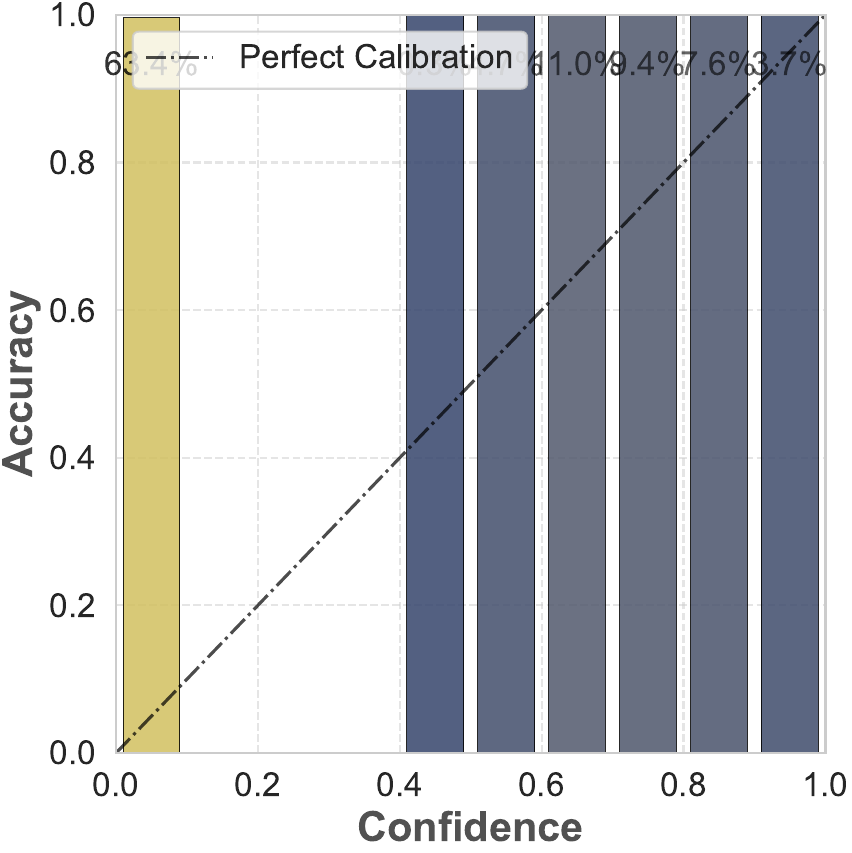}
        \label{fig:cot_seq_likelihood_gpt4_truthful_qa}
    }
    \subfloat[Verbalized CoT]{
        \includegraphics[width=0.188\textwidth]{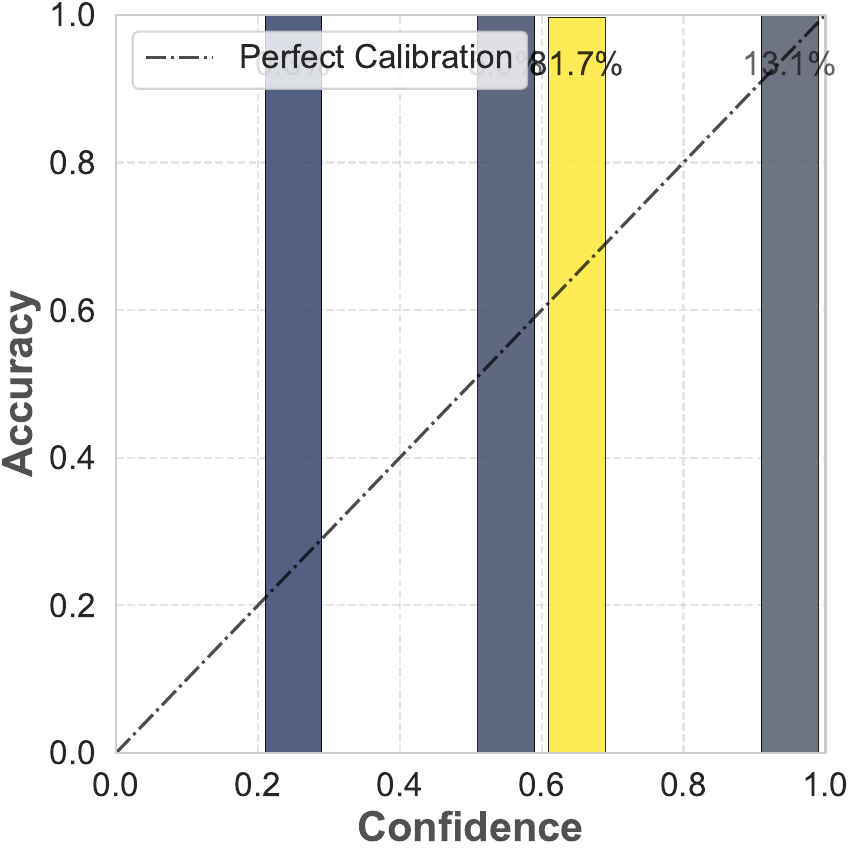}
        \label{fig:verbalized_cot_qual_gpt4_truthful_qa}
    }
    \subfloat[Platt Scaling]{
        \includegraphics[width=0.188\textwidth]{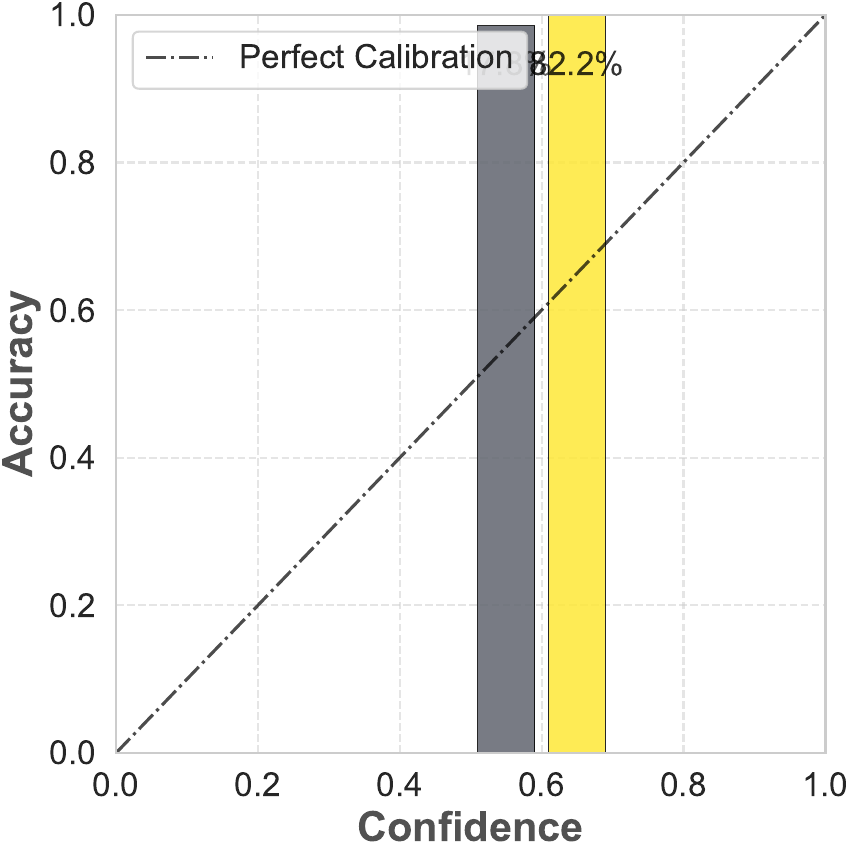}
        \label{fig:ps_seq_likelihood_gpt4_truthful_qa}
    }
    \subfloat[Temp. Scaling]{
        \includegraphics[width=0.188\textwidth]{figures/truthfulqa/test_ts_seq_likelihood_truthful_qa_gpt-4.pdf}
        \label{fig:ts_seq_likelihood_gpt4_truthful_qa}
    }
    \caption{Reliability diagrams for calibration methods using GPT-4 on TruthfulQA. The number of diagram bins is 10. The color as well as the percentage number within each bar indicate the proportion of total points contained in each bin. }
    \label{fig:gpt4_truthful_qa_comparison}
\end{figure*}

\begin{figure*}[!ht]
    \centering
    \subfloat[Few-shot]{
        \includegraphics[width=0.188\textwidth]{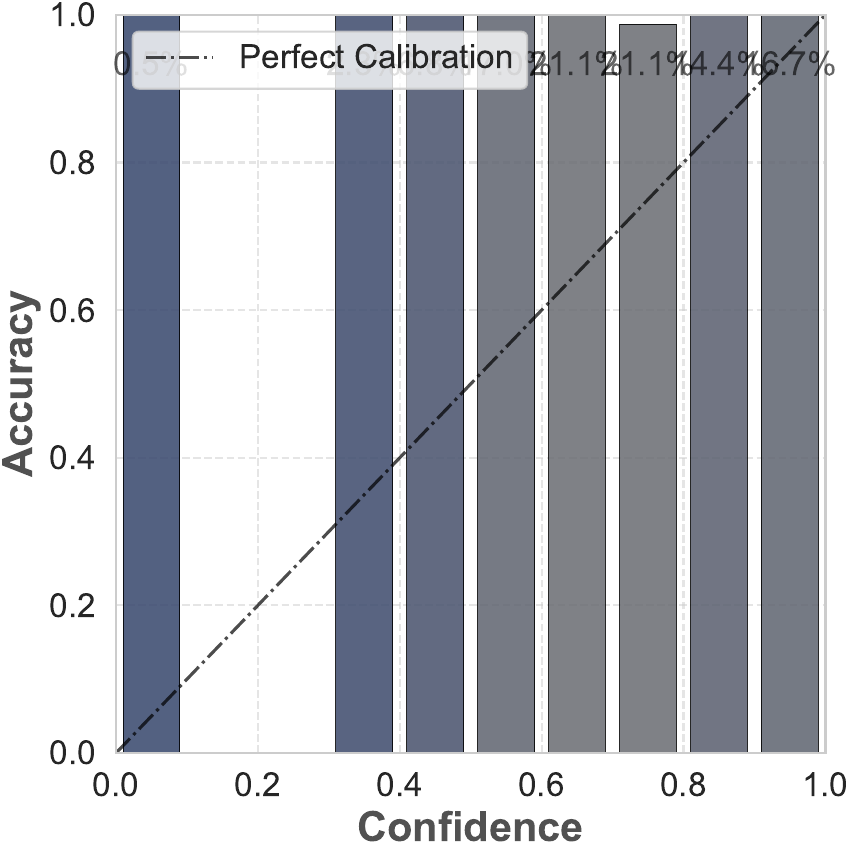}
        \label{fig:seq_likelihood_gpt4o_truthful_qa}
    }
    \subfloat[Few-Shot CoT]{
        \includegraphics[width=0.188\textwidth]{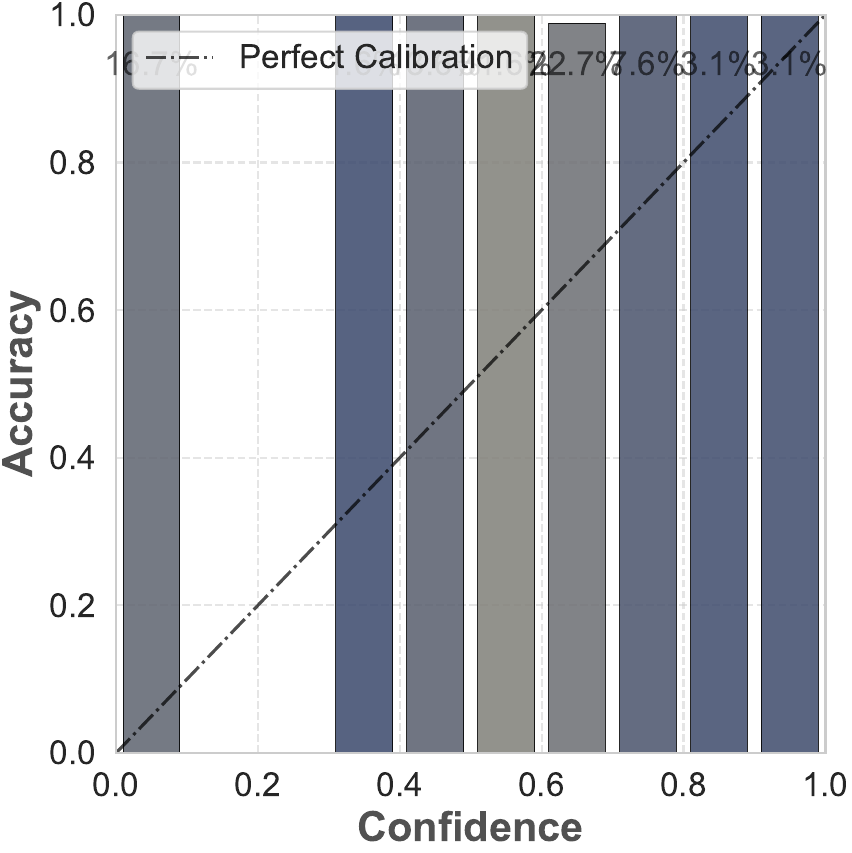}
        \label{fig:cot_seq_likelihood_gpt4o_truthful_qa}
    }
    \subfloat[Verbalized CoT]{
        \includegraphics[width=0.188\textwidth]{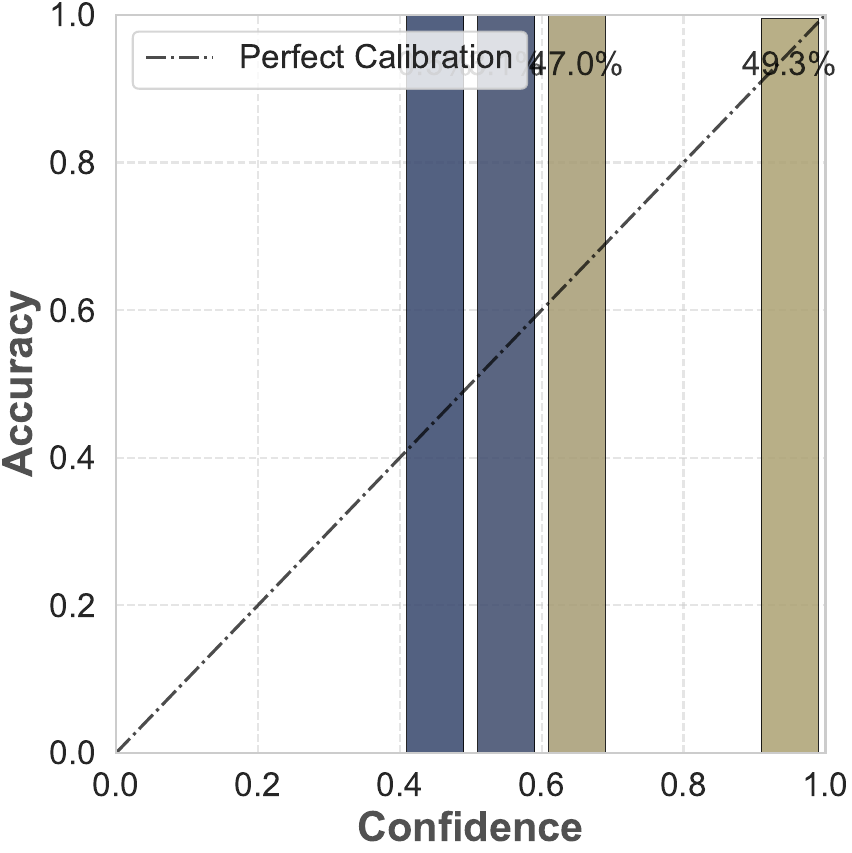}
        \label{fig:verbalized_cot_qual_gpt4o_truthful_qa}
    }
    \subfloat[Platt Scaling]{
        \includegraphics[width=0.188\textwidth]{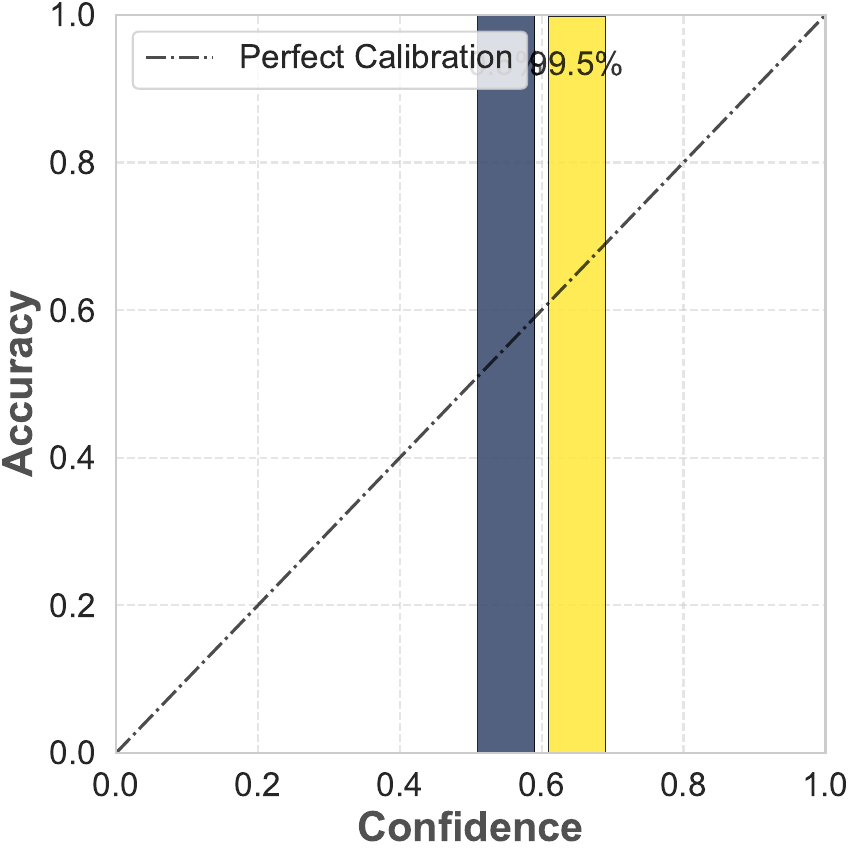}
        \label{fig:ps_seq_likelihood_gpt4o_truthful_qa}
    }
    \subfloat[Temp. Scaling]{
        \includegraphics[width=0.188\textwidth]{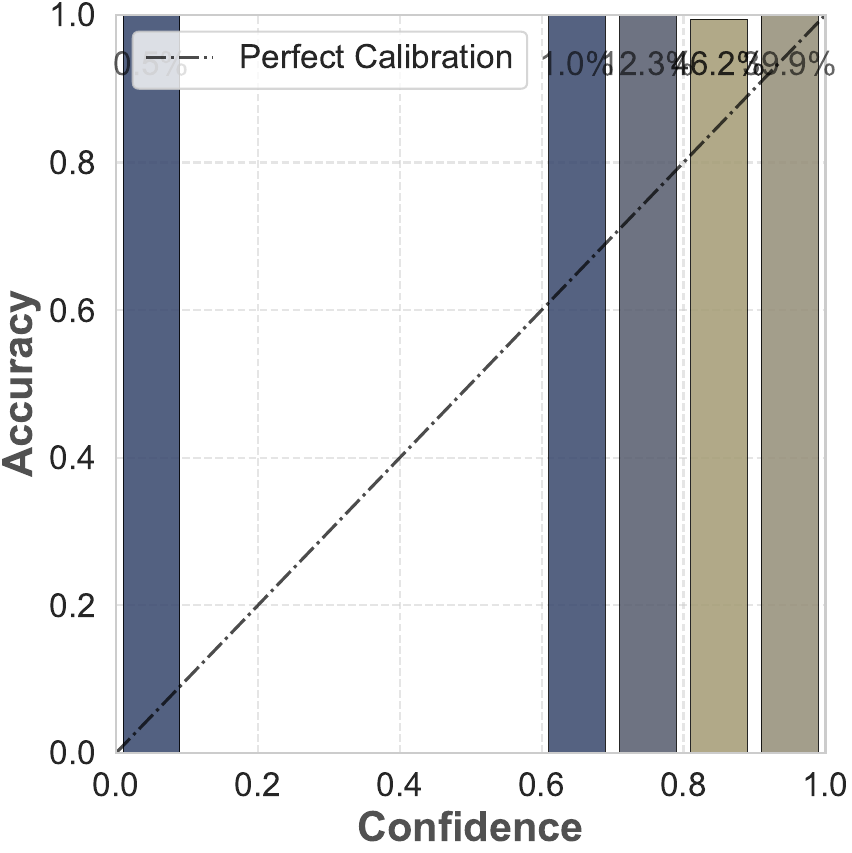}
        \label{fig:ts_seq_likelihood_gpt4o_truthful_qa}
    }
    \caption{Reliability diagrams for calibration methods using GPT-4o on TruthfulQA. The number of diagram bins is 10. The color as well as the percentage number within each bar indicate the proportion of total points contained in each bin. }
    \label{fig:gpt4o_truthful_qa_comparison}
\end{figure*}

\subsection{Datasets}
\label{appendix:exp_data}

\subsubsection{TriviaQA}
\label{appendix:exp_data_xxx}
The TriviaQA dataset \cite{joshi2017triviaqa} comprises trivia questions scraped from the web, featuring over 650K question-answer-evidence triples. Additionally, it includes 95K question-answer pairs authored by trivia enthusiasts. Following the setup in \cite{ulmer2024apricot}, we sample 12K examples for training and 1.5K examples for testing. As TriviaQA is well-suited for few-shot learning, we include 10 diverse in-context examples \cite{kuhn2023semanticuncertainty, farquhar2024detecting}.

\subsubsection{TruthfulQA}
\label{appendix:exp_data_yyy}
TruthfulQA \cite{lin2021truthfulqa} is a dataset designed to evaluate the ability of language models to answer questions truthfully, particularly when faced with common misconceptions and falsehoods. The questions in the dataset are crafted to reflect scenarios where some humans might provide incorrect answers due to false beliefs or misconceptions. To perform well, models must avoid generating false answers typically learned from imitating human-written texts. The test set comprises 817 questions across 38 categories, including health, law, finance, and politics.

In our experiments, we use a zero-shot setting where each question is paired with both true and false target answers, accompanied by supporting sources \cite{lin2021truthfulqa}. The dataset is evenly divided, with one half used for training and the other for testing. Additionally, since TriviaQA is well-suited for few-shot learning, we augment the training set by adding 10 different in-context examples to the questions.

\subsubsection{Natural Questions}
\label{appendix:exp_data_zzz}
NQ \cite{kwiatkowski-etal-2019-nq} is a large-scale, open-domain question answering dataset sourced from anonymized Google search queries paired with Wikipedia passages that contain the answers. It includes both long and short answer formats, reflecting the variability and complexity of real user information needs. For our evaluation, we sample 1,000 NQ questions, evenly dividing them into 500 for training and 500 for testing. To leverage few-shot capabilities—similar to our approach on TruthfulQA—we augment each training question with five in-context examples.

\subsection{Verbalized Qualitative Mapping}
\label{appendix:exp_verb_map}

Table~\ref{tab:verbalized_qual_map} presents the mapping of verbalized confidence expressions to their corresponding probabilistic values, enabling a clear relationship between verbalized confidence levels and numerical probabilities

\begin{table}[!t]
    \centering
    \caption{Mapping of verbalized qualitative confidence expressions to numerical probabilities used in the empirical evaluation.}
    \label{tab:verbalized_qual_map}
    \begin{tabular}{lc}
        \toprule
        \textbf{Verbal Expression} & \textbf{Numerical Probability} \\
        \midrule
        Very low       & 0.00 \\
        Low            & 0.30 \\
        Somewhat low   & 0.45 \\
        Medium         & 0.50 \\
        Somewhat high  & 0.65 \\
        High           & 0.70 \\
        Very high      & 1.00 \\
        \bottomrule
    \end{tabular}
\end{table}

\subsection{Evaluation Metrics}
\label{appendix:exp_eval}

\subsubsection{ECE}
\label{appendix:ece}
ECE can be calculated as the difference between bi$n$-wise accuracy and confidence of the model, and can be calculated as:

\begin{equation}
    \text{ECE} = \sum^{k}_{i=1} \frac{\abs{B_{i}}}{N} \abs{\text{acc}(B_i) - \text{conf}(B_i) },
    \label{eq:ece}
\end{equation}

where $N$ is the number of samples and $k$ is the number of bins or intervals $B_i = \Large[\frac{i - 1}{k}, \frac{i}{k}\Large]$. For confidence, we divide probabilities between [0, 1] into the number of bins, and calculate accuracy using indicator function that takes value of 1 when the prediction was correct and 0 otherwise, by these formulas:

\begin{subequations}\label{eq:ece_sub}
\begin{align}
    \text{acc}(B_i) &= \frac{1}{\abs{B_i}} \sum_{i \in B_i} \mathds{1}(y_i = y_{pred}) \label{eq:ece_acc} \\
    \text{conf}(B_i) &= \frac{1}{\abs{B_i}} \sum_{i \in B_i} p(y_{pred} | x) \label{eq:ece_conf}
\end{align}
\end{subequations}

\subsubsection{UCE}
\label{appendix:uce}

UCE can defined as in Equation~\ref{eq:uce}:

\begin{equation}
    \text{UCE} = \sum^{k}_{i=1} \frac{\abs{B_{i}}}{N} \abs{\text{err}(B_i) - \text{unc}(B_i) },
    \label{eq:uce}
\end{equation}

where $\text{err}(B_i)$ is for bi$n$-wise error of the model and 
$\text{unc}(B_i)$ bi$n$-wise obtained uncertainty of model:

\begin{subequations} 
\label{eq:uce_sub}
\begin{align}
    \text{err}(B_i) &= \frac{1}{\abs{B_i}} \sum_{i \in B_i} \mathds{1}(y_i \neq y_{pred})     \label{eq:uceerr} \\
    \text{unc}(B_i) &= \frac{1}{\abs{B_i}} \sum_{i \in B_i} \hat{u}_i(p) \label{eq:uceuncert}
\end{align}
\end{subequations}

\subsubsection{RCE}
\label{appendix:rce}

RCE metric can be empirically calculated as:

\begin{equation}
    \text{RCE} =  \frac{1}{N} \sum^{N}_{i=1}  \abs{\hat{P}(reg(U) \geq reg(u_i)) - \hat{P}(U \leq u_i) },
    \label{eq:rce}
\end{equation}

where the regression function can be defined as:

\begin{subequations} 
\label{eq:rce_sub}
\begin{align}
    \hat{P}(reg(U) \geq reg(u_i)) &= \frac{1}{B_i - 1} \sum_{b' \neq b} \mathds{1}[acc(b') \geq acc(b)]     
    \label{eq:rceconf} \\
    \hat{P}(U \leq u_i) &= \frac{1}{B_i - 1} \sum_{b' \neq b} \mathds{1}[unc(b') \leq unc(b)] \label{eq:rcesub}
\end{align}
\end{subequations}

\subsubsection{Class-wise ECE}
\label{appendix:class_wise_ece}

Given the above definition of ECE, the class-wise extension of ECE can be defined as in Equation~\ref{eq:cece}:

\begin{equation}
    \text{cECE} = \sum^{C}_{c=1} \frac{1}{C} \text{ECE}(c) ,
    \label{eq:cece}
\end{equation}

where ECE metric function applied per class $C$ of the dataset, and by taking their mean per class \cite{kull2019beyond}.

\subsubsection{ACE}
\label{appendix:ace}

ACE can be defined as:

\begin{equation}
    \text{ACE} =  \frac{1}{CR} \sum^{C}_{c=1} \sum^{R}_{r=1}  \abs{\text{acc}(B_{cr}) - \text{conf}(B_{cr}) },
    \label{eq:ace}
\end{equation}

where $r$ is the calibration range for class $c$.

\subsubsection{SmoothECE}
\label{appendix:smece}

SmoothECE, also defined as smECE, can be obtained by applying Gaussian Kernel Smoothing instead binning as follows:

\begin{equation}
    \text{smECE}_{K, \sigma} = \int \frac{1}{N} \sum_{i}^N \abs{K(t, \text{conf}(y_i))( \text{conf}(y_i) - \text{acc}(y_i))} dt,
    \label{eq:sm_ece}
\end{equation}

where $K(t_1, t_2) = \text{exp}(-(t_1 - t_2)^2 / 2 \sigma) / \sqrt{2\pi\sigma}$, $\sigma$ is the bandwidth or smoothing scale, where $\sigma \mapsto \text{smECE}_{\sigma}(\{y_i, y_{pred}\})$ is a non-increasing function and $\sigma^{*} = \text{smECE}_{\sigma^{*}}(\{y_i, y_{pred}\})$ that can be found using binary search. The goal is to approximate the true Wasserstein distance (Earth Mover's Distance)  to the distribution of perfect calibration.

\subsubsection{Brier Score}
\label{appendix:exp_eval_brier}
Brier score (BS) can be defined as the following probabilistic formulation:
\begin{equation}
    \text{BS} = \frac{1}{N} \sum_{i}^N (\text{acc}(y_i, y_{pred}) - \text{conf}(y_{pred}, x_i))^2
    \label{eq:brier_score_eq}
\end{equation}

where $N$ is the number of samples, $\text{conf}(y_{pred}, x_i)$ model's prediction confidence for the token at the correct position, $\text{acc}(y_i, y_{pred})$ is the shows the accuracy of inference, that it gets 0 when the result is not correct and 1 if it is correct. Another name for Brier score is Mean Squared Error (MSE) \cite{detommaso2024multicalibrationiglb}.

\subsubsection{BLEU Score and EM}
\label{appendix:exp_eval_bleu}

The exact match (EM) and BLEU score \cite{papineni2002bleu} can be used to show accuracy and confidence of language model. EM is the ratio of the test samples that model accurately inferred the entire sequence. BLEU Score is denoted as follows:

\begin{equation}
    \text{BLEU} = \text{BP} * \text{exp}\Big( \sum_{i}^N w_i \text{log} p_n \Big),
    \label{eq:bleu_score}
\end{equation}

where $w_i$ are positive weights that sum to one, $p_i$ are geometric average of $n$-gram precisions, where $n$-gram is the sequence of $n$ words, that is defined as follows:

\[
\centering
p_n = \frac{{\sum_{n\text{-gram} \in y_{pred}} Count_{clip}(n\text{-gram})}}{{\sum_{n\text{-gram}' \in {y_{pred}'}} Count(n\text{-gram}')}}
\]

where $Count(\text{$n$-gram'})$ is the number of $n$-grams in the prediction, $Count_{clip}(\text{$n$-gram})$ is the number of $n$-grams in the prediction that appear in the target sequence, but their count is clipped by the maximum amount of appearance in the target sequence. As a baseline, $N$ can be set to 4 and $w_n$ to $1/N$\cite{papineni2002bleu}. BP part is the exponential Brevity Penalty for the short translation and can be defined as:

\[
    \text{BP} 
    =
    \begin{cases}
        1 & \text{if } c > r \\
        e^{(1-r/c)} & \text{if } c \leq r
    \end{cases}
    ,
\]

where $c$ is the length of the prediction and $r$ is the length of the target sequence. BLEU score represents the fluency and adequacy of the sequence generation \cite{papineni2002bleu}. 

\subsubsection{Area Under the Receiver Operator Curve (AUROC)}

For the calculation of the AUROC metric, we need to aggregate inference outcomes according to decision thresholds as in binary misprediction detection task on confidence scores \cite{ulmer2024apricot}. This curve put threshold as a boundary between false positive (FPR) and true positive rate (TPR) of the inference. We assume that there exists a threshold $t$ so that our prediction set $Y_{t}$ contains only the predictions that the model made right or true positive. Formula:

\begin{equation}
    \text{AUROC} = \int_0^1 \text{TPR}(\text{FPR}^{-1}(t))\text{d}t.
\end{equation}

\subsubsection{AURC}
\label{appendix:aurc}

Given the subset $S$ of high confidence predictions is greater than rank $r$ from subset $R$, we calculate the tradeoff between subset size with $K$ highest elements and accuracy as:

\begin{equation}
    \text{AURC}(R, S) = \sum^{N}_{K=1} acc(S_K)\frac{K}{N}.
    \label{eq:aurc}
\end{equation}

\subsection{\reviewerTwo{Semantic Entropy Baseline}}
\label{appendix:semantic_entropy}

\begin{table*}[!t]
    \centering
    \caption{\reviewerTwo{Semantic Entropy (SE) baseline results on TriviaQA and Natural Questions (NQ) for Llama-3.1-8B-Instruct, Mistral-7B-Instruct-v0.3, and GPT-5.2. SE samples $K{=}10$ generations at temperature $1.0$, clusters them by bidirectional NLI entailment (DeBERTa-large-MNLI), and reports the cluster-marginal entropy as a confidence score. For the two open-box models we compare SE against sequence likelihood (Seq.\ Lik.) and chain-of-thought sequence likelihood (CoT Seq.\ Lik.); for the closed-box GPT-5.2 we compare against verbalized qualitative confidence (Verb.\ Qual), reproduced from Tables~\ref{tab:calibration_results} and~\ref{tab:calibration_results_nq}. BLEU is identical across the Seq.\ Lik.\ and SE rows within a model/dataset block because SE only re-scores the greedy generation used by sequence likelihood; the CoT row uses a different generation, hence a different BLEU. Best result per model/dataset column is in \textbf{bold}.}}
    \label{tab:semantic_entropy_results}
    \setlength{\tabcolsep}{3pt}
    \renewcommand{\arraystretch}{0.95}
    \footnotesize
\begin{tabular}{
    >{\arraybackslash}p{0.13\textwidth}
    >{\centering\arraybackslash}p{0.05\textwidth}
    >{\centering\arraybackslash}p{0.06\textwidth}
    >{\centering\arraybackslash}p{0.04\textwidth}
    >{\centering\arraybackslash}p{0.06\textwidth}
    >{\centering\arraybackslash}p{0.05\textwidth}
    >{\centering\arraybackslash}p{0.05\textwidth}
    >{\centering\arraybackslash}p{0.05\textwidth}
    >{\centering\arraybackslash}p{0.06\textwidth}
    >{\centering\arraybackslash}p{0.04\textwidth}
    >{\centering\arraybackslash}p{0.06\textwidth}
    >{\centering\arraybackslash}p{0.05\textwidth}
    >{\centering\arraybackslash}p{0.05\textwidth}
}
\toprule
\textbf{Models \& Baselines} &
\multicolumn{6}{c}{\textbf{TriviaQA}} & \multicolumn{6}{c}{\textbf{NQ}} \\
\cmidrule(lr){2-7} \cmidrule(lr){8-13}
& ECE $\downarrow$ & smECE $\downarrow$ & BS $\downarrow$ & AUROC $\uparrow$ & AURC $\downarrow$ & BLEU $\uparrow$
& ECE $\downarrow$ & smECE $\downarrow$ & BS $\downarrow$ & AUROC $\uparrow$ & AURC $\downarrow$ & BLEU $\uparrow$ \\
\midrule
\rowcolor[HTML]{EFEFEF} \multicolumn{13}{l}{\textbf{Llama-3.1-8B-Instruct}} \\
\reviewerTwo{Seq.\ Lik.}         & \reviewerTwo{\bf 0.045} & \reviewerTwo{\bf 0.039} & \reviewerTwo{0.287}     & \reviewerTwo{0.505}     & \reviewerTwo{0.523}     & \reviewerTwo{\bf 0.041} & \reviewerTwo{0.197}     & \reviewerTwo{0.126}     & \reviewerTwo{0.417}     & \reviewerTwo{0.514}     & \reviewerTwo{0.783}     & \reviewerTwo{0.018} \\
\reviewerTwo{CoT Seq.\ Lik.}     & \reviewerTwo{0.087}     & \reviewerTwo{0.065}     & \reviewerTwo{\bf 0.273} & \reviewerTwo{0.531}     & \reviewerTwo{\bf 0.436} & \reviewerTwo{0.033}     & \reviewerTwo{0.429}     & \reviewerTwo{0.360}     & \reviewerTwo{0.410}     & \reviewerTwo{0.520}     & \reviewerTwo{\bf 0.736} & \reviewerTwo{\bf 0.020} \\
\reviewerTwo{Semantic Entropy}   & \reviewerTwo{0.213}     & \reviewerTwo{0.208}     & \reviewerTwo{0.436}     & \reviewerTwo{\bf 0.542} & \reviewerTwo{0.484}     & \reviewerTwo{\bf 0.041} & \reviewerTwo{\bf 0.028} & \reviewerTwo{\bf 0.024} & \reviewerTwo{\bf 0.199} & \reviewerTwo{\bf 0.577} & \reviewerTwo{0.737}     & \reviewerTwo{0.018} \\
\midrule
\rowcolor[HTML]{EFEFEF} \multicolumn{13}{l}{\textbf{Mistral-7B-Instruct-v0.3}} \\
\reviewerTwo{Seq.\ Lik.}         & \reviewerTwo{0.044}     & \reviewerTwo{0.043}     & \reviewerTwo{\bf 0.265} & \reviewerTwo{0.621}     & \reviewerTwo{0.367}     & \reviewerTwo{0.228}     & \reviewerTwo{0.328}     & \reviewerTwo{0.298}     & \reviewerTwo{0.442}     & \reviewerTwo{0.633}     & \reviewerTwo{0.631}     & \reviewerTwo{\bf 0.080} \\
\reviewerTwo{CoT Seq.\ Lik.}     & \reviewerTwo{\bf 0.025} & \reviewerTwo{0.034}     & \reviewerTwo{0.283}     & \reviewerTwo{\bf 0.702} & \reviewerTwo{\bf 0.306} & \reviewerTwo{\bf 0.310} & \reviewerTwo{0.215}     & \reviewerTwo{0.211}     & \reviewerTwo{0.438}     & \reviewerTwo{\bf 0.652} & \reviewerTwo{\bf 0.612} & \reviewerTwo{0.059} \\
\reviewerTwo{Semantic Entropy}   & \reviewerTwo{\bf 0.025} & \reviewerTwo{\bf 0.025} & \reviewerTwo{0.480}     & \reviewerTwo{0.559}     & \reviewerTwo{0.431}     & \reviewerTwo{0.228}     & \reviewerTwo{\bf 0.023} & \reviewerTwo{\bf 0.025} & \reviewerTwo{\bf 0.256} & \reviewerTwo{0.510}     & \reviewerTwo{0.717}     & \reviewerTwo{\bf 0.080} \\
\midrule
\rowcolor[HTML]{EFEFEF} \multicolumn{13}{l}{\textbf{GPT-5.2}} \\
\reviewerTwo{Verb.\ Qual}        & \reviewerTwo{\bf 0.201} & \reviewerTwo{\bf 0.101} & \reviewerTwo{\bf 0.236} & \reviewerTwo{0.595}     & \reviewerTwo{0.354}     & \reviewerTwo{\bf 0.458} & \reviewerTwo{0.342}     & \reviewerTwo{\bf 0.188} & \reviewerTwo{0.411}     & \reviewerTwo{\bf 0.648} & \reviewerTwo{0.492}     & \reviewerTwo{\bf 0.120} \\
\reviewerTwo{Semantic Entropy}   & \reviewerTwo{0.328}     & \reviewerTwo{0.262}     & \reviewerTwo{0.361}     & \reviewerTwo{\bf 0.624} & \reviewerTwo{\bf 0.203} & \reviewerTwo{\bf 0.458} & \reviewerTwo{\bf 0.263} & \reviewerTwo{0.225}     & \reviewerTwo{\bf 0.320} & \reviewerTwo{0.603}     & \reviewerTwo{\bf 0.472} & \reviewerTwo{\bf 0.120} \\
\bottomrule
\end{tabular}
\end{table*}

\reviewerTwo{To complement the descriptive treatment of meaning-level uncertainty in Section~\ref{sec:llm_uncertainty}, we report a Semantic Entropy (SE) baseline~\cite{kuhn2023semanticuncertainty,farquhar2024detecting} on the same open-ended generative QA setup used in our main evaluation. We evaluate on TriviaQA and Natural Questions (NQ) with two open-box backbones (Llama-3.1-8B-Instruct and Mistral-7B-Instruct-v0.3) and the closed-box GPT-5.2. For each question we sample $K{=}10$ generations at temperature $1.0$, cluster them into semantic-equivalence classes via bidirectional NLI entailment using DeBERTa-large-MNLI, and aggregate length-normalized sequence log-likelihoods within each cluster via log-sum-exp before computing the cluster-marginal entropy. Confidence is reported as $1-H/\log M$, where $M$ is the observed number of clusters; lower entropy maps to higher confidence. For the open-box models we compare SE against sequence likelihood (Seq.\ Lik.) and chain-of-thought sequence likelihood (CoT Seq.\ Lik.); for GPT-5.2 we compare against verbalized qualitative confidence (Verb.\ Qual). Table~\ref{tab:semantic_entropy_results} reports ECE, smECE, BS, AUROC, AURC, and BLEU under an identical protocol.}

\reviewerTwo{The dominant pattern is that SE's benefit is \emph{dataset- and metric-dependent} rather than uniform. On the more ambiguous NQ benchmark, where sequence likelihood over-concentrates mass on lexically distinct rewordings of the same answer, SE reduces ECE by roughly 7--14$\times$ on both open-box models (Llama: $0.197\to0.028$; Mistral: $0.328\to0.023$) and also lowers Brier score (Llama: $0.417\to0.199$; Mistral: $0.442\to0.256$), matching the prediction of~\cite{kuhn2023semanticuncertainty,farquhar2024detecting} that meaning-level aggregation is particularly informative under high paraphrase variability. On TriviaQA, by contrast, Seq.\ Lik.\ is already near-calibrated ($\mathrm{ECE}\!\le\!0.05$) so the $K$-sample clustering step introduces additional variance and hurts ECE / BS, while still yielding a modest AUROC gain for Llama ($0.505\to0.542$). CoT Seq.\ Lik.\ occupies a distinct point on the Pareto front: it dominates AUROC/AURC on Mistral TriviaQA ($0.702$ / $0.306$) and remains the strongest single-pass ranking signal on NQ, at the cost of a slightly worse ECE than SE when the task is ambiguous.}

\reviewerTwo{On the closed-box GPT-5.2, SE improves selective-prediction quality noticeably over Verb.\ Qual --- AUROC rises from $0.595$ to $0.624$ and AURC drops from $0.354$ to $0.203$ on TriviaQA, with a similar AURC gain on NQ ($0.492\to0.472$) --- but it underperforms on ECE for TriviaQA ($0.328$ vs $0.201$). This reflects a well-known SE trade-off: the cluster-marginal entropy is a strong ranking signal but a coarse absolute-probability estimate, especially when the base model already emits well-formed verbal confidences. BLEU is identical across the Seq.\ Lik.\ / SE rows within a model/dataset block, because SE only re-scores the greedy generation rather than changing it; the CoT row differs because it uses a different generation.}

\reviewerTwo{These trends carry two practical implications that we integrate into Section~\ref{sec:evaluation} and Section~\ref{sec:future_research}. First, no single family (token likelihood, CoT likelihood, semantic clustering, or verbalized qualitative) dominates across calibration \emph{and} ranking on open-ended generative QA; the appropriate baseline depends on downstream use (selective prediction vs.\ calibrated risk score) and on whether the base likelihood is already well-calibrated. Second, SE's $K$-sample overhead ($10\times$ generation cost plus $\mathcal{O}(K^2)$ NLI calls per question) and its reliance on an off-the-shelf NLI classifier bound its practicality: false-negative entailment edges over-fragment the cluster space and inflate entropy estimates, particularly on short-span TriviaQA answers where lexical proximity is not a reliable proxy for semantic identity. These observations reinforce the open-challenge framing of meaning-level uncertainty in Section~\ref{sec:llm_uncertainty} and the practical-Pareto discussion in Section~\ref{sec:future_research}.}

\subsection{Additional results}

This section contains the reliability diagrams for baselines on on TruthfulQA dataset. Figure~\ref{fig:llama_truthful_qa_comparison} shows evaluation results for Llama 3.1, Figure~\ref{fig:mistral_truthful_qa_comparison} shows experimental results for Mistral v0.3,Figure~\ref{fig:qwen2.5_truthful_qa_comparison} shows results for Qwen2.5, Figure~\ref{fig:gpt4_truthful_qa_comparison} and Figure~\ref{fig:gpt4o_truthful_qa_comparison} shows results for  GPT-4 and GPT-4o models respectively.

\vskip 0.2in

\end{document}